\documentclass{article}
\usepackage{arxiv}

\usepackage{comment}
\usepackage[utf8]{inputenc} % allow utf-8 input
\usepackage[T1]{fontenc}    % use 8-bit T1 fonts
\usepackage{hyperref}       % hyperlinks
\usepackage{url}            % simple URL typesetting
\usepackage{booktabs}       % professional-quality tables
\usepackage{amsfonts}       % blackboard math symbols
\usepackage{nicefrac}       % compact symbols for 1/2, etc.
\usepackage{microtype}      % microtypography
\usepackage{lipsum}		% Can be removed after putting your text content
\usepackage{graphicx}
\usepackage{natbib}
\usepackage{doi}
\usepackage{mathtools}
\usepackage{subfig}
\usepackage{colortbl}
\usepackage{multirow}
\usepackage{supertabular,booktabs}
\usepackage[x11names,dvipsnames,table]{xcolor} % for use in color links
\usepackage{tabularray}
\usepackage{tikz}
\usetikzlibrary{shapes.geometric, arrows}
\usepackage{forest}
\usepackage{tikz-qtree}
\usepackage{hyperref}
\usepackage{lscape}
\usepackage{longtable}
\usepackage{rotating} % for rotating the table

\newcommand{\subfour}[1]{\vspace*{3mm}{\noindent\bf #1}} 

\begin{document}

\title{Visual Question Answering: From Early Developments to Recent Advances - A Survey}

\author{ 
        {\hspace{1mm}Ngoc Dung Huynh} \\
	Deakin University\\
	Australia \\
        \texttt{ndhuynh@deakin.edu.au}\\
	\And
        {\hspace{1mm}Mohamed Reda Bouadjenek} \\
	Deakin University\\
	Australia \\
        \texttt{reda.bouadjenek@deakin.edu.au}\\ 
        \And
        {\hspace{1mm}Sunil Aryal} \\
	Deakin University\\
	Australia \\
        \texttt{sunil.aryal@deakin.edu.au}\\ 
	\AND
        {\hspace{1mm}Imran Razzak} \\
        Mohamed bin Zayed University of Artificial Intelligence,  Abu Dhabi, \\
        University of New South Wales, Sydney, \\
        \texttt{imran.razzak@mbzuai.ac.ae}\\
	\And
        {\hspace{1mm}Hakim Hacid} \\
	Technology Innovation Institute\\
	United Arab Emirates \\
        \texttt{hakim.hacid@tii.ae}
}

\maketitle

\begin{abstract}

With the rapid development of research on multimodal data, VQA has  attracted a lot of attention from the research community. 
VQA is a growing field of research that aims at enabling machines 
% \sout{combines computer vision and natural language processing techniques to enable machines} 
to answer questions about visual content. 
This task requires reasoning capabilities and a combination of image and language processing techniques, including feature extraction, object detection, text embedding, natural language understanding, and language generation. 
VQA has diverse applications ranging from creating interactive educational tools, diagnosing medical images, and aiding customer service, to enhancing entertainment experiences and generating captions for social media content. 
VQA can also be used to help and support blind and visually impaired individuals to ``see'' images through natural language descriptions, providing them with access to visual content that they would otherwise be unable to perceive.
In this survey paper, we introduce a taxonomy for VQA architectures based on their key components and design choices, which provides a structured framework for comparing and evaluating different VQA approaches. 
We then analyze the various approaches to VQA, including deep learning-based methods, and provide a critical review of the current state of the field. 
We also review the emerging field of Large Visual Language Model (LVLM) models, which have shown promising results in a variety of multimodal tasks such as VQA. 
Additionally, we examine existing VQA datasets and the evaluation metrics used to assess the performance of VQA systems before discussing some real-world applications of VQA in various domains.
Finally, we discuss the current challenges and future directions of VQA research, including open research questions and potential applications. 
This survey paper serves as a valuable resource for researchers and practitioners who are interested in understanding the current state-of-the-art in VQA and exploring future research opportunities in VQA.

\end{abstract}

% Note that keywords are not normally used for peer-review papers.

% make the title area
\maketitle

\section{Introduction}
\label{sec:introduction}

Humans are capable of processing information from their surroundings through various sensory modalities, such as hearing, smelling, vision, and touch. 
Although these types of data are absorbed individually and are incompatible, humans have the remarkable ability to align and fuse them to better sense and understand the world around them. 
For instance, when watching a TV show, a human can simultaneously process both the visual and auditory components of the program to enhance their comprehension and enjoyment of the content. 
This integration of different sensory modalities is critical to our ability as humans to perceive and interpret the world effectively, and it highlights the incredible flexibility and adaptability of the human brain.

% In recent years, multi-modal computing has become a research topic attracting much attention from researchers. 

Multi-modal computing is a field of research that has attracted a lot of attention from the research community~\cite{8269806,8237712,bagher-zadeh-etal-2018-multimodal,Zellers_2022_CVPR,ektefaie2023multimodal}, which aims to replicate this ability in machines, by developing algorithms that can integrate information from multiple sources, such as images, audio, and text. 
By enabling machines to process information from different modalities, multi-modal computing can enhance their ability to understand and interact with the world in a more human-like manner.
% It works as the human way of receiving information is to process multiple input types, such as vision and sound, at the same time. 
This approach has paved the way for the creation of advanced algorithms that can achieve new capabilities that were previously inaccessible to computers. 
For example, computers can now figure out which object in an image has the best relevance to a text query.
This integration of different sensory modalities is critical to the development of more intelligent and intuitive machines, and it has the potential to transform many industries, from healthcare to entertainment.

\begin{figure}[t]
\centering  
\includegraphics[width=1\linewidth]{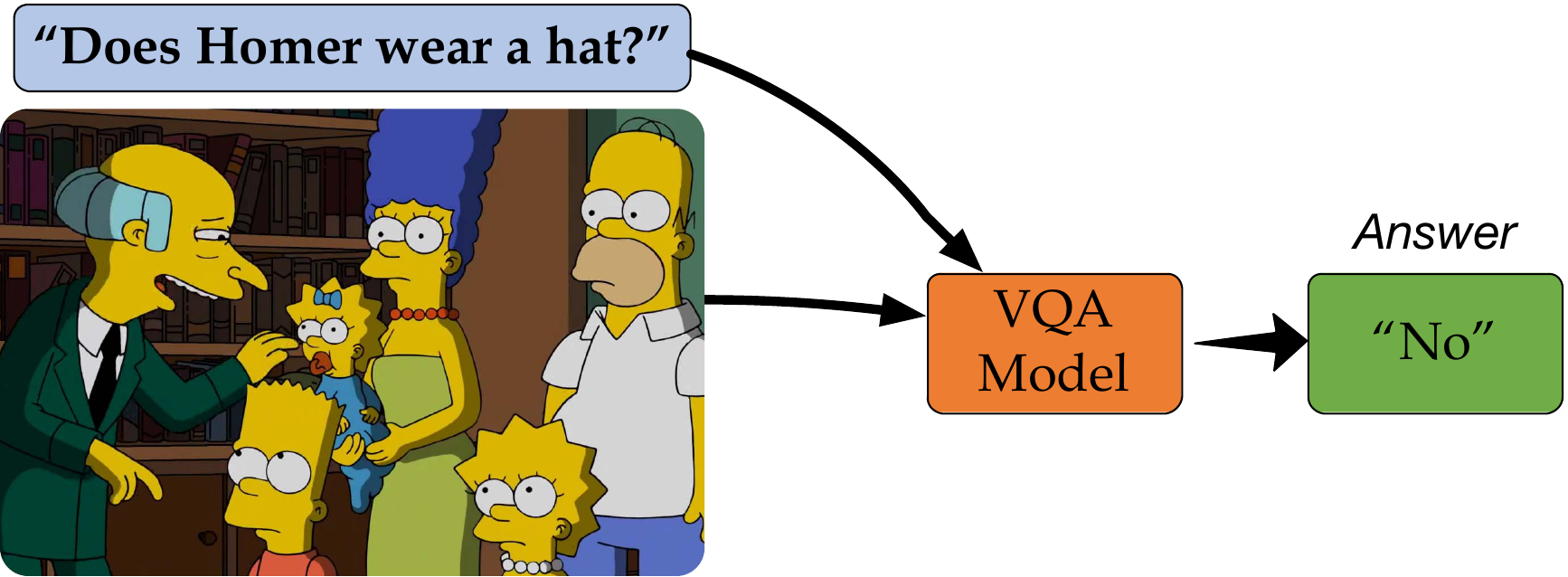}
\caption{Overview of a VQA system.}
\label{fig:VQA_architecture}
\vspace{-10pt}
\end{figure}
% \end{wrapfigure}

\begin{table*}[t]
 \caption{Classification of VQA questions based on the type of information required to answer them.}
  % \rowcolors{2}{gray!10}{white}
  \rowcolors{2}{gray!10}{white}
  \resizebox{\linewidth}{!}{%
  \begin{tabular}{lll}
  \toprule
    \textbf{Category} &  \textbf{Description} & \textbf{Example -- related to Figure~\ref{fig:VQA_architecture}.} \\
  \midrule
    \textbf{Closed-ended} & Fixed set of answers &  \\
    \hspace{3mm}Single-choice & Only one correct answer & ``What color is Mr. Burns suit? (a) Blue (b) Red (c) Green '' \\
    \hspace{3mm}Multiple-choice & Multiple possible answers & ``What is shown in the image? (a) baby (b) persons (c) cup''\\
    \hspace{3mm}Yes/No (true/false) & Only two possible answers & ``Are there kids in the image?''\\
    \textbf{Open-ended} & No fixed set of answers, often requiring NL answers &  \\
    \hspace{3mm}Factual  & Factual information about the image & ``What color is Marge's dress?'' \\
    \hspace{3mm}Reasoning & Require reasoning or inference to answer & ``How is the man on the right feeling?'' \\
    \hspace{3mm}Spatial  & About the spatial relationship between objects or regions & ``Who is on the right of the image?'' \\
    \hspace{3mm}Comparative  & Comparison between objects or regions & ``Who between Lisa and Bart is the tallest?'' \\
    \hspace{3mm}Attribute classification & About the attributes or properties of objects & ``What is the shape of Marge's hair?'' \\
    \hspace{3mm}Action recognition  & About actions or events  & ``What are these people doing?'' \\
    \hspace{3mm}Counting & About the number of objects or regions & ``How many people are there in the image?'' \\
    \hspace{3mm}Object detection & Information about objects & ``Is there a dog in the image?'' \\
       \bottomrule

  \end{tabular}
}%
    \label{table:AI problems} 
\end{table*}

Visual Question Answering (VQA) is an example of a multi-modal Visual-Language (VL) task~\cite{wang2021simvlm,alayrac2022flamingo,PARK2023100548} that combines computer vision and natural language processing techniques to enable machines to answer questions about visual content~\cite{Antol_2015_ICCV,ren2015exploring,Zhang_2016_CVPR,ma2016learning,Goyal_2017_CVPR}. 
For instance, as shown in Figure~\ref{fig:VQA_architecture}, given an image of ``The Simpsons'' TV show and a natural language question such as ``Does Homer wear a hat?'', a VQA system should be able to recognize  ``Homer'' in the image and generate the correct answer, ``No''. 
Supported by various large-scale datasets~\cite{VQA:2014,Goyal_2017_CVPR,gva,young-etal-2014-image}, the VQA task has allowed remarkable advancements in a diverse range of applications including creating interactive educational tools~\cite{8100054}, diagnosing medical images~\cite{lin2021medical,10.1145/3460426.3463584,8987108, Allaouzi2019AnEM, Gong2021SYSUHCPAV,ImageCLEFVQA-Med2018,Lau2018ADO,ImageCLEF-VQA-Med2021}, aiding customer service~\cite{8269806}, enhancing entertainment experiences~\cite{Gordon_2018_CVPR}, and generating captions for social media content~\cite{Vinyals_2015_CVPR}.
VQA is widely recognized as one of the most challenging tasks in the field of VL computing, particularly when compared to other VL tasks like Visual Grounding~\cite{Mao_2016_CVPR,Deng_2021_ICCV} and Visual Captioning~\cite{8620348,Gan_2017_CVPR}. 
One of the reasons for this is that VQA's input question form is much more diverse than other VL tasks. 
While the text queries in other tasks like Visual Grounding are fixed and describe specific objects or features in an image, VQA questions can vary widely in format and content. 
Table~\ref{table:AI problems} provides examples of VQA questions organized by their respective AI problem domains, where we observe that VQA can tackle a wide range of AI problems, including image classification, object detection, counting, scene classification, attribute classification, activity recognition, and reasoning understanding.

Several surveys on VQA have been published in recent years to review and evaluate different aspects of VQA (e.g.,~\cite{zhang2019information, kallooriyakath2020visual, manmadhan2020visual, zou2020survey, sharma2021survey}). 
In particular, Zhang et al.~\cite{zhang2019information} and Kallooriyakath et al.~\cite{kallooriyakath2020visual} focused on reviewing existing fusion techniques, while Manmadhan and Kovoor~\cite{manmadhan2020visual} provided a comprehensive view of VQA, covering vision encoder, language encoder, fusion strategies, datasets, and evaluation metrics. 
% Zou et al. (2020) \cite{zou2020survey} attended to review available datasets on VQA with their limitations. 
% Sharma and Jalal (2021) \cite{sharma2021survey} did not only review aspects of VQA's main branch, but they also reviewed the extensions of VQA consisting of text-VQA and Knowledge-based VQA. 
 Zou and Xie~\cite{zou2020survey}  reviewed available datasets on VQA with their limitations. 
Sharma and Jalal~\cite{sharma2021survey} extended their review beyond traditional VQA models and included text-VQA and knowledge-based VQA.

However, with the rapid progress of the field, combined with the introduction of new techniques, models, and datasets, we argue that there is a need for an up-to-date and comprehensive survey on the topic. 
First, the surveys mentioned above have focused only on specific aspects of VQA, such as fusion techniques or datasets, and do not provide a comprehensive view of the field. 
Second, there has been a significant advancement in VQA models with the introduction of novel approaches such as LVLM (Visual Language Pre-training) models like ViLBERT~ \cite{lu2019vilbert}, VisualBERT~\cite{li2019visualbert}, and VL-BERT~\cite{su2019vl}. 
These models are pre-trained on a large amount of image and language data to learn the joint representation of vision and language, and then, are fine-tuned on Visual-Language tasks such as VQA. 
By doing so, these models have significantly improved the accuracy of the VQA task.
To the best of our knowledge, these models have not been reviewed in any existing surveys.
Third, VQA has been applied in diverse domains such as healthcare, education, and customer service, which have their unique challenges and require specific solutions. 
Finally, VQA's practical applications have been extended to more complex tasks, such as multi-modal dialogue systems and open-domain VQA, which require a different approach from traditional VQA. 
Therefore, a new survey on VQA that covers recent advancements, practical applications, and emerging trends is necessary to provide researchers and practitioners with a comprehensive overview of the field.

The rest of this paper is organized as follows:
First, Section~\ref{sec:the_vqa_task_and_taxonomy_of_common_architectures} presents an introduction to the VQA task that we address in this paper, 
% along with other VQA tasks that may belong to distinct categories.
% Then, Section \ref{sec:vqa_architecture} presents
and 
a taxonomy of VQA systems based on Vision Encoder, Language Encoder, and techniques for fusing image and question features to generate answers, including LVLM models. 
Following this, in Section \ref{datasets}, some of the most widely used datasets in VQA research are explored. 
Next, in Section \ref{metrics}, the most commonly used evaluation metrics for VQA models are discussed. 
Section \ref{applications} provides an overview of various real-world applications of VQA in fields such as medical imaging, visually impaired individuals, cultural heritage, advertisement, and education. 
Section \ref{Discussion} analyzes and compares existing VQA models based on accuracy and offers future work directions for building VQA models and real-world VQA applications. 
Finally, the paper is concluded in Section \ref{sec:conslusion}.

\section{The VQA Task and Taxonomy of Common Architectures}
\label{sec:the_vqa_task_and_taxonomy_of_common_architectures}

VQA is a task that requires the system to fuse information from both visual and linguistic modalities to produce a single answer, making it a prime example of a multi-modal computing task.
Over the years, various sub-fields of VQA have gained attention from researchers. 
These include OCR-VQA (or Text-VQA), Knowledge-Based VQA (or KB-VQA), and Answer Grounding VQA, which require additional technologies or knowledge beyond traditional VQA to answer questions related to images. 
OCR-VQA involves integrating Optical Character Recognition (OCR) technologies into VQA models to recognize and interpret text within images to answer questions~\cite{singh2019towards, wang2022tag, 8978122, tanaka2021visualmrc, biten2019scene, mathew2021docvqa}. 
On the other hand, KB-VQA utilizes external knowledge from sources like Wikipedia to answer complex questions beyond image information~\cite{wang2015explicit, wu2016ask, marino2019ok, schwenk2022okvqa}. 
Finally, Answer Grounding VQA identifies relevant regions or objects in an image to answer questions instead of providing textual answers like other VQA systems~\cite{chen2022grounding, TolokaWSDMCup2023}.
The primary focus of this paper is to survey the conventional VQA task, which involves providing an accurate natural language answer given only a natural language question and an associated image.

\begin{figure}[t]
  \centering
  \includegraphics[width=\linewidth]{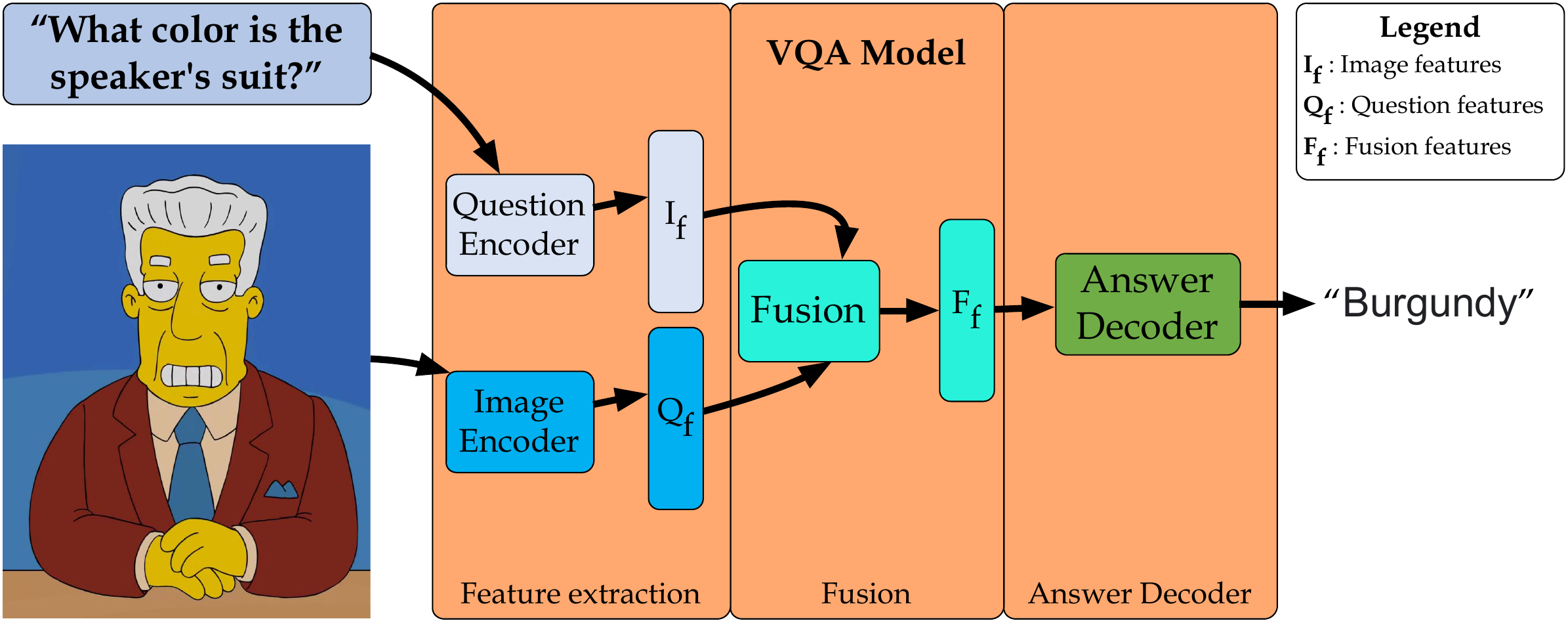}
  \caption{General VQA system.}
  \label{fig:VQA_system}
\end{figure}

As illustrated in Figure~\ref{fig:VQA_system}, a typical VQA system often consists of three stages: Feature Extraction, Fusion, and Answer Decoder. 
The first two steps involve the extraction of visual features and textual representations from the input image and question, respectively. 
The Vision Encoder is responsible for extracting visual features from the image using Computer Vision techniques such as CNNs, Vision Transformers, etc. The Language Encoder, on the other hand, extracts textual representations from the input question using Natural Language Processing (NLP) techniques like RNNs, Transformers, etc. 
Further details on both encoders can be found in Section~\ref{sec:Vision_Encoder} and Section~\ref{sec:Language_Encoder}, respectively.
The output vectors from the Image and Language Encoders are then passed through a fusion component, which combines them using a technique such as element-wise product or concatenation~\cite{VQA:2014}. 
The resulting fused vector is then passed through the Answer Decoder to generate an answer to the given question based on the input image.
More information on this process can be found in Section~\ref{sec:Fusion_Machine}.
% A basic VQA system consists of these three phases and the overall process is illustrated in Figure~\ref{fig:VQA_system}. 
By using this framework, VQA systems can handle a wide range of questions and images, making them applicable in different scenarios.

\begin{figure*}
    \centering   
\tikzset{
    basic/.style  = {draw, text width=1cm, align=center, font=\sffamily, rectangle},
    root/.style   = {basic, rounded corners=2pt, thin, align=center, fill=green!30},
    onode/.style = {basic, thin, rounded corners=2pt, align=center, fill=green!60,text width=3cm,},
    tnode/.style = {basic, thin, align=left, fill=pink!60, text width=1.5cm, align=center},
    LVLMnode/.style = {basic,  rounded corners=2pt,thin, align=left, fill=purple!50, text width=2.5cm, align=center},
    LVLMnodeLong/.style = {basic,  rounded corners=2pt,thin, align=left, fill=purple!50, text width=5cm, align=center},
    RelationNetworkNode/.style = {basic,  rounded corners=2pt, thin, align=left, fill=cyan!50, text width=2.5cm, align=center},
    BilinearPoolingNode/.style = {basic,  rounded corners=2pt,thin, align=left, fill=lime!50, text width=2.5cm, align=center},    
    mnmnode/.style = {basic, thin,  rounded corners=2pt, align=left, fill=yellow!50, text width=2.5cm, align=center},
    xnode/.style = {basic, thin, rounded corners=2pt, align=center, fill=blue!20,text width=2.5cm,},    
    BoWNode/.style = {basic, thin, rounded corners=2pt, align=center, fill=brown!20,text width=2.5cm,}, 
    BoWNodeLong/.style = {basic, thin, rounded corners=2pt, align=center, fill=brown!20,text width=5cm,}, 
    RNNNode/.style = {basic, thin, rounded corners=2pt, align=center, fill=gray!40,text width=2.5cm,}, 
    RNNNodeLong/.style = {basic, thin, rounded corners=2pt, align=center, fill=gray!40,text width=5cm,}, 
    CNNNode/.style = {basic, thin, rounded corners=2pt, align=center, fill=teal!70,text width=2.5cm,},    
    TransformerNode/.style = {basic, thin, rounded corners=2pt, align=center, fill=magenta!60,text width=2.5cm,},
    TransformerNodeLong/.style = {basic, thin, rounded corners=2pt, align=center, fill=magenta!60,text width=5cm,},
    OpenEndednode/.style = {basic, thin, rounded corners=2pt, align=center, fill=olive!50,text width=2.5cm,},    
    CloseEndednode/.style = {basic, thin, rounded corners=2pt, align=center, fill=cyan!40,text width=2.5cm,},    
    GridBasedNode/.style = {basic, thin, rounded corners=2pt, align=center, fill=green!20,text width=2.5cm,},
    GridBasedNodeLong/.style = {basic, thin, rounded corners=2pt, align=center, fill=green!20,text width=5cm,},
    ObjectBasedNode/.style = {basic, thin, rounded corners=2pt, align=center, fill=pink!50,text width=2.5cm,},
    ObjectBasedNodeLong/.style = {basic, thin, rounded corners=2pt, align=center, fill=pink!50,text width=5cm,},
    ViTBasedNode/.style = {basic, thin, rounded corners=2pt, align=center, fill=yellow!80,text width=2.5cm,},
    ViTBasedNodeLong/.style = {basic, thin, rounded corners=2pt, align=center, fill=yellow!80,text width=5cm,},
    fusionnode/.style = {basic, thin, rounded corners=2pt, align=center, fill=red!50,text width=2.5cm,},
    fusionnode2/.style = {basic, thin, rounded corners=2pt, align=center, fill=red!50,text width=3.5cm,},
    fusionnode2Long/.style = {basic, thin, rounded corners=2pt, align=center, fill=red!50,text width=5cm,},
    attentionnode/.style = {basic, thin, rounded corners=2pt, align=center,  fill=orange!50,text width=2.5cm,},
    wnode/.style = {basic, thin, align=left, fill=pink!10!blue!80!red!10, text width=6.5em},
    edge from parent/.style={draw=black, edge from parent fork right}

}
\resizebox{0.7\linewidth}{!}{
\begin{forest} 
for tree={
    grow=east,
    growth parent anchor=west,
    parent anchor=east,
    child anchor=west,
    edge path={\noexpand\path[\forestoption{edge},->, >={latex}] 
         (!u.parent anchor) -- +(10pt,0pt) |-  (.child anchor) 
         \forestoption{edge label};}
}
% lsep is used for arrow distance
[VQA, basic,  l sep=7mm,
    [Answer Encoder Sec.~\ref{sec:Answer_Decoder}, xnode,  l sep=7mm 
    [Close-Vocabulary, CloseEndednode]
    [Open-Vocabulary, OpenEndednode,  l sep=7mm 
    [Open-End, OpenEndednode]
    [Multiple Choice, OpenEndednode]
    ]] 
    [Fusion Machine Sec.~\ref{sec:Fusion_Machine}, xnode,  l sep=7mm,        
        [LVLM\\ Sec.~\ref{LVLM}, LVLMnode, l sep=7mm, 
        [Masking Training, LVLMnode, l sep=7mm,
            [
            Flava~\cite{singh2022flava}{,} MaskVLM~\cite{kwon2022masked}
            I-JEPA \cite{assran2023self}{,} MAE \cite{he2022masked}, LVLMnodeLong
            ]
        ]
        [Contrasting learning, LVLMnode, l sep=7mm, 
            [
            CLIP~\cite{radford2021learning}{,} ALIGN~\cite{jia2021scaling}
            SigLIP \cite{zhai2023sigmoid} , LVLMnodeLong
            ]
        ]
        [Generative Models, LVLMnode, l sep=7mm, 
            [
            Chameleon~\cite{chameleonteam2024chameleonmixedmodalearlyfusionfoundation}{,} ChatGPT4o~\cite{openai2023gpt4} {,}
            Vilbert \cite{lu2019vilbert}{,} OFA\cite{wang2022ofa},LVLMnodeLong
            ]
        ]
        [Pre-Train LLM Backbones, LVLMnode, l sep=7mm, 
            [
            LLaVA~\cite{liu2023llava}{,} Frozen~\cite{tsimpoukelli2021multimodalfewshotlearningfrozen} {,}
            BLIP2. \cite{li2023blip2bootstrappinglanguageimagepretraining}{,} Qwen \cite{bai2023qwentechnicalreport},LVLMnodeLong
            ]
        ]
        ]
        [Relation Net. Sec.~\ref{sec:Relation_Networks}, RelationNetworkNode, l sep=7mm, [
        Simple RN~\cite{santoro2017simple}{,} Graph-VQA~\cite{teney2017graph}, RelationNetworkNode
        ]]
        [Bilinear Pooling Sec.~\ref{sec:bilinear_pooling}, BilinearPoolingNode, l sep=7mm, [
        MCB + Att~\cite{fukui2016multimodal}{,} MUTAN~\cite{ben2017mutan}, BilinearPoolingNode
        ]]
        [MNM\\ Sec.~\ref{sec:nmns}, mnmnode, l sep=7mm,[
        NMN \cite{andreas2016neural}{,} N2NMN \cite{hu2017learning}, mnmnode
        ]]
        [Attention-based\\ Sec.~\ref{Attention}, attentionnode, l sep=7mm,
        [Transformer, attentionnode, l sep=7mm,[
        MCAN \cite{yu2019deep}{,} MCAoA \cite{rahman2021improved}, attentionnode
        ]]
        [Co-Attention, attentionnode, l sep=7mm,[
        HieCoAtt \cite{lu2016hierarchical}{,} DAN \cite{nam2017dual},attentionnode
        ]]
        [Visual Attention, attentionnode, l sep=7mm,[
        SANs \cite{Yang_2016_CVPR}{,} BUTD \cite{anderson2018bottom},attentionnode ]]]
        [Simple Fusion Sec.~\ref{Simple Fusion}, fusionnode, l sep=7mm,
        [VQA \cite{VQA:2014}{,}  Neural-Image-QA \cite{malinowski2015ask}, fusionnode2Long]
        ]
        ]
    [Language Encoder Sec.~\ref{sec:Language_Encoder}, xnode,  l sep=7mm 
        [Transformers Sec.\ref{sec:Transformers}, TransformerNode,  l sep=7mm, [
        BERT\cite{devlin-etal-2019-bert}{,} RoBERTa \cite{DBLP:journals/corr/abs-1907-11692}{,} Llama\cite{touvron2023llama}{,} Vicuna \cite{vicuna2023},  TransformerNodeLong]
        ]
        [CNN Sec.~\ref{sec:cnn}, CNNNode]
        [RNN Sec.~\ref{Sec:RNN}, RNNNode,  l sep=7mm, [
            LSTM \cite{schmidhuber1997long}{,} Bi-LSTM \cite{650093} {,} GRU \cite{DBLP:journals/corr/ChoMGBSB14}, RNNNodeLong   
        ]]
        [BoW Sec.~\ref{sec:BoW}, BoWNode,  l sep=7mm, [
            Word2Vec \cite{mikolov2013efficient}{,} Glove \cite{pennington-etal-2014-glove}, BoWNodeLong   
    ]]
    ] 
    [Vision Encoder Sec.~\ref{sec:Vision_Encoder}, xnode,  l sep=7mm
        [ViT-based patch Sec.~\ref{sec:ViT-based_patch}, ViTBasedNode,  l sep=7mm  
        [ViT \cite{dosovitskiy2020image}{,} CLip \cite{radford2021learning}{,} SigLip \cite{zhai2023sigmoid},  ViTBasedNodeLong, l sep=7mm]]
        [Object-based\\ Sec.~\ref{sec:Object_based}, ObjectBasedNode,  l sep=7mm
        [Anchor-based \cite{Redmon_2016_CVPR}{,} Region-based \cite{NIPS2015_14bfa6bb}, ObjectBasedNodeLong]
        ]
        [Grid-based\\ Sec.~\ref{sec:Grid_based}, GridBasedNode,  l sep=7mm 
        [CNN {,}\\ Spatial Pyramid Pooling (SPP) \cite{7005506},GridBasedNodeLong]
    ]
    ] ]
\end{forest}
 }
    \caption{VQA Taxonomy.}
    \label{fig:lit_surv}
\end{figure*}
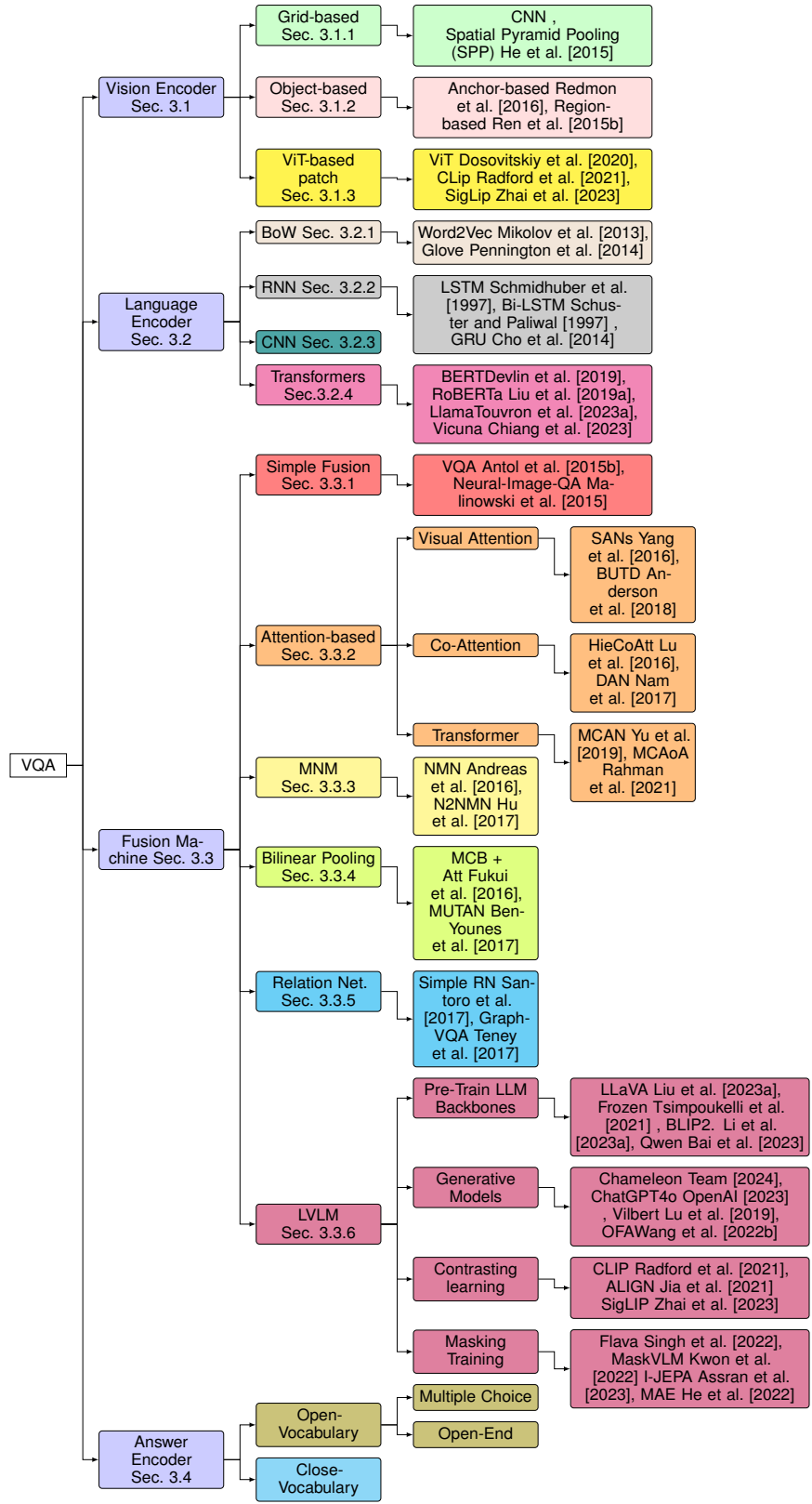

In the following subsections, we will cover the primary techniques utilized in each step involved in operating a VQA system.
These techniques are also summarized in the general taxonomy illustrated in Figure~\ref{fig:lit_surv}.

\section{VQA Architecture}
\label{sec:vqa_architecture}

VQA is a task that was first introduced by Antol et al. (2015) \cite{VQA:2014}, is a task in artificial intelligence that involves answering natural language questions about images. 
It requires a system to have a wide range of knowledge and the ability to reason about the content of the images to provide accurate answers. A VQA system generally consists of four main components: Vision Encoder, Language Encoder, Fusion Machine, and Answer Decoder (as shown in Fig \ref{fig:VQA_system}). In this section, Subsections \ref{sec:Vision_Encoder} and  \ref{sec:Language_Encoder} 
consist of the techniques used to extract image representations and question representations, respectively. 
Subsection \ref{sec:Fusion_Machine} includes the taxonomy of models used to combine image and question representations. 
Lastly, the answer decoder is discussed in Subsection \ref{sec:Answer_Decoder}.

The organization of this section follows the three primary steps involved in operating a VQA system as depicted in Figure~\ref{fig:VQA_system}: Image and Question Feature Extraction respectively in Sections~\ref{sec:Vision_Encoder} and  \ref{sec:Language_Encoder}, Fusion~\ref{sec:Fusion_Machine}, and Answer Decoder~\ref{sec:Answer_Decoder}.

\subsection{Vision Encoder}
\label{sec:Vision_Encoder}

The Vision Encoder is a critical component of a VQA system as it extracts important visual features from the input image and encodes them in a concise and meaningful way. 
The output of the Vision Encoder is then fed into the fusion module, where it is combined with the question embedding to generate a prediction for the answer to the given question.
Below, we will discuss three categories in which Vision Encoders can be classified based on how images are encoded.

\subsubsection{Grid-based encoder} 
\label{sec:Grid_based} 
% Visual encoder divides the input image into a grid of cells and processes each cell independently to produce a latent image representation.
% Pre-train CNN models by eliminating their last layer are usually used to extract the grid-based features.
In VQA systems, a Grid-based encoder refers to an Vision Encoder that partitions an input image into a grid of cells and processes each cell independently. 
The rationale behind this approach is that various cells of an image may contain unique visual information that is relevant to answering a given question. 
The visual features extracted from each cell are subsequently combined to  produce a latent  representation of the input image.

Convolution Neural Networks (CNNs)~\cite{6795724} are typical examples of Grid-based encoders.
In recent years, several models based on CNNs have been developed for image classification, typically pre-trained using large-scale image datasets such as ImageNet~\cite{5206848}. 
Among these models, VGGNet~\cite{DBLP:journals/corr/SimonyanZ14a}, ResNet~\cite{7780459}, and GoogleNet~\cite{7298594} are widely used as Vision Encoders due to their popularity and effectiveness.

On the other hand, Spatial Pyramid Pooling (SPP)~\cite{7005506} has been utilized in VQA to extract features from individual cells of an image. 
A spatial pyramid pooling layer is applied to combine the features from each cell at multiple scales. 
This allows the model to capture both local and global visual information from the image, which can be important for accurately answering a question~\cite{Jiang_2020_CVPR,banerjee-etal-2021-weaqa,YU2023119148}.
% In these models, SPP has been shown to improve the performance of VQA systems by allowing them to capture more fine-grained visual information from the input image.

Also, Spatial Transformer Networks (STNs)~\cite{NIPS2015_33ceb07b} have been used in VQA to dynamically transform input images to enhance their salient features before encoding. 
STNs allow the network to learn spatial transformations that can align image regions with corresponding question semantics, thus improving the overall performance of the VQA system~\cite{7965954}.

\subsubsection{Object-based}
\label{sec:Object_based}

The idea behind object-based VQA models is to capture the relationships between objects in an image and their relevance to a given question, as objects are often the primary focus of questions in VQA. 
By localizing objects and extracting their features, these models can better understand the visual context of a question and generate more accurate answers. 
Various techniques have been used in VQA for identifying and localizing objects in an image~\cite{Gordon_2018_CVPR,anderson2018bottom,guo2023from}.
Region-based methods use a region proposal network to generate a set of candidate object bounding boxes in an image, followed by a classifier to predict the category and location of each object within the proposed boxes. 
Faster-RCNN \cite{NIPS2015_14bfa6bb} is a commonly used region-based method to detect important objects in input images. 
Single-shot methods accomplish object detection and classification in one forward pass of a neural network. They commonly use a set of default bounding box shapes, known as anchors, to forecast the category and location offsets for each anchor. 
YOLO~\cite{Redmon_2016_CVPR} is an instance of a single-shot method. 
Anchor-based methods also use anchor boxes to predict object locations and categories, but they create anchor boxes at several scales and aspect ratios to better handle objects of different sizes and shapes. 
RetinaNet~\cite{Lin_2017_ICCV} is a type of anchor-based method.
A backbone CNN model such as VGGNet~\cite{DBLP:journals/corr/SimonyanZ14a} or ResNet~\cite{7780459} is used to extract visual features of the detected bounding box, which facilitate the identification of object attributes, quantity, and categories for accurate inferences~\cite{anderson2018bottom}.
It is worth noting that as VQA models that use an object-based Vision Encoder are two-stage models, they cannot be trained end-to-end.

% Because VQA models adopting an object-based Vision Encoder are two-stage models, they cannot be end-to-end trained.

\subsubsection{ViT-based Patch Encoder}
\label{sec:ViT-based_patch}
Inspired by the Vision Transformer (ViT) \cite{dosovitskiy2020image}, a \textit{ViT-based patch encoder} splits an image into small patches and linearly embeds them to create patch embeddings. To produce a single feature vector summarizing the entire input sequence, a learnable token is added at the beginning of the input sequence.

Some VQA models, such as ViLT~\cite{kim2021vilt} and VLMo~\cite{bao2021vlmo}, combine patch embeddings with question tokens and input them into a multi-layer Transformer block. However, other models like X-VLM~\cite{zeng2022multi} and X$^2$-VLM~\cite{zeng2022x} independently send patch embeddings into a multi-layer Transformer block to generate final output image features. This approach has shown promising results, allowing for more flexible and efficient processing of images in VQA tasks.

Recently, using encoders from pretrained models like CLIP \cite{radford2021learning}, ALIGN \cite{li2021align}, and SigCLIP \cite{zhai2023sigmoid} has gained traction because they offer robust and high-quality visual features that significantly enhance performance. These models align images with text effectively, making them highly compatible with natural language queries. Incorporating these encoders into VQA models improves accuracy and robustness, reduces the need for extensive task-specific training, and achieves state-of-the-art performance across various benchmarks, advancing capabilities in VQA.

In addition to the aforementioned models, some approaches adapt other types of computer vision models such as Segmentation \cite{Kirillov_2023_ICCV} and Depth Estimation \cite{birkl2023midas}, or combine different vision encoders together, as seen in recent trends \cite{lee2024moai, tong2024cambrian1}.

\subsection{Language Encoder}
\label{sec:Language_Encoder}
In a VQA system, a Language Encoder is responsible for transforming the original text of a question into a numerical latent representation. 
Usually, it utilizes natural language processing (NLP) techniques to examine the structure and significance of the question. 
Different techniques have been applied to obtain the question features as discussed below.

\subsubsection{Bag-of-Words (BoW)} 
\label{sec:BoW} 
is the most simple for question encoding. Firstly, a fixed vocabulary is created from the words from all questions. For each question, a feature vector is then produced by counting the frequency of each word in the vocabulary and representing it as a histogram. This method can not encode the dependencies between neighbor words.
    
\subsubsection{Recurrent Neural Network (RNN)}
\label{Sec:RNN}
is one of the most popular Language Encoders used in VQA. This method can capture text structures and word dependencies of input questions. The question is firstly processed by one-hot encoding and then passed into a word embedding layer that can be used with pre-trained models such as GloVE \cite{pennington-etal-2014-glove} or extracted from word2vec \cite{mikolov2013efficient}, or also can be trained from scratch during the training process. After that, these word embeddings are fed to an RNN, such as Long Short Time Memory (LSTM) \cite{schmidhuber1997long}, biLSTM, GRU \cite{DBLP:journals/corr/ChoMGBSB14}, and Skip-thought Vectors \cite{kiros2015skip} to extract a contextual representation of the questions.
    
\subsubsection{Convolutional Neural Network (CNN)} 
\label{sec:cnn} 
can be used to encode the question representations by using convolutional filters to extract features from local regions of the input question \cite{Yang_2016_CVPR}. The questions are also firstly processed by one-hot encoding and then passed into a word embedding layer. These word embeddings are then fed into a 2D matrix in which each row represents a word. Consequently, the 2D matrix is passed through several convolutional layers and max-pooling layers to extract the question representations.

\subsubsection{Transformers} 
\label{sec:Transformers} 
\cite{vaswani2017attention} It has been the most popular in NLP tasks, and unsurprisingly, Transformer Networks have become a popular method for extracting question representations in VQA. Following BERT \cite{devlin2018bert}, and RoBERTa \cite{liu2019roberta}, the input question is broken into smaller subwords or tokens using WordPiece tokenization. The tokenized input is then mapped to high-dimensional vector embeddings, which capture the semantic meaning of each token in the context of the surrounding text. A special learnable token [CLS] is inserted at the beginning of the embedding to summarize the entire input sequence during training. On the one hand, these word embeddings can be concatenated with the visual token and then directly fed into a multi-layer Transformer block that is initialized with BERT or RoBERTa as ViLT \cite{kim2021vilt}, and VLMo \cite{bao2021vlmo}. On the other hand, they are sent into a Language Model like BERT, RoBERTa, mT5 \cite{xue2020mt5}, or Llama family \cite{touvron2023llama, touvron2023llama2} to capture the language features first and then interact with the visual features.

\subsection{Multimodal Fusion}
\label{sec:Fusion_Machine}

The Fusion machine, the critical element of a VQA system, aims to take the image representations from the Vision Encoder and question representation from the question decoder as inputs and provides the fused multi-modal feature as an output. This section will provide a taxonomy of fusion techniques ranging from simple operations such as concatenation or element-wise multiplication to complex attention machines. We categorize fusion techniques into five categories: Simple Fusion, Attention, Bilinear Pooling, Neural Module Networks, and Visual-Language Pre-train Models.

\subsubsection{Simple Fusion} 
\label{Simple Fusion}
In the first previous works, image and question features are fused using simple operations such as element-wise multiplica tion, sum, or concatenation \cite{VQA:2014, jabri2016revisiting, saito2017dualnet}. Apart from that, then some slightly more complex methods are used, such as CNN \cite{ma2016learning}, LSTM \cite{malinowski2015ask, ren2015exploring, gao2015you}, and Bayesian Model \cite{kafle2016answer}. However, these simple fusion techniques obtain lower performances for some reasons. Firstly, they cannot capture the complex relationships between the visual and textual inputs necessary for accurate VQA. Secondly, simple fusion techniques need to provide insight into how the visual and textual inputs are combined, making it difficult to understand how the VQA system arrives at its answers.

\subsubsection{Attention-based Methods}
\label{Attention}

Attention Machine, which was first introduced by Bahdanau et al. (2014) \cite{bahdanau2014neural}, is mainly used to align and translate source language input to target language output in translation machines. In other words, the attention machine aims to focus on certain parts of the input data rather than considering the entire input at once. In VQA, Attention Machine aims to selectively focus on the most relevant region features in grid-based models or objects in object-based models in an image and a text input question when generating answers. More specifically, VQA needs to learn the image's critical region part or objects and the essential words. For instance, we assume that an image includes a yellow dog on the right side and another black dog is the left side with the given question \textit{"what color of the left dog"}. For the image, the responsibility of the attention machine is to allow VQA to pay more attention to the left dog than the right dog or other parts of the image. For the question, VQA needs to focus on the keywords including \textit{"color"}, \textit{"left"}, and \textit{"dog"}. Compared to simple fusion, attention-based models reach a higher performance.

There are several methods to use Attention Machine in VQA:

\subfour{Visual-Attention:} Guided attention refers to the use of an attention mechanism to help the model focus on specific parts of an image when generating an answer to a question about the image. The attention mechanism is guided by the question, which is used as a query to direct the model's attention to relevant regions of the image.  Yang et al. (2016) \cite{Yang_2016_CVPR} proposed a stack attention networks (as known as SANs) that stacks the attention networks using the softmax function  for multi-step reasoning to select a narrow region in visual information according to the given question. After that, Residual Networks is applied into SANs to improve the performance \cite{kim2016multimodal}. Inspired by End-to-End Memory Networks \cite{sukhbaatar2015end}, Xu and Saenko \cite{xu2016ask} proposed a Spatial Memory Network (SMen-VQA) that used spatial attention to compute the correlation of image representations with single words in the question in the first hop. In the second hope, the visual and question are combined and then looped into the attention machine again to refine the attention distribution. Similarly, Xiong et al. (2016) \cite{pmlr-v48-xiong16} propsed DMN+ based on a Dynamic Memory Network \cite{kumar2016ask} from Textual-QA. Instead of the uniform grid features, another approach to implement attention  is to generate all bounding boxes in the image and then use question-guided visual attention to determine the relevant boxes \cite{shih2016look, ilievski2016focused, anderson2018bottom, teney2018tips, ijcai2018p126}.

\subfour{Co-Attention:} Co-attention mechanism allows the model to attend to both the visual and textual input simultaneously. In other words, this mechanism uses the visual representations to guide the question attention, and uses the question representations to guide the visual attention. Lu et al. (2016) \cite{lu2016hierarchical} proposed the first co-attention mechanism named HieCoAtt for VQA to produce attention on the image feature and the question simultaneously. After that, Kim et al (2018) \cite{kim2018bilinear} improved the previous model by publishing the Bilinear Attention Networks (BAN); that is created by applying co-attention into bilinear attention which considers every pair of question words and image regions. Nguyen et al. (2018) \cite{nguyen2018improved} also said that HieCoatt considers only visual attention from a full sentences and language attention on a entire image leading the performance limitation. Therefore, they proposed a dense co-attention network (DCN) that stacks multiple dense co-attention layers to fuse the language and visual representation repeatly. Besides, the dense co-attention layers is dense that consider every interaction between any word and any region.
  %  Additionally, the proposed model is a hierarchical architecture which co-attends the question and the image in three different levels: word level, phrase level, and question level. This approach makes the co-attention more meaningful, hence, improves the overall performance of Image QA.

\subfour{Transformer:} Although the above co-attention models present little better performance, the bottleneck is their obstacle that limits the self-attention ability within each modality simultaneously, such as word-to-word relationships in a question and region-to-region relationships in an image. Therefore, Inspired by Transformers \cite{vaswani2017attention}, Yu et al. (2019) \cite{yu2019deep} introduced MCAN that used the self-attention unit from Transformers to obtain the word-to-word relationships and the region-to-region relationships. In addition, the authors also use the guided-attention unit to get the cross-attentions, such as word-to-region relationships. MCAN adopts two architectures (stack and encoder-decoder) consisting of multiple MCA layers cascaded in-depth to refine the attended image and question features gradually. The result presented that the encoder-decoder MCAN architecture outperforms the other architecture and the existing models. After that,   Rahman et al. (2021) \cite{rahman2021improved} proposed MCAoAN that extends MCAN by using Attention-on-Attention \cite{huang2019attention}. Zhou et al. (2021) \cite{zhou2021trar} then introduced the first example-dependent routing scheme for Transformer to schedule global and local attention in VQA named TRAR dynamically. These models proved that Transformers is the better approach to fuse the language and question future than other attention-based methods. Steitz et al. (2022) \cite{steitz2022txt} proposed an end-to-end network named TxT based on the transformer-based object detector called DETR. The model shows a clear improvement in VQA accuracy compared to the Fast R-CNN features.

\subsubsection{Neural Module Networks (NMNs)} 
\label{sec:nmns}
Apart from attention-based models, another method to improve the performance of VQA is Neural Module Networks (NMNs). Particularly, the input question in VQA may require multiple steps of reasoning to answer correctly. For example, the question "What is the object on the refrigerator?" requires some steps: the first is finding the refrigerator and then identifying the object on the table. NMNs are specially designed to handle the compositional structure of questions.

The architecture of NMNs consists of a set of shallow neural networks called "module". Each neural network is designed to perform a specific function that performs a single well-defined task. These modules are combined and connected in different ways to form an instance-specific network for each input question. Andreas et al. \cite{andreas2016neural} use a natural language parser to find the question's sub-tasks and infer the network of the sub-tasks that, when executed in sequence, would produce an answer to the given question. This architecture cannot be trained end-to-end because it involves two stages: program synthesis and program execution.
Consequently, IPE \cite{johnson2017inferring} and N2NMNs \cite{hu2017learning} are proposed to address the limitation of the previous works by making end-to-end trainable NMNs via reinforcement learning. Later, Stack-NMM \cite{hu2018explainable} makes a soft layout selection so that the whole model is fully differentiable. Finally, neural-Symbolic VQA models \cite{yi2018neural, vedantam2019probabilistic} perform symbolic reasoning by encoding images into the scene graph.

\subsubsection{Bilinear Pooling Methods} 
\label{sec:bilinear_pooling}
VQA strongly relies on the way to join the image and question features. Early models use simple operations such as concatenation and element-wise product between the question and image features. However, these operations need to be more expensive to capture the complex relationships between image and question features. Bilinear Pooling is a straightforward way to create more complex interactions of inputs.

Firstly, Fuikui et al. (2016) \cite{fukui2016multimodal} proposed Multimodal Compact Bilinear (MCB) pooling to compact bilinear models. However, using the outer product on the original question and image features would be very highly dimensional, leading to an infeasible number of parameters, so they use Count Sketch Function \cite{charikar2002finding} and Fast Fourier Transformation (FFT) to produce a lower dimensional space to avoid computing the outer product directly. However, this method is still computationally expensive because MCB uses an approximate outer product. Therefore, multimodal low-rank bilinear Pooling (MLB) \cite{kim2016hadamard}  use the Hadamard product to produce a low-rank bilinear pooling. Nevertheless, Yu et al. \cite{yu2017multi} pointed out that even though MLB delivers low dimensional features, has fewer parameters, and reaches a higher performance than MCB, MLB has a slow convergence rate and is sensitive to the learned hyper-parameters. Therefore, Multimodal Factorized Bilinear pooling (MFB) was created to overcome the limitation of MCB and MLB. Firstly, the visual and language features are expanded to a higher dimensional space. They are then compressed into a dense output feature using a sum pooling layer. Other more advanced bilinear pooling methods consist of MUTAN \cite{ben2017mutan}, Film \cite{perez2018film}, and BLOCK \cite{ben2019block}.

\subsubsection{Relation Networks}
\label{sec:Relation_Networks}
Relation reasoning refers to the ability of a machine learning model to understand and reason relationships between objects or entities. This involves identifying patterns and connections between objects and using those patterns to make predictions or decisions. The relationship between objects may be a simple binary relationship (e.g., Are these two objects the same color?) or a more complex multi-class relationship (e.g., "What is the relationship between these two objects?"). Relation Networks aim to learn and reason about relationships between objects.

Santoro et al. (2017) \cite{santoro2017simple} consider the relationships between image regions in image conditioning on the question context to build a universal pairwise relation network. Because they use a CNN to extract the image features, the object could be the background or a physical object. Although this architecture improves the VQA performance, this obtains only extracts superficial relationships. Teney et al. (2017) \cite{teney2017graph} used the Stanford dependency parser to obtain graph representations of questions. They then proposed a graph-based approach utilizing graph neural networks to combine abstract images and graph representations of questions. Their model achieved high performance but used scene graph representation of abstract images. Therefore, it takes work to apply it to real-world images.

% \subsubsection{Visual-Language Pre-train} 
\subsubsection{Large Visual-Language Models}
\label{LVLM}

Large language models (LLMs) have revolutionized AI by leveraging vast datasets to learn complex representations, significantly improving performance on various tasks. Building on the success of LLMs in NLP likes BERT, researchers have developed Visual-Language Pre-training (LVLM) models since 2019. These models are trained on large-scale visual and linguistic datasets to capture cross-modal relationships, enhancing performance in tasks such as image captioning and VQA. LVLM models are categorized into four training methods: Contrasting Training, Masking Training, Pretrained Backbones, and Generative Models.

% \subfour{Contrasting Training:} Contrasting learning \cite{lecun2006tutorial} is a paradigm of machine learning that unlabeled data point are compared to teach a model which point are similar and which are different. In the other words, samples belonging to the same distributions are pull closer in the embedding, while those from different distribution are push away. This method can leverages large amounts of unlabeled data, bypassing the need for extensive labeled datasets required in supervised learning.

\subfour{Contrasting learning} \cite{lecun2006tutorial} is a paradigm of machine learning where unlabeled data points are compared to teach a model which points are similar and which are different. In other words, samples belonging to the same distribution are pulled closer in the embedding space, while those from different distributions are pushed away. This approach can utilize vast amounts of unlabeled data, eliminating the need for the extensive labeled datasets typically required in supervised learning.

% Contrastive Language–Image Pre-training (CLIP) \cite{radford2021learning} is a notable contrastive learning model that maps images and their corresponding caption to similar vector, otherwise, unrelated ones are distant. This capability enables the model to understand and retrive images based textual queries and vice versa. However, contrastive leaning is aligns images and text in a shared embedding space, it lacks the fine-grained understanding and explicit answer prediction mechanisms required for complex reasoning tasks as VQA. Consequently, CLIP is often used to extract rich image features that can be fed into other models specifically designed for tasks like VQA, leveraging its strong visual representations while addressing its limitations. Other kind of this method are , ALIGN \cite{jia2021scaling}, SigLIP \cite{zhai2023sigmoid}, Llip \cite{lavoie2024modeling}.
Contrasting learning \cite{lecun2006tutorial} Contrastive Language–Image Pre-training (CLIP) \cite{radford2021learning} is a notable contrastive learning model that maps images and their corresponding captions to similar vectors, while unrelated ones are mapped to distant vectors.
This capability enables the model to understand and retrieve images based on textual queries and vice versa.
However, while contrastive learning aligns images and text in a shared embedding space, it lacks the fine-grained understanding and explicit answer prediction mechanisms required for complex reasoning tasks such as VQA.
Consequently, CLIP is often used to extract rich image features that can be fed into other models specifically designed for tasks like VQA, leveraging its strong visual representations while addressing its limitations.
Other similar methods include ALIGN \cite{jia2021scaling}, SLIP \cite{mu2022slip}, SigLIP \cite{zhai2023sigmoid}, and Llip \cite{lavoie2024modeling}.

\subfour{Masking Training:} Inspired by BERT's success with Masked Language Modeling (MLM) \cite{devlin2018bert}, a similar technique is used in Vision-Language Models (LVLMs). Mask Training involves masking parts of images and sentences during training, akin to how MLM masks words in text. Image patches are randomly dropped, compelling the model to predict the missing visual information from the surrounding context, while masked words in text are predicted based on surrounding words. This approach enhances the model's ability to generate coherent representations from incomplete inputs, improving its contextual understanding and robustness \cite{singh2022flava, kwon2022masked, assran2023self, he2022masked}.

\subfour{Generative model:}
Generative LVLMs have demonstrated significant advancements in VQA by leveraging their ability to generate both captions and images.
Initially, models focused on text-based tasks using a complete encoder-decoder architecture to produce descriptive captions \cite{lu2019vilbert,tan2019lxmert,yu2022coca}.
However, recent developments have expanded their capabilities to include image generation, as exemplified by models like Stable Diffusion \cite{sauer2024fast, Stable_Diffusion_3}.
This multimodal proficiency allows models like ChatGPT \cite{openai2023gpt4} and Chameleon \cite{chameleonteam2024chameleonmixedmodalearlyfusionfoundation} to excel in VQA tasks, where understanding and generating both visual and textual information are crucial.
These advancements enable more accurate and contextually relevant answers to visual queries, enhancing the overall effectiveness of VQA systems.

\subfour{Pretrained Backbones:}
Training from scratch involves significant costs, including the need for vast amounts of vision-language data and the computational power of hundreds to thousands of GPUs. However, many LLMs are open-source and more accessible, leading to a shift towards fine-tuning existing LLMs to become LVLMs through the use of adapters that map visual tokens to language tokens. These adapters serve as a bridge, enabling the model to effectively process and understand both image and language modalities. Open-source LLMs such as Llama, Qwen, and Falcon can be utilized as backbones for this purpose \cite{llama3, bai2023qwentechnicalreport, falcon2}. Adapters can take various forms, including simple Multilayer Perceptrons (MLPs) \cite{tsimpoukelli2021multimodalfewshotlearningfrozen, liu2023llava, liu2024llavanext, zhu2023minigpt}, Perceiver Resamplers \cite{alayrac2022flamingo}, and Q-Formers \cite{li2023blip2bootstrappinglanguageimagepretraining}.
This trend towards using adapters for fine-tuning presents several advantages. Fine-tuning pre-trained models requires significantly less computational power and data than training from scratch, making it more accessible and cost-effective.

\subsection{Answer Decoder} 
\label{sec:Answer_Decoder}
% Depending on the question styles, different designs of an answer decoder are required to deliver answer predictions. There are two popular representative tasks in VQA: Open-End Question Answering and Multiple-choice Question Answering.
VQA requires specialized answer decoders to handle different types of questions. Depending on whether the task is open-vocabulary or closed-vocabulary, different designs and strategies are employed:

\subfour{Close-Vocabulary Decoder}
Closed-vocabulary question answering restricts the possible answers to a predefined set. This makes it easier for the model to select the correct answer from a limited pool. There are two main types of tasks within closed-vocabulary VQA:

\textit{1. Open-End Question Answering:} Despite traditionally referring to free-form text responses, in a closed-vocabulary context, open-end question answering involves selecting the correct answer from a predefined set of responses. For instance, if the task is to determine what a man in an image is doing, the model chooses from predefined answers like \textit{``running"}, \textit{``walking"}, and \textit{``sleeping"}. The answer decoder uses several fully connected layers. A softmax function computes the probabilities for each predefined answer, and the answer with the highest probability is selected as the final output.

\textit{2. Multiple-Choice Question Answering:} In multiple-choice question answering, the model is presented with a question and a fixed set of possible answers. The task is to select the correct answer from these options. For example: \textit{``What is the man in the image doing?" (1) Running (2) Walking (3) Sleeping (4) Swimming}. The decoder ranks the similarity between each answer choice and the question-context pair. A fully connected layer scores each option, and a softmax function identifies the most likely correct answer by picking the one with the highest score.

\subfour{Open-Vocabulary Decoder}
Open-vocabulary VQA is a challenging task that involves generating free-form text responses to natural language questions about images, allowing for more flexible and natural responses compared to closed-set VQA systems with predefined answer options. The answer decoder, a crucial component in open-vocabulary VQA, typically leverages advanced language models based on transformer architectures like Vicuna \cite{vicuna2023}. These models excel at understanding context and generating human-like text, making them well-suited for producing diverse and accurate answers. By employing these sophisticated language models, open-vocabulary VQA systems can generate coherent and relevant responses that adapt to a wide range of questions and image content, pushing the boundaries of artificial intelligence in visual understanding and natural language processing.

\section{Datasets} 
\label{datasets}
VQA is a challenging and interdisciplinary task that lies at the intersection of computer vision and natural language processing. VQA datasets play a crucial role in advancing the state-of-the-art in this field by providing diverse and annotated data for training and evaluating VQA models. In this section, we will review some of the existing and notable VQA datasets.

\begin{table*}[ht]
\centering
\rowcolors{2}{gray!20}{white}
\captionsetup{justification=centering}
\caption{Comparison of popular VQA datasets. Notes: VIP - Visual Impaired People; AG - Answer Grounding. \\ * - Benchmark dataset without training set. }
\resizebox{0.9\linewidth}{!}{
\begin{tabular}{|c |l |c |c|c |c |c |} 
\hline
\rowcolor{gray!50}
\textbf{ID}&\textbf{Name} & \textbf{Year} & \textbf{Domain} &\textbf{Images} & \textbf{Questions} & \textbf{Type}\\
\hline

1&\href{https://arxiv.org/abs/1505.00468}{VQA v1.0} \cite{Antol_2015_ICCV} & 2015 & General &204,721 & 614,163  & OE \& MC\\
2&\href{https://arxiv.org/abs/1505.00468}{VQA v1.0} \cite{Antol_2015_ICCV}& 2015 & Abstract sense  &50,000 & 150,000 & OE \& MC\\
3&\href{https://arxiv.org/abs/1505.02074}{COCO-QA} \cite{ren2015exploring}& 2015 & General &123,287 & 117,684 & OE\\
4&\href{https://arxiv.org/abs/1511.05099}{Binary-VQA} \cite{Zhang_2016_CVPR}& 2015 & Abstract sense & 50,000 & 150,000 & MC\\
5&\href{https://proceedings.neurips.cc/paper/2015/hash/fb508ef074ee78a0e58c68be06d8a2eb-Abstract.html}{FM-IQA} \cite{gao2015you}& 2015 & General & 158,392 & 316,193 & OE\\
6&\href{https://arxiv.org/pdf/1511.02570.pdf}{KB-VQA} \cite{wang2015explicit}& 2015 & KB-VQA & 700 & 2,402 & OE\\
7&\href{https://arxiv.org/abs/1602.07332}{VG} \cite{krishna2017visual}& 2016 & General & 108,077 & 1,700,000  & OE\\
8&\href{https://arxiv.org/abs/1511.02799}{SHAPE} \cite{andreas2016neural}& 2016 & Abstract shapes & 15,616 & 244 & MC\\
9&\href{https://www.semanticscholar.org/paper/A-Dataset-for-Multimodal-Question-Answering-in-the-Sheng-Gool/10be82098017fc2d60b0572cea8032afabad5d1a}{Art-VQA} \cite{sheng2016dataset}& 2016 & Cultural Heritage & 16 & 805 & OE\\
10&\href{https://arxiv.org/pdf/1603.07396}{AI2D} \cite{sheng2016dataset}& 2016 & Textbook & 5000 & 15,000 & MC\\
11&\href{https://arxiv.org/abs/1606.05433}{FVQA}  \cite{wang2017fvqa}  & 2017 & KB-VQA & 1,906 & 4,608 & OE\\
12&\href{https://arxiv.org/abs/1410.0210v4}{DAQUAR}  \cite{malinowski2014multi}  & 2017 & General & 1,449 & 12,468 & OE\\
13&\href{https://arxiv.org/abs/1511.03416}{Visual7W} \cite{zhu2016visual7w} & 2017 & General &47,300 & 327,939 &  MC\\
14&\href{https://arxiv.org/abs/1612.00837}{VQA v2.0} \cite{goyal2017making} & 2017 & General &200,000 & 1,100,000 & OE \& MC\\

15&\href{https://arxiv.org/abs/1612.06890}{CLEVR} \cite{johnson2017clevr}& 2017 & 3D shapes &100,000 & 853,554 & OE\\
16&\href{https://arxiv.org/abs/1712.00377}{VQA-CP1} \cite{agrawal2018don}& 2017 & General & 205,000 & 370,000 & OE\\
17&\href{https://arxiv.org/abs/1712.00377}{VQA-CP2} \cite{agrawal2018don}& 2017 & General & 219,000 & 658,000 & OE\\
18&\href{https://openaccess.thecvf.com/content_cvpr_2017/papers/Hussain_Automatic_Understanding_of_CVPR_2017_paper.pdf}{AD-VQA} \cite{hussain2017automatic}& 2017 &  Advertisement & 64,832 & 202,090 & OE\\
19&\href{https://openaccess.thecvf.com/content_cvpr_2017/papers/Kembhavi_Are_You_Smarter_CVPR_2017_paper.pdf}{TQA} \cite{kembhavi2017you}& 2017 & Textbook & 1,496 &  	6,501 & MC\\
20&\href{https://arxiv.org/abs/1710.07300}{FigureQA} \cite{kahou2017figureqa}& 2018 & Chart & 125,000 & 1,550,000 & OE\\
21&\href{https://ceur-ws.org/Vol-2125/paper_212.pdf}{VQA-MED-2018} \cite{hasan2018overview}& 2018 & Medical & 2,866 & 6,413 & OE\\
22&\href{https://www.nature.com/articles/sdata2018251}{VQA-RAD} \cite{Lau2018ADO}& 2018 & Medical & 315 & 3,515 & OE \\
23&\href{https://vizwiz.org/tasks-and-datasets/vqa/}{VizWiz} \cite{8578478} & 2018 & VIP & 32,842 & 32,842& OE\\
24&\href{https://ojs.aaai.org/index.php/AAAI/article/view/4815}{TallyQA} \cite{acharya2019tallyqa} & 2018 & General & 	98,680 & 183,986& OE\\
25&\href{https://arxiv.org/abs/1801.08163v2}{DVQA} \cite{kafle2018dvqa} & 2018 & Chart & 	300,000 & 3,487,194& OE\\
26&\href{https://www.imageclef.org/2019/medical/vqa}{VQA-MED-2019} \cite{abacha2019vqa}& 2019 &  Medical & 4,200 & 15,292 & OE\\
27&\href{https://arxiv.org/pdf/1904.08920v2.pdf}{TextVQA} \cite{singh2019towards}& 2019 & Text-VQA &28,408 & 45,336 & OE\\
28&\href{https://ieeexplore.ieee.org/document/8978122}{OCR-VQA} \cite{8978122}& 2019 & Text-VQA &207,572 & 1,002,146 & OE\\
29&\href{https://arxiv.org/pdf/2002.10215.pdf}{STE-VQA} \cite{wang2020general}& 2019 & Text-VQA & 21,047 & 23,887 & OE\\
30&\href{https://arxiv.org/abs/1905.13648}{ST-VQA} \cite{biten2019scene}& 2019 & Text-VQA & 22,020 & 30,471 & OE\\
31&\href{https://arxiv.org/pdf/1906.00067v2.pdf}{OK-VQA} \cite{marino2019ok}& 2019 & KB-VQA & 14,031 & 14,055 & OE\\
32&\href{https://arxiv.org/abs/1902.09506}{GQA} \cite{hudson2019gqa}& 2019 & General &113,000 & 22,000,000 & OE\\
33&\href{https://arxiv.org/pdf/1907.12861v1.pdf}{LEAF-QA} \cite{chaudhry2020leaf}& 2019 & FigureQA &240,000 & 2,000,000& OE\\
34&\href{https://openaccess.thecvf.com/content_CVPR_2019/html/Zellers_From_Recognition_to_Cognition_Visual_Commonsense_Reasoning_CVPR_2019_paper.html}{CRV} \cite{Zellers_2019_CVPR} & 2019 & Movie Scenes &110,000 & 290,000 & OE \& MC\\
35&\href{https://arxiv.org/abs/2007.00398}{DOC-VQA} \cite{mathew2021docvqa}& 2020 & Text-VQA&12,767 & 50,000 & OE\\
36&\href{https://arxiv.org/abs/2008.12520}{AQUA} \cite{garcia2020dataset}& 2020 & Cultural Heritage & 21,383 & 32,345 & OE\\
37&\href{https://arxiv.org/abs/2003.07333}{RSVQA-low} \cite{8578478} & 2020 & Remote Sensor  & 772 & 77,232 & OE\\
38&\href{https://arxiv.org/abs/2003.07333}{RSVQA-high} \cite{8578478} & 2020 & Remote Sensor & 10,659 & 1,066,316 & OE\\
39&\href{https://ceur-ws.org/Vol-2696/paper_106.pdf}{VQA-MED-2020} \cite{abacha2020overview}& 2020 &  Medical & 5,000 & 5,000 & OE\\
40&\href{https://aclanthology.org/2020.bionlp-1.6.pdf}{RadVisDial} \cite{kovaleva2020towards}& 2020 & Medical & 91,060 & 455,300& OE\\
41&\href{https://arxiv.org/abs/2003.10286}{PathVQA} \cite{he2020pathvqa}& 2020 & Medical& 4,998 & 32,799 & OE\\
42&\href{https://arxiv.org/abs/1909.00997}{PlotQA} \cite{Methani_2020_WACV}& 2020 & Chart& 224,377 & 28,952,641 & OE\\
43&\href{https://arxiv.org/abs/2005.04790}{HatefulMemes} \cite{kiela2020hateful}& 2020 & General& 	8,500 & 	8,500 & MC\\
44&\href{https://arodes.hes-so.ch/record/9062}{VQA-MED-2021} \cite{ImageCLEF-VQA-Med2021}& 2021 & Medical &  5,500 &  5,500 & OE\\
45&\href{https://arxiv.org/abs/2102.09542}{SLAKE} \cite{liu2021slake}& 2021 & Medical & 642 & 14,000 & OE\\
46&\href{https://arxiv.org/abs/2105.14517}{GeoQA} \cite{chen2021geoqa}& 2021 & Geometry Problems & 5,010 & 5,010 & MC\\
47&\href{https://arxiv.org/abs/2101.11272}{VisualMRC} \cite{tanaka2021visualmrc}& 2021 & Text-VQA & 10,197 & 30,562 & OE\\
48&\href{https://arxiv.org/pdf/2110.13214}{IconQA} \cite{lu2021iconqa}& 2021 & Logic & 96,817 & 107,439  & OE \& MC\\
49&\href{https://arxiv.org/abs/2208.05358}{CLEVR-MATH } \cite{lindstrom2022clevr}& 2022 &   3D shapes & 	70,000 & 788,650 & OE\\
50&\href{https://arxiv.org/pdf/2209.09513}{ScienceQA} \cite{lu2022learn}& 2022 & Textbook  & 10,332 & 21,208 & MC\\
51&\href{https://arxiv.org/abs/2202.01993}{ VizWiz-Grounding} \cite{chen2022grounding}& 2022 & VIP + AG & 9,998 & 9,998 & -\\
52&\href{https://aclanthology.org/2022.findings-acl.177/}{ChartQA-H} \cite{masry-etal-2022-chartqa}& 2022 & Chart & 4,804 &  9,608 & OE\\
53&\href{https://aclanthology.org/2022.findings-acl.177/}{ChartQA-M} \cite{masry-etal-2022-chartqa}& 2022 & Chart & 17,141 &  23,111 & OE\\
54&\href{https://ceur-ws.org/Vol-3357/invited1.pdf}{WSDM} \cite{TolokaWSDMCup2023}& 2023 & AG & 45,119 & 45,119 & -\\
55&\href{https://arxiv.org/pdf/2209.06794}{WebLI-VQA} \cite{650093}& 2023 & Website Images & - & 150,000,000 & OE \& MC\\
56&\href{https://arxiv.org/pdf/2307.06281}{MMBENCH*} \cite{liu2023mmbench}& 2023 & Mix & - & 3,217 & OE \& MC\\
57&\href{https://arxiv.org/abs/2307.16125}{SEED-Bench*} \cite{li2023seed}& 2023 & Mix & - & 24,000  & MC\\
58&\href{https://arxiv.org/abs/2308.02490}{MM-VET*} \cite{yu2023mm}& 2023 & Mix  & 200 & 218 & OE\\
59&\href{https://mmmu-benchmark.github.io/}{MMMU*} \cite{Yue_2024_CVPR}& 2024 & Academic  & 11,264 & 11,550 &OE \& MC \\
60&\href{https://arxiv.org/abs/2408.00765}{MM-VET-v2*} \cite{yu2024mm}& 2024 & Mix  & 469  & 517 & OE\\
61&\href{https://arxiv.org/abs/2310.02255}{MathVista*} \cite{lu2023mathvista}& 2024 & Mathematics  & 5,487  &  6,141 & MC\\

\hline
\end{tabular}
}
\end{table*}

\subsection{VQA datasets}
The VQA dataset is a large-scale dataset for the VQA tasks. It consists of images, associated questions, and answers and is used to train and evaluate VQA models. This dataset is one of the most popular for creating and validating VQA models. There is various versions of the VQA dataset have developed over the years.

\subfour{VQA v1.0~\cite{VQA:2014}}: This is the original VQA dataset that consists of 204,721 real-world images from the MS COCO dataset \cite{lin2014microsoft} with 614,163 questions. Additionally, This dataset also contains 50,000 abstract scenes with 150,000 questions. Each question has ten answers from 10 people with different confidence levels. VQA v1.0 has both open-end and multiple-choice questions for real-world and abstract images. However, a part of the questions was answered without looking at the images leading the language bias.

\subfour{VQA v2.0 \cite{goyal2017making}}: Because the VQA v1.0 dataset has a strong language bias, VQA v2.0 was proposed to address this problem. This dataset contains twice as many question-answer pairs as the old version, with 200K images and 1.1M question-answer pairs. Each image has three questions, and each question also has ten answers. VQA v2.0 has both open-end and multiple-choice questions. This version is now more used than the old version.

% \subsection{DAQUAR}
% DAQUAR \cite{malinowski2014multi} consists of indoor scenes with 1,449 images, each paired with five questions and corresponding answers. The questions in this dataset are based on spatial relationships and object recognition, and the answers are either binary or free-form. Because this is the first VQA dataset and its size is small, this dataset needs to train and evaluate VQA models effectively. Secondly, the images in DAQUAR \cite{malinowski2014multi} are just in-door scenes, and the visual context needs to be richer to build models. Finally, since the images were taken in-door with poor lighting conditions.

\subsection{Visual Genome}
Visual Genome \cite{krishna2017visual} is a large-scale visual knowledge base that provides information about the objects, attributes, and relationships in images. It is designed to be a resource for various computer vision and natural language processing tasks, such as image captioning, VQA, and image retrieval.

Visual Genome consists of 108,077 images, each of which has an average of 35 objects, 26 attributes, and 21 pairwise relationships between these objects. Questions must begin with one of six letters: \textit{what, where, how, why, who},  and \textit{when}. The questions must be clear, direct, precise, and only relevant to what is presented in the images. There are two types of question answering: free-form and region-based question answering. For freeform question answering, data labelers must write eight question-answer pairs about everything in images freely. On the other hand, for region-based question answering, the data labelers must write a question-answer pair according to a given object. Generally speaking, free-form question answering has more diverse question-answer pairs, but question-answer pairs in region-based question answering are more accurate.

\subsection{Visual7W}
Visual7W \cite{zhu2016visual7w} is a subset of the Visual Genome dataset by sampling 47,300 images from Visual Genome. Apart from six W question categories like Visual Genome, Visual7W is added another W question category \textit{which}. The fundamental difference between the two data sets is Visual7W is designed for the multiple-choice task, while Visual Genome is proposed for the open-end task. There are 327,939 question-answer pairs and 561,459 object groundings from 36,579 categories. Each question has four categories to choose from.

% \subsection{COCO-QA}
% COCO-QA \cite{ren2015exploring} is created from the MS-COCO dataset. The author used NLP algorithms to create automatic question-answer pair generators based on the description of the images. In the MS-COCO dataset, each image has five descriptions. Assuming that there is a description of "a dog is running". The generated question may be "what is the dog doing" and the generated answer may be "running". The dataset consists of 123,287 images and 117,684 questions that are split into a training set with 78,736 open-end questions and a test set with 38,948 questions. the categories of these questions are objects, color, location, and number. All answers are only one word that makes the evaluation of VQA models more accessible and more robust.

% Because the question-answer pairs are automatically generated, the data doesn't have absolute accuracy because of grammatical errors. In addition, the question richness could have been higher because it was made up of descriptive sentences of pictures and sentence-level repetition of questions.

\subsection{CLEVR}
The CLEVR (Compositional Language and Elementary Visual Reasoning) \cite{johnson2017clevr} dataset is a dataset that researchers at Stanford University created. The images in the dataset are rendered 3D scenes that depict simple objects and scenes, such as spheres and blocks of various colors, sizes, and shapes. The questions and answers are designed to test the ability of a VQA model to reason about visual concepts such as object properties, relationships, and logical operations. The dataset is designed to be challenging for current machine learning models and focuses on testing compositional reasoning, counting, and arithmetic. The whole dataset contains 100,000 images and 853,554 unique questions. It is divided into three according to the ratio of 70\%: 15\%: 15\% for the purpose of training, evaluating, and testing.

\subsection{GQA}
GQA \cite{hudson2019gqa} is another VQA dataset created by researchers at Stanford University. This is a real-world dataset for visual reasoning and compositional answering. GQA consists of 113K real-world images and 22M questions. Each image comes with a dense scene graph representing the objects, attributes, and relations it contains. GQA is the first dataset to provide "programs" which are logical representations of the question, providing a way to decompose questions into simpler sub-questions, making it more suitable to evaluate the compositional reasoning abilities of the model. In addition, each answer is augmented with textual and visual justifications, pointing to the relevant region within the image. Overall, GQA is a widely used benchmark dataset for evaluating the performance of VQA models due to the diversity and complexity of the questions and images it contains and its large size.

\section{Question Relevance in VQA } \label{question relevance}
The common belief when gathering responses to visual questions is that a questions can be answered using the provided image~\cite{Antol_2015_ICCV, andreas2016neural, gao2015you, goyal2017making, johnson2017clevr, krishna2017visual, malinowski2014multi, ren2015exploring, wang2015explicit, wang2017fvqa}. 
However, in practice, not everyone asks questions directly related to the visual content~\cite{davis2020unanswerable}, especially early-age learners.
% Hence, the posed questions might remain unanswered as they do not correspond with the image. 
% For instance, if a photo displays a cat but the question asks, ``What color is the dog?" it is not applicable. 
In VQA v1.0~\cite{Antol_2015_ICCV}, Ray et al.~\cite{ray2016question} conducted a study where they randomly selected 10,793 question-image pairs from a pool of 1,500 unique images. Their findings revealed that 79\% of the questions were unrelated to the corresponding images.
Hence, a VQA system should avoid answering an irrelevant question to an image,  as doing so may lead to considerable confusion and a lack of trust.
The exploration of question relevance has been extensively explored in the literature, leading to the development of numerous methods and algorithms aimed at avoiding answering irrelevant questions. 
Notable contributions include works such as \cite{ray2016question, stengel2022did, bhattacharya2019does, rajpurkar2018know, van2023document, 8578478, toor2017question, chandrasekaran2018explanations, mashrur2023robust,mahendru2017promise}.
The SimpsonsVQA dataset relates to existing datasets as it includes \textit{``Relevant"} question and \textit{``Irrelevant"} questions.

\section{Metrics} \label{metrics}
The evaluation of models is very diverse, depending on the type of problem that these models solve. Several metrics are commonly used to evaluate the performance of VQA models:
\subsection{Accuracy}
Accuracy is a fundamental metric used to evaluate VQA models, with two main variants:

\subsubsection{Simple Accuracy}
Simple accuracy is the most common metric for evaluating multiple-choice VQA tasks. It is computed by calculating the ratio of correct answers to total answers:

\begin{equation}
    Accuracy = \frac{\text{\textit{Number of questions answered correctly}}}{\text{\textit{Total questions}}}
\end{equation}

While simple accuracy can be used for open-ended multiple tasks, it can lead to incorrect evaluations. For example, if the question is "What animals are shown in the images?", the label is "dogs," and the prediction is "dog," it could be considered correct, but the system might regard it as incorrect.

\subsubsection{Top-3 Accuracy}
To address the limitations of simple accuracy for open-ended VQA tasks, Top-3 accuracy was introduced by Antol et al. (2014) \cite{VQA:2014}. This metric takes into account that each question in the VQA dataset typically has ten answers from ten different people. An answer is considered 100\% accurate if at least three people provide the same answer as the prediction. The Top-3 accuracy is computed as:

\begin{equation}
    Accuracy = min (\frac{n}{\text{\textit{3}}}, 1)
\end{equation}

where n is the number of people who provide the same answer as the prediction.
However, Top-3 accuracy has a significant practical limitation: it necessitates a dataset with multiple human responses per question, which can be resource-intensive and time-consuming to create.

% While both simple accuracy and Top-3 accuracy offer valuable perspectives on model performance, it's crucial to weigh their respective advantages and drawbacks when assessing VQA models. The choice of metric should be guided by the specific requirements of the task and the available resources for dataset creation.
\subsection{WUP Set Score}

Wu-Palmer Similarity (WUP) is a measure of the semantic similarity between two words or concepts in the WordNet lexical database \cite{miller1995wordnet} that is a large lexical database of English that groups words into sets of synonyms called synsets and describes semantic relationships between them. WUP calculates the similarity between two concepts as the depth of the least common subsumer divided by the sum of the depths of the two concepts in the WordNet hierarchy. The WUP score is in the range [0, 1], with the larger value meaning the higher the similarity of two words. For example, the WUP score of (dog, puppy) is 0.9677. Meanwhile, the WUPS score of (tall, short) is 0.3102. Under this metric, the answer, which is more semantically similar to the label, would be penalized less. Based on WUP, the WUP Set score (WUPS) \cite{malinowski2014multi} is defined as follows:

\begin{equation}
\begin{aligned}
    {\mathop{\rm WUPS}\nolimits} \left( {A,T} \right) = \frac{1}{N}\sum\limits_{i = 1}^N \min   \Bigg\{&\prod\limits_{a \in {A^i}} \max \limits_{t \in {T^i}}  {\mathop{\rm WUP}\nolimits} (a,t),\,\,
    &\prod\limits_{t \in {T^i}} {\mathop {\max }\limits_{a \in {A^i}} {\mathop{\rm WUP}\nolimits} } (a,t) \Bigg\} 
\end{aligned}
\end{equation}

Where $A, T$ is the answers and ground-truth, $A^i$ and $T^i$ are the i-th answer and the i-th ground-truth, respectively.

However, the WUPS score is difficult to be used to evaluate VQA models because of some limitations. Firstly, although some pairs of answers and correct answers are completely different, their similarity index is relatively high. For example, the WUP score of (dog, cat) is  0.8667. Secondly, it is only used to evaluate the questions which need to be answered in one word.

\subsection{Human-Like Evaluation}
As some generative LVLM models increasingly produce more natural, free-form answers, traditional evaluation metrics like accuracy and multiple-choice questions often fall short in capturing the true quality of these responses.
For example, when asked, ``What is the person in the image likely feeling based on their expression?'', a traditional system might expect a specific word like ``happy'' or ``sad''.
However, an advanced model might generate a more nuanced answer such as, ``The person appears to be content, possibly enjoying the moment''.
In this case, a simple metric check could incorrectly mark the response as wrong if it does not exactly match the expected word, despite it being a more human-like and contextually appropriate answer.
To address this, recent approaches LLMs like GPT-4 to simulate human-like evaluation. These LLMs assess responses based on semantic similarity, relevance, and overall coherence, offering a more flexible and accurate evaluation that aligns with how a human might judge the answer's quality. For instance, MM-VET \cite{liu2023mmbench} utilizes GPT-4 to grade model outputs on a scale from 0 to 1, reflecting how well the response aligns with human-like reasoning across multiple dimensions such as knowledge, spatial awareness, and language generation.
Similarly, MMBench \cite{liu2023mmbench} uses a hierarchical evaluation framework where LLMs are employed to determine the correctness of responses by comparing them against structured multiple-choice options, ensuring a robust and human-like assessment process.

However, this method has its limitations, primarily the cost associated with setting up and maintaining API access to powerful LLMs.
This can be a significant barrier, especially for research teams or smaller organizations with limited budgets, making it a less accessible option compared to traditional, less expensive evaluation metrics.

\section{Applications} \label{applications}
Although VQA is a promising and active area of research, there are still relatively few real-world applications of the technology, as it is a relatively new and complex task. However, some examples of real-world applications of VQA include the following:

\subsection{VQA for Medical Application}

VQA has the potential to be a powerful tool in medical imaging, aiding healthcare professionals in clinical decision-making and disease diagnosis. Unlike many AI applications in medicine that are tailored for specific tasks, VQA is designed to interpret a wide range of questions and provide accurate answers, making it highly versatile. However, applying VQA in the medical field poses significant challenges, particularly in the need for large, specialized datasets. Creating these datasets is expensive, requiring experts to carefully annotate medical images such as X-rays and CT scans, formulate relevant questions, and provide accurate answers. These images often require segmentation techniques to enhance analysis, and the language used in questions must be specifically tailored to the medical context. As a result, existing medical VQA datasets tend to be smaller and more specialized than general-domain datasets. Notable datasets include VQA-MED \cite{hasan2018overview, abacha2019vqa, abacha2020overview}, VQA-RAD \cite{lau2018dataset}, and PathVQA \cite{he2020pathvqa}.

Research in medical VQA accelerated with the VQA-Med challenge in 2018, leading to the adoption of various methods, many inspired by general-domain models. Attention-based models like Stacked Attention Networks (SAN) \cite{zhan2020medical, do2021multiple} and Bilinear Attention Networks (BAN) \cite{nguyen2019overcoming, liu2021contrastive} are commonly used, along with multimodal pooling techniques that improve model performance through advanced operations such as Multimodal Factorized Bilinear Pooling (MFB) \cite{peng2018umass}.

\subsection{VQA for Visually Impaired People}

Visually impaired individuals face significant challenges, such as limited access to visual information, which impacts their ability to navigate and interact with their environment. Traditional deep learning tools like object detection and image captioning offer some help but often lack the depth needed for a full understanding. VQA provides a more comprehensive solution by allowing users to ask questions about images and receive detailed answers in natural language.

The VizWiz dataset, introduced by Gurari et al. (2018) \cite{gurari2018vizwiz}, was pivotal in advancing VQA for the visually impaired. It includes 31,000 images taken by visually impaired users and their corresponding questions, making it more challenging due to the low quality and complexity of real-world images. VizWiz-Priv, a variation of this dataset, includes privacy-sensitive information that has been anonymized for safety. Additionally, answer-grounding datasets like VizWiz-VQA-Grounding \cite{chen2022grounding} enhance the relevance of answers by linking them to specific visual regions. Unlike medical VQA, where specialized models are often required, general-domain models like LXMERT \cite{tan2019lxmert} and OSCAR \cite{li2020oscar} are commonly adapted to assist the visually impaired.

 \subsection{VQA for Remote Sensing Data (RSVQA)}
% RSVQA was first explored by Lobry et al. (2019)  \cite{lobry2019visual} when they applied VQA to extract information from satellite images of the land.
% The questions are used to ask about aspects of the environment, such as the number of houses in the photo, the shape of the areas, and the size. They used a simple VQA using LSTM and ResNet followed by simple point-wise multiplication. Although the results of their work are high, with 80\% accuracy, their data set has some limitations. The diversity of questions could be higher, and the data set needs to be balanced. Therefore, the authors extended their work by introducing larger data containing very high-resolution images \cite{lobry2020rsvqa}. After that, some general-domain architectures including Multiple Pooling \cite{chappuis2021find}, LVLM \cite{felix2021cross, bazi2022bi, chappuis2022prompt}, Attention-based \cite{zheng2021mutual}, Relation Networks \cite{zhang2023spatial} are used for RSVQA.
RSVQA was first explored by Lobry et al. (2019) \cite{lobry2019visual} to extract information from satellite images, answering questions about environmental features like the number of houses or area shapes. They used a simple VQA model combining LSTM and ResNet with point-wise multiplication, achieving 80\% accuracy. However, their dataset had limitations, such as low question diversity and imbalance. To improve this, they later introduced a larger dataset with very high-resolution images \cite{lobry2020rsvqa}. Subsequently, other models from general VQA were adapted for RSVQA, including Multiple Pooling methods \cite{chappuis2021find}, LVLMs \cite{felix2021cross, bazi2022bi}, attention-based models \cite{zheng2021mutual}, and Relation Networks \cite{zhang2023spatial}.

 \subsection{VQA for Cultural Heritage}
Sheng et al. (2016) \cite{sheng2016dataset} introduced the first VQA dataset for Cultural Heritage, focused on the ancient Egyptian Amarna period. This small dataset includes 16 artworks and related questions, using a bottom-up model with Faster R-CNN. Garcia et al. (2020) \cite{garcia2020dataset} expanded this work with the AQUA dataset, which includes paintings and detailed commentary. They developed the VIKING model, which integrates visual and external knowledge, though the generated questions were often simple and lacked variety. To overcome the need for expert-generated descriptions, Bongini \cite{brown2020language} used GPT-3 to automatically generate detailed descriptions of artworks.

\subsection{VQA for Advertisement}
 % Advertisements play a crucial role in business because they attract customers' attention to products or services. By using images or videos with stimulating content to attract more viewers, advertising helps companies grow and achieve their goals. Ads can be simple to convey content, but some ads present complex messages that require a lot of reasoning steps. Therefore, the field has attracted several researchers who use VQA to address this challenge. Hussain et al. (2017) \cite{hussain2017automatic} proposed the first ads VQA dataset that contains image and video datasets. The image dataset comprises about 64,000 image ads, while the video dataset reaches about 3,500. Apart from the proposed dataset, the authors use a basic VQA model architecture with LSTM and VGGNet for image QA and LSTM and ResNets for video QA. However, the accuracy achieved is low because of the model's simplicity. Using this dataset, Zhou et al. (2020) \cite{zhou2020recommending} proposed a more complex VQA that uses a cross-modality encoder architecture followed by a feed-forward network. Additionally, they use the text generated by OCR in the image ads and the extra knowledge extracted from Wikipedia along with image and question features. The result presents a better performance compared to the previous work.
 Advertisements are vital in business as they capture customer attention and drive sales. By using engaging images or videos, ads help companies grow and achieve their objectives. While some ads convey simple messages, others present complex ideas requiring advanced reasoning, making VQA an important tool for analyzing such content. Hussain et al. (2017) \cite{hussain2017automatic} introduced the first VQA dataset for ads, including around 64,000 image ads and 3,500 video ads. Their basic VQA models, using LSTM with VGGNet for images and LSTM with ResNet for videos, achieved low accuracy due to their simplicity. Building on this, Zhou et al. (2020) \cite{zhou2020recommending} developed a more advanced VQA model with a cross-modality encoder and a feed-forward network. They incorporated OCR-generated text from the ads and external knowledge from Wikipedia, leading to significantly improved performance over the earlier models.
 
 \subsection{VQA for Education}

Recent years have witnessed the emergence of several specialized educational VQA datasets, each tailored to specific subjects and designed to challenge various aspects of reasoning and comprehension. In the realm of mathematics, datasets such as MathVista \cite{lu2023mathvista}  cater to high school and advanced mathematical reasoning. MathVista encompasses a wide range of topics including algebra, geometry, and calculus, requiring models to solve equations, interpret graphs, and perform geometric reasoning. Science-focused datasets such as ScienceVQA \cite{lu2022learn} and the Textbook Question Answering (TQA) dataset \cite{kembhavi2017you} address a wide array of scientific disciplines. ScienceVQA covers topics from physics, chemistry, and biology, integrating questions based on textbook diagrams and experimental setups to evaluate a model’s understanding of scientific principles. TQA, derived from middle school science textbooks, combines visual data from scientific diagrams with textual information, posing complex, multimodal challenges. The AI2D dataset \cite{kembhavi2016diagram} further contributes to this field by focusing specifically on diagram understanding in science, where models must comprehend and reason over typical textbook diagrams.

kembhavi et al. (2017 \cite{kembhavi2017you}) introduced an initial implementation of VQA for educational purposes. They created a dataset called Textbook Question Answering from middle school science curricula. Their baseline model, which used the CNN+VGG architecture without attention mechanisms, faced several limitations. The dataset was small, containing only around 3000 images, and the images were too complex for existing machine comprehension capabilities. As a result, it was not widely adopted by researchers and potential funders. Their model also struggled to capture essential information from both images and questions, achieving only a modest accuracy of 30\%. Attention-based models \cite{gomez2020isaaq}, Relation Network \cite{haurilet2018moqa,li2018textbook, 9996417, ma2022weakly}   Bilinear Pooling \cite{li2018essay} were then applied to improve the performance. some works \cite{chen2021geoqa,chen2022unigeo, zhang2023multi, cao-xiao-2022-augmented} explored geometric VQA models using small datasets including GeoQA, GeoQA+, which included  geometric problems for middle school. However, they encountered similar limitations.

He et al. (2017, \cite{he2017educational}) conducted a noteworthy study that employed VQA to develop a robot aimed at assisting preschool students. This robot was designed to respond to inquiries from preschoolers regarding their surroundings, such as cats or dogs, utilizing voice recognition, image capturing, and answering systems. However, the study lacked an evaluation system to assess the effectiveness of the robot in enhancing children's cognitive development. 
Furthermore, the accuracy of the system was not presented, which is crucial considering the significant impact early learning has on a child's subsequent development.
 
\section{Discussion, challenges, and future research directions} 
\label{Discussion}
According to Table \ref{tab:long}, There have been many developments in Vision Encoder, Language Encoder, as well as Fusion Technique over time. It would be very difficult to discuss these issues in the same section. The development of these modules over the years is also different. So we will discuss the modules separately through the following 4 time periods:

\begin{itemize}
  \item[] Period I: From May 2015 to August 2017
  \item[] Period II: From August 2017 to August 2019
  \item[] Period III: From August 2019 to August 2021
  \item[] Period IV: From August 2021 to Now
\end{itemize}

% \begin{center}
% \centering
%   \begin{figure}[!th]
%   \includegraphics[width=1\linewidth]{Figures/Vision Encoder.pdf}
%   \caption{Trends of using Vision Encoders through the four investigating periods. \textit{Note: GB: Grid-based Encoders; OB: Object-based Encoders; VIT: ViT-based Patch Encoders.}}
%   \label{fig:Vision Encoder}
% \end{figure}  
% \end{center}
\subsection{\textit{Vision Encoder}}
The architecture for building a model depends on how the input image is processed. Models using object-based vision encoders yield better results than grid-based encoders that cannot provide fine-grained inference \cite{anderson2018bottom, teney2018tips}. Grid-based vision encoders have been most widely used in the first and second periods. Object-based encoders were first introduced by Anderson et al. (2018) \cite{anderson2018bottom}, gradually becoming mainstream during the third period. The reason is that the object-based encoders can capture more semantic information than the grid-based encoders, so it achieves better results \cite{anderson2018bottom, teney2018tips}. However, Li et al.(2021) \cite{li2021align} suggest object-based models are both compute-expensive and annotations-expensive because they require bounding box annotations during pre-training and a larger size image during interference. Dou et al. (2022) \cite{dou2022empirical} also agree that the object-based vision encoders have some limitations as they are usually kept frozen during training, and it takes time to extract the local visual features. Kim et al. (2021) \cite{kim2021vilt} studied that the ViT-based Patch vision encoder reduces the model's size, resulting in a four-time and 60-time reduction in runtime compared to the grid-based and object-based, respectively. Apart from that, the ViT-based LVLM models reach better performances than the other \cite{kim2021vilt, dou2022empirical, li2021align}. Therefore, this method has become the most popular encoder. In the future, the ViT-based patch encoder will become a hot trend in LVLM because it provides a powerful and flexible framework for integrating visual and textual information and has shown impressive performance across a range of LVLM tasks.

\begin{table*}[th!]
\centering
\rowcolors{2}{gray!20}{white}
% \begin{spacing}{1.1}
\caption{Comparison of different VQA models on various popular VQA datasets} 
\begin{flushleft}
\footnotesize 
\textbf{Notes:} (-): means unavailable data. (*): Result on Abstract-Scene dataset.  (**): Evaluated on the development test set.\\
The result on VQA v2.0 is OE. \\
\textbf{VA}: Visual Attention. 
\textbf{BP}: Bilinear Pooling. 
\textbf{RN}: Relation Network. 
\textbf{OE}: Open-End. 
\textbf{MC}: Multiple Choice.\\
\end{flushleft}
\label{tab:long}
\resizebox{\linewidth}{!}{
\begin{tabular}{|l|c |c |c |c |c |c |c |c |c |c| c|c|}
% \multicolumn{13}{l}{{ \textbf{Notes:} OE: Open-End. MC: Multiple Choice. -: Unavailable data. *: Result on Abstract-Scene dataset. The result on VQA v2.0 is OE. LVLM: Singe-Stream Fusion Encoder LVLM. LVLM: Dual-Stream Fusion Encoder LVLM.}} \\
% \multicolumn{12}{l}{{ **: Evaluated on the development test set.}} \\

\hline
%\rowcolor{gray!50}
\rowcolor{gray!50}
 \multirow{2} {*} \textbf{\textbf{Model}} &{\begin{tabular}{c}\textbf{Image}\\ \textbf{Encoder}\end{tabular}} & {\begin{tabular}{c}\textbf{Question}\\ \textbf{Encode}\end{tabular}} & {\begin{tabular}{c}\textbf{Fusion}\\ \textbf{Method}\end{tabular}} & \multicolumn{8}{c|}{\textbf{Acc (\%)}}\\
\cline{5-12}
%\rowcolor{gray!50}
\rowcolor{gray!50}
 & & & & \multicolumn{2}{c|}{\textbf{VQA v1}} & \textbf{DAQUAR} & \textbf{Visual7W} & \textbf{COCO-QA} & \textbf{VQA v2} & \textbf{CLEVR} &\textbf{GQA}\\
 \cline{5-6} 
% \rowcolor{gray!50}
\rowcolor{gray!50}
 & & & & EO & MC & & & &  & &\\
\hline
 VQA 2015 \cite{VQA:2014}                            & VGGNet           & LSTM                   & Simple Fusion        & 58.16 & 63.09 & -     & -     & -     & -     & 52.30 & -     \\
Neural-QA 2015 \cite{malinowski2015ask}             & GoogleNet        & LSTM                   & Simple Fusion        & -     & -     & 34.68 & -     & 52.10 & -     & -     & -     \\ 
 2Vis-BiLSTM 2015  \cite{ren2015exploring}           & VGGNet           & BiLSTM                 & Simple Fusion        & -     & -     & 35.78 & -     & 51.61 & -     & -     & -     \\
 ReVisit 2016 \cite{jabri2016revisiting}             & Resnet           & Word2Vec               & Simple Fusion        & -     & 65.22 & -     & \textbf{68.50}& -     & -     & -     & -     \\ 
 CNN-QA 2016 \cite{ma2016learning}                   & VGGNet              & CNN                 & Simple Fusion        & -     & -     & 42.76 & -     & 58.40 & -     & -     & -     \\
 Type-Answer 2016 \cite{kafle2016answer}             & ResNet           & Skip-Thought           & Simple Fusion        & 60.06 & -     & 45.17 & 43.58 & 63.18 &       & -     & -     \\ 
 DualNet 2017 \cite{saito2017dualnet}                & ResNet \& VGGNet & LSTM                   & Simple Fusion        & 61.72 & 66.72 & -     & -     & -     & -     & -     & -     \\
 DualNet 2017 \cite{saito2017dualnet}                & ResNet \& VGGNet & LSTM                  & Simple Fusion        & 69.73*& 74.02*& -     & -     & -     & -     & -     & -     \\ 
\hline 
 ABC-CNN 2015 \cite{chen2015abc}                     & VGGNet           & LSTM                   & VA     & -     & -     & 42.76 & -     & 58.10 & -     & -     & -     \\
 Word+Region 2016 \cite{shih2016look}              & VGGNet       & Word2Vec               & VA     & -     & 62.43 & -     & -     & -     & -     & -     & -     \\ 
Vis7W 2016 \cite{zhu2016visual7w}                & VGGNet           & LSTM                   & VA     & -     & -     & -     & 55.60 & -     & -     & -     & -     \\
 FDA 2016 \cite{ilievski2016focused}                 & ResNet           & LSTM                   & VA     & 59.54 & 64.18 & -     & -     & -     & -     & -     & -     \\ 
 SMem-VQA (2-hop) 2016 \cite{xu2016ask}              & GoogLeNet        & One-Hot                & VA     & 58.24 & -     & 40.07 & -     &       & -     & -     & -     \\
 SANs 2016 \cite{Yang_2016_CVPR}                     & VGGNet           & CNN                    & VA     & 58.90 & -     & \textbf{45.50} & -     & 61.60 & -     & 68.50 & -     \\ 
 MRN 2016 \cite{kim2016multimodal}                   & Resnet           & LSTM                   & VA     & 61.84 & 66.33 & -     & -     & -     & -     & -     & -     \\
 DMN+ 2016 \cite{pmlr-v48-xiong16}                   & Resnet           & GRU                    & VA     & 60.40 & -     & -     & -     & -     & -     & -     & -     \\
  MF-SIG+VG \cite{chen2015abc}                     & ResNet           & GRU                   & VA     & -     & -     & -     & -     & 58.10* & -     & -     & -     \\
 Tips and Tricks 2018 \cite{teney2018tips}           & Faster R-CNN     & GloVe + GRU            & VA     & -     & -     & -     & -     & -     & 64.73 &  & - \\
 BUTD 2018 \cite{anderson2018bottom}                 & Faster R-CNN     & GRU                    & VA     & -     & -     & -     & -     & -     & 63.20 & -     & -     \\
Pythia  2018 \cite{jiang2018pythia}                 & Faster R-CNN     & GloVe + GRU                    & VA     & -     & -     & -     & -     & -     & 72.27 & -     & -     \\ 
 CVA 2018 \cite{ijcai2018p126}                       & Faster R-CNN     & GRU                    & VA     & 66.20 & 70.41 & -     & 63.8  & 67.51 & -     &       & -     \\ 
 \hline

HieCoAtt 2016 \cite{lu2016hierarchical}               & ResNet          & CNN+LSTM               & Co-Attention         & 62.10 & 66.10 & -     & -     & 65.40 & -     & -     & -     \\ 
DAN 2017 \cite{nam2017dual}                           & ResNet          & BiLSTM                 & Co-Attention         & 64.20 & 69.00 & -     & -     & -     & -     & -     & -     \\
DCN 2018 \cite{nguyen2018improved}                    & ResNet          & GloVe + BiLSTM         & Co-Attention         & 67.89 & -     & -     & -     & -     & 67.04 & -     & -     \\ 

BAN 2018 \cite{kim2018bilinear}                       & Faster R-CNN    & GloVe + GRU            & Co-Attention         & -     & -     & -     & -     & -     & 70.35 & -     & -     \\
\hline 
MCAN 2019 \cite{yu2019deep}                           & Faster R-CNN    & GloVe + LSTM           & Transformer          & -     & -     & -     & -     & -     & 70.90 & -     & -     \\
MCAoA 2019 \cite{yu2019deep}                          & Faster R-CNN    & GloVe + LSTM           & Transformer          & -     & -     & -     & -     & -     & 71.14 & -     & -     \\ 
TRAR 2019 \cite{zhou2021trar}                           & ResNet         & GloVe + LSTM           & Transformer          & -     & -     & -     & -     & -     & 72.93 & 99.10 & -     \\ 
TxT 2022 \cite{steitz2022txt}                         & ResNet          & GloVe + LSTM           & Transformer        & -     & -     & -     & -     & -     & 69.93** & -     & -     \\ 
\hline
NMN 20
16 \cite{andreas2016neural}                     & VGGNet          & LSTM                   & NMN                  & 55.10 & -     & -     & -     & -     & -     & 75.71  & -     \\
N2NMN 2017 \cite{hu2017learning}                      & ResNet          & LSTM                   & NMN                  & 64.90*& -     & -     & -     & -     & -     & 83.70  & -     \\
IEP 2017 \cite{johnson2017inferring}                  & ResNet          & LSTM                   & NMN                  & -     & -     & -     & -     & -     & -     & 96.90  & -     \\
Stack-NMN 2018 \cite{hu2018explainable}               & ResNet          & Bi-LSTM                & NMN                  & 66.00 & -     & -     & -     & -     & 64.10 & 96.50  & -     \\
NS-NMN 2018 \cite{yi2018neural}                       & Mask R-CNN      & LSTN                   & NMN                  & -     & -     & -     & -     & -     & -     & \textbf{99.80} & -     \\
Prob-NMN 2019 \cite{vedantam2019probabilistic}        & ResNet          & LSTM                   & NMN                  & -     & -     & -     & -     & -     & -     & 97.73  & -     \\
ProTo 2021 \cite{zhao2021proto}                       & Faster R-CNN    & Transformers           & NMN                 & -     & -     & -     & -     & -     & -     & -     & \textbf{65.14}\\
\hline 

MCB + Att 2016 \cite{fukui2016multimodal}             & ResNet          & GloVe + LSTM           & BP     & 66.47** & 70.10** & -     & -     & 62.20 & 62.27** & -     & -      \\
MLB 2016 \cite{kim2016hadamard}                       & ResNet          & GRU                    & BP    & 66.89 & 70.29 & -     & -     & -     & -     & -     & -       \\
MFB 2017 \cite{yu2017multi}                           & ResNet          & GloVe + LSTM           & BP     & 68.40 & 72.50 & -     & -     & -     & -     & -     & -     \\
MUTAN 2017 \cite{ben2017mutan}                        & ResNet          & GRU                    & BP     & 67.36 & -     & -     & -     & -     & -     & -     & -     \\
Film 2018 \cite{perez2018film}                        & ResNet          & GRU                    & BP     & -     & -     & -     & -     & -     & -     & 97.7  & -     \\
BLOCK 2019 \cite{ben2019block}                        & Faster R-CNN    & Skip-Thought           & BP     & -     & -     & -     & -     & -     & 67.92 & -     & -     \\
\hline 
Simple RN 2017 \cite{santoro2017simple}               & CNN             & LSTM                  & RN     & -     & -     & -     & -     & -     & -     & 95.50 & -      \\
Graph-VQA 2017 \cite{teney2017graph}                  & ResNet          & LSTM                  & RN    & \textbf{70.42*}& \textbf{74.37*}& -     & -     & -     & -     & -& -\\
Graph learner 2018 \cite{norcliffe2018learning}       & Faster R-CNN    & GRU                   & RN     & -     & -     & -     & -     & -     & 66.18 & -       & -      \\
ReasonNet 2019 \cite{gao2019multi}                    & ResNet          & BOW          & RN     & 67.9  & -     & -     & -     & -     & 64.61 & -       & -      \\
MLIN-BERT 2019 \cite{gao2019multi}                    & Faster R-CNN    & BERT                  & RN     & -     & -     & -     & -     & -     & 71.27 & -       & -      \\
CoR 2018 \cite{wu2018chain}                            & Faster R-CNN    & GRU                   & RN    & \textbf{68.54}& \textbf{72.93}& -     & -     & \textbf{69.38}& 69.14 & -& -\\\
ReGAT 2019 \cite{li2019relation}                      & Faster R-CNN    & GloVe + GRU           & RN     & -     & -     & -     & -     & -     & 70.58 & -       & -      \\
MLIN-BERT 2019 \cite{gao2019multi}                    & Faster R-CNN    & BERT                  & RN     & -     & -     & -     & -     & -     & 71.27 & -       & -      \\
NSM 2019 \cite{hudson2019learning}                    & Mask R-CNN      & GloVe+LSTM            & RN     & -     & -     & -     & -     & -     & -     & -       & 63.17   \\

\hline 

VisualBERT 2019 \cite{li2019visualbert}               & Faster R-CNN    & BERT        & LVLM            & -     & -     & -     & -     & -     & 71.00 & -     & -      \\

VL-BERT 2019 \cite{su2019vl}                          & Faster R-CNN    & BERT                  & LVLM    & -     & -     & -     & -     & -     & 72.22 & -     & -     \\
ViLT 2019 \cite{kim2021vilt}                          & Patch Embedding &  BERT                 & LVLM    & -     & -     & -     & -     & -     & 71.26** & -     & -     \\
ViLBERT 2019 \cite{lu2019vilbert}                     & Faster R-CNN    & BERT                   & LVLM      &       & -     & -     & -     & -     & 70.92 & -  & -     \\
LXMERT 2019 \cite{tan2019lxmert}                      & Faster R-CNN    & BERT                   & LVLM      & -     & -     & -     & -     & -     & 72.50 & -  & 60.33 \\
Pixel-BERT 2020 \cite{huang2020pixel}                 & ResNet          & BERT                  & LVLM    & -     & -     & -     & -     & -     & 74.55 & -     & -     \\
UNITER 2020 \cite{chen2020uniter}                     & Faster R-CNN    & Word Embedding        & LVLM    & -     & -     & -     & -     & -     & 74.02 & -     & -     \\
OSCAR 2020 \cite{li2020oscar}                         & Faster R-CNN    &  Word Embedding       & LVLM    & -     & -     & -     & -     & -     & 73.82 & -     & 61.62 \\
CLIP-ViL 2020 \cite{shen2021much}                     & CLIP          &  Word Embedding       & LVLM    & -     & -     & -     & -     & -     & 76.70 & -     & 62.93 \\
UNIMO 2020 \cite{li2020unimo}                         & Faster R-CNN    & Word Embedding        & LVLM    & -     & -     & -     & -     & -     & 75.27 & -     & -     \\
Oscar+ 2021 \cite{zhang2021vinvl}                     & Faster R-CNN    & Word Embedding        & LVLM    & -     & -     & -     & -     & -     & 76.60 & -     & 64.65 \\
VL-T5 2021 \cite{cho2021unifying}                     & Faster R-CNN    & Word Embedding        & LVLM    & -     & -     & -     & -     & -     & 70.30 & -     & 60.80  \\
VL-BART 2021 \cite{cho2021unifying}                   & Faster R-CNN    & Word Embedding        & LVLM    & -     & -     & -     & -     & -     & 71.3 & -      & 60.5  \\
MDETR 2021 \cite{kamath2021mdetr}                     & ResNet          & RoBERTa               & LVLM    & -     & -     & -     & -     & -     & -     & 99.70 & 62.45 \\
ERNIE-ViL 2021 \cite{yu2021ernie}                     & Faster R-CNN    & Word Embedding         & LVLM    & -     & -     & -     & -     & -     & 75.10 & -     & -     \\
ALBEF 2021 \cite{li2021align}                         & ViT             & BERT                   & LVLM      & -     & -     & -     & -     & -     & 76.04 & -  & -     \\
Florence 2021 \cite{yuan2021florence}                 & CNN + ViT       T& BERT               & LVLM      & -     & -     & -     & -     & -    & 80.16 & -     & -     \\

UNIMO-2 2022 \cite{li2022unimo}                       & ViT             & RoBERTa               & LVLM    & -     & -     & -     & -     & -     & 76.42 & -     & -     \\
SimVLM 2021 \cite{wang2021simvlm}                     & Patch + CNN & Word Embedding        & LVLM    & -     & -     & -     & -     & -     & 80.34 & -     & -     \\
VLMo 2021 \cite{bao2021vlmo}                          & Patch Embedding & BERT                  & LVLM    & -     & -     & -     & -     & -     & 82.78 & -     & -     \\
GIT 2022 \cite{wang2022git}                           & ViT             & Word Embedding        & LVLM    & -     & -     & -     & -     & -     & 78.81 & -     & -     \\
OFA-1B 2022 \cite{wang2022ofa}                           & ResNet           & Word Embedding     & LVLM    & -     & -     & -     & -     & -     & 82.00 & -     & -     \\
X-VLM  2022 \cite{zeng2022multi}                      & ViT            & BERT                  & LVLM    & -     & -     & -     & -     & -     & 78.37 & -     & -     \\
X$^2$-VLM  2022 \cite{zeng2022x}                      & ViT             & BERT                  & LVLM    & -     & -     & -     & -     & -     & 80.50 & -     & -     \\
BEiT-3 2B 2022 \cite{wang2022image}                      & ViT             & Transformer           & LVLM    & -     & -     & -     & -     & -     & 84.03 & -     & -     \\

METER 2022 \cite{dou2022empirical}                    & ViT             & RoBERTa & LVLM      & -     & -     & -     & -     & -     & 77.64 & -     & -     \\
Flamingo-80B 2022 \cite{alayrac2022flamingo}              & NFNet \cite{brock2021high} & Word Embedding & LVLM      & -     & -     & -     & -     & -     & 82.1  & -     & -     \\
CoCa-2.1B 2022 \cite{yu2022coca}                           & ViT            & Transformers                    & LVLM      & -     & -     & -     & -     & -     & 82.3  & -     & -     \\
PaLi-17B 2022 \cite{chen2022pali}                         & ViT             & mT5 \cite{xue2020mt5} & LVLM    & -     & -     & -     & -     & -     & \textbf{84.30}& -     & -     \\
mPLUG 2022 \cite{li2022mplug}                        & ViT             & BERT & LVLM    & -     & -     & -     & -     & -     & 82.41& -     & -     \\
mPLUG 2022 \cite{li2022mplug}                        & ViT             & BERT & LVLM    & -     & -     & -     & -     & -     & 82.41& -     & -     \\
BLIP-13B 2023 \cite{li2023blip2bootstrappinglanguageimagepretraining}   & VIT             & OPT & LVLM    & -     & -     & -     & -     & -     & 82.19& -     & -     \\
Qwen-VL 7B 2023\cite{bai2023qwen}                       & ViT             & QwenLM \cite{bai2023qwen} & LVLM    & -     & -     & -     & -     & -     & 78.80& -     & 63.80     \\
LLAVA-1.5 13B 2024\cite{liu2024improved}                       & ViT             & Vicuna-1.5 13B \cite{zheng2023judging} & LVLM    & -     & -     & -     & -     & -     & 80.80& -     & 64.70     \\
\hline
\end{tabular}
}
% \end{spacing}
\end{table*}
% \subsection{\textit{Language Encoder}}

% \begin{figure}[!th]
% \centering
% \includegraphics[width=1\linewidth]{Figures/Language Encoder.pdf}
% \caption{Trends of using Language Encoders through the four investigating periods.}
% \label{fig:Language Encoder}
% \end{figure}  
% Acoording to Fig. \ref{fig:Vision Encoder}, 

\subsection{\textit{Language Encoder}}
RNN models such as LSTM and GRU mainly was used before in the first and second periods. After the birth of BERT \cite{devlin2018bert}, Transformer language encoders are mainly used to extract the question features in the third and fourth periods. In addition to these two popular methods, some researchers have used some simple methods such as BOW \cite{shih2016look, gao2019multi} and CNN \cite{Yang_2016_CVPR, lu2016hierarchical} in the early stages. 

Among the used language encoders, pre-train Transformer language models have shown impressive performance (Table \ref{tab:long}). For example, Dou et al. (2022) \cite{dou2022empirical} have studied various languages models, including BERT, RoBERTa, ELECTRA \cite{clark2020electra}, ALBERT \cite{lan2019albert}, and DeBERTa \cite{he2020deberta} for text encoding and shown that RoBERTa is the best model among them. Chen et al. (2022) \cite{chen2022pali} and  Alayrac et al. (2022) \cite{alayrac2022flamingo} use huge pre-train language models with respectively 13 billion and 70 billion parameters for text encoding and reach the state-of-the-art in VQA. Recently, the trend of using a LLM \cite{liu2023llava, alayrac2022flamingo, liu2024llavanext, li2023blip2bootstrappinglanguageimagepretraining} as a backbone and then employing a vision encoder for training, in order to save resources like time and data, has demonstrated remarkable results and is expected to continue in the future.

\subsection{\textit{Fusion Method}}

\begin{landscape}
% \mbox{}\vfill
  \begin{figure}
  \centering
  \includegraphics[width=\linewidth]{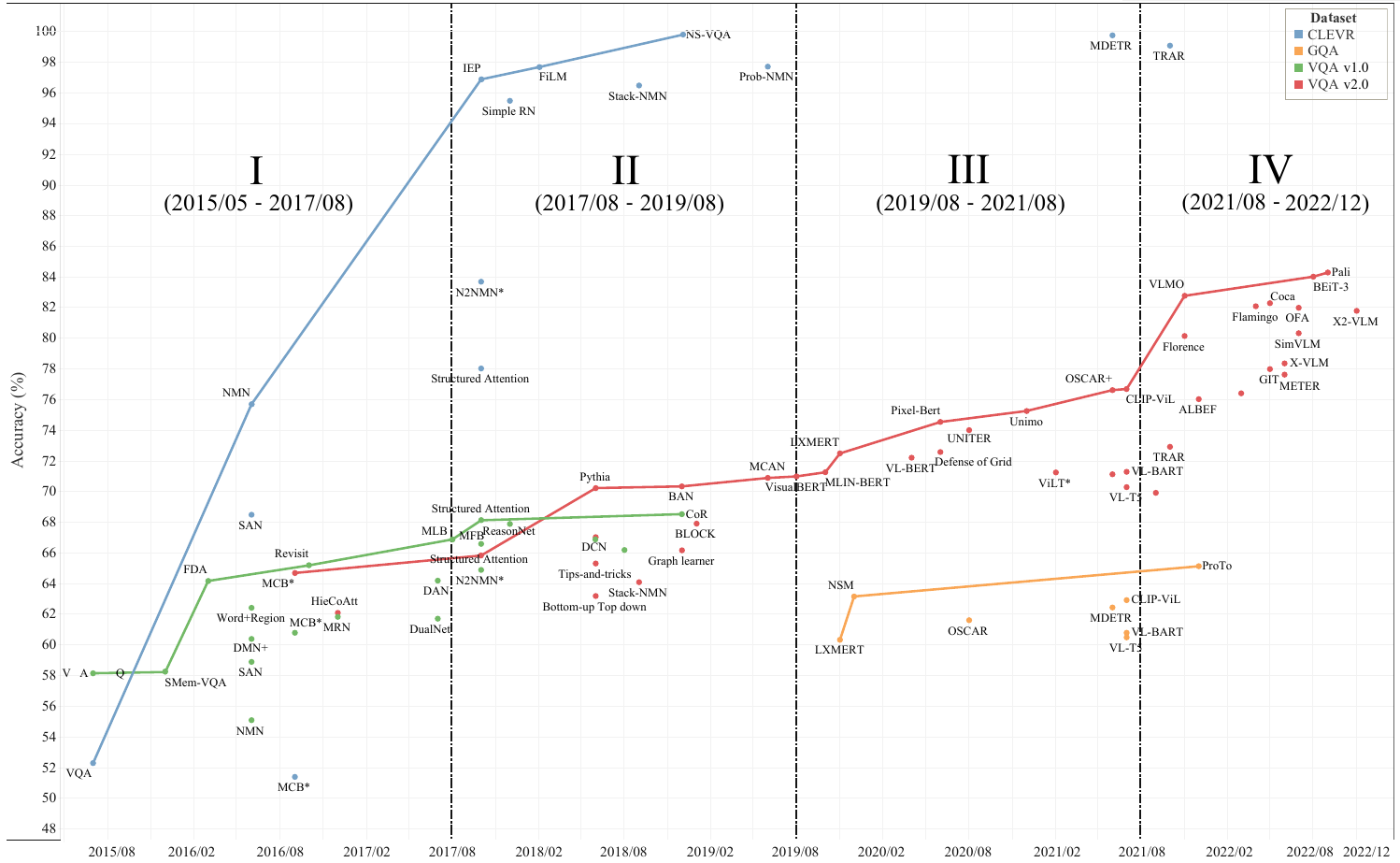}
  \caption{The accuracy of the model on different datasets over the years (from May 2015 to December 2022). \textit{Note: the results on the VQA v1.0 and VQA v2.0 sets are from the test-std sets. * represents the results from the test-dev set. The different colored lines represent the state-of-the-art models in the datasets over time.}}
  \label{fig:accuracylinechart}
\end{figure}
% \vfill
\end{landscape}

Most researchers have focused on improving this part rather than others. Therefore, the development of fusion methods is very diverse. Fig. \ref{fig:accuracylinechart} presents the state-of-art of VQA models as well as others from 2015 to No.

\subfour{Period I:} Initially, RNN-CNN, followed by a simple fusion operation, is used to build a VQA model. This end-to-end approach is straightforward, reaching deficient performance \cite{VQA:2014}. From 2015 to 2017, an visual-based model was proposed in VQA, which uses attention machines to weigh the contribution of different regions of the image and the input question. Therefore, the viusal-attention-based models outperform the previous works \cite{Yang_2016_CVPR, ren2015exploring}. At the same time, some bilinear pooling fusion techniques were introduced and obtained better results since they can capture the interaction between the image and the question \cite{fukui2016multimodal, kim2016hadamard, yu2017multi,  ben2017mutan}. Besides, Fukui et al. \cite{fukui2016multimodal} integrated these pooling techniques into the attention-based models to improve the VQA performance. Apart from these approaches, Relation Networks \cite{santoro2017simple, teney2017graph} and NMN models \cite{andreas2016neural, hu2017learning, johnson2017inferring} also achieved certain achievements. The accuracy on the VQA v1.0 dataset in this period ranged from 58\% to 64\%.

\subfour{Period II:} Co-attention machines are proposed to capture both visual and question attention \cite{nguyen2018improved, yu2018beyond, nam2017dual}.  However, Yu et. al (2019) \cite{yu2019deep} recommended that these co-attention models are limited to capturing the word-to-word relationship for questions, and the region-to-region relationship for images. Therefore, they used Transformer to build MCAN \cite{yu2019deep}. The performance on the VQA v1.0 and v2.0 was from 66\% to 71\%. 

\subfour{Period III:} Some Transformer-based VQA models were proposed \cite{yu2019deep, zhou2021trar, zhao2021proto}. However, LVLM appeared and have become the main topic to solve not only VQA but also other VL tasks such as Visual Caption. Initially, the LVLM models used object-based vision encoders for visual encoding \cite{li2019visualbert, su2019vl, chen2020uniter, li2020oscar, li2020unimo, zhang2021vinvl, cho2021unifying, lu2019vilbert, tan2019lxmert, yu2021ernie}. Later, the researchers replaced that with grid-based and ViT-based patch encoders to build the end-to-end LVLM models \cite{kim2021vilt, huang2020pixel, shen2021much, kamath2021mdetr, li2021align}. During this period, these model sizes are usually medium-scale, for instance ALIGN with 820M parameters and Florence with 893M parameters. The accuracy on the VQA v2.0 ranged from 71\% to 78\%.

\subfour{Period IV:} LVLMs continue to dominate the landscape for solving Visual-Language tasks, including VQA. Recent advancements have seen the introduction of highly expansive LVLMs from leading companies such as Google, with models like SimVLM \cite{wang2021simvlm}, Coca \cite{yu2022coca}, and PaLI \cite{chen2022pali}; Microsoft, with METER \cite{dou2022empirical}, VLMo \cite{bao2021vlmo}, GIT \cite{wang2022git}, and BEit-3 \cite{wang2022image}; and DeepMind, with Flamingo \cite{alayrac2022flamingo}. The scale of these models ranges from a few billion to tens of billions of parameters, such as PaLI \cite{wang2022image} with 16.9 billion parameters and Flamingo with 80.2 billion parameters. These models have consistently achieved top-tier performance, with accuracy on the VQA v2.0 benchmark ranging from 78% to 84%, led by PaLI \cite{wang2022image}.

In the future, these LVLMs are expected to significantly enhance zero-shot and few-shot learning capabilities, reducing the need for extensive fine-tuning. By leveraging the vast knowledge encoded within these large-scale models, future iterations will likely excel at generalizing across tasks and domains with minimal additional training. This improvement will make VQA systems more adaptable, capable of performing new tasks and answering complex queries without requiring task-specific fine-tuning, thereby streamlining the deployment of these models in diverse real-world applications.
In fact, the construction of the original VQA models has been replaced by LVLM because LVLM can both solve many tasks at the same time and achieve high results in VL tasks. LVLM will continue to be a hot area that will attract the attention of future researchers.

\subsection{\textit{Application}}
Future developments in VQA are poised to significantly impact education and healthcare, particularly in complex mathematical problem-solving and medical imaging. In education, VQA systems are expected to advance by tackling more sophisticated mathematical reasoning, moving beyond simple calculations to interpreting complex equations, geometric shapes, and graphs. These systems could assist students by answering queries like “What is the derivative of this function?” or “Explain the steps to solve this equation,” offering interactive, real-time learning support.

In healthcare, VQA’s role in medical imaging will likely expand, providing more accurate and detailed analysis of complex medical data, such as MRIs, CT scans, and X-rays. 

\noindent
Future systems may not only answer questions like ``Is there a tumor present?'' but also provide detailed insights, offering suggestions for potential diagnoses or treatment plans. These advancements could reduce diagnostic errors and enhance personalized patient care. To achieve this, research needs to focus on improving the reasoning capabilities of VQA systems and developing larger, domain-specific datasets to train models effectively in these critical fields. 

\section{Conclusion}
\label{sec:conslusion}
VQA represents a significant advancement in the field of artificial intelligence, enabling machines to interpret and respond to questions about visual content. Over the years, VQA has evolved from simple image-based question-answering systems to sophisticated models capable of complex reasoning, grounded in both vision and language understanding.

The key takeaway from this survey is the versatility of VQA across various domains. From medical diagnostics and educational tools to accessibility for visually impaired individuals, the applications of VQA are broad and impactful. However, challenges remain, such as the need for domain-specific datasets, enhanced reasoning capabilities, and real-time system performance.

Looking ahead, the integration of LVLMs has the potential to push VQA into new territories, improving accuracy and extending applicability to more diverse and intricate tasks. As VQA continues to grow, addressing the gaps in dataset availability, computational efficiency, and real-world deployment will be crucial for realizing its full potential.

\bibliographystyle{unsrtnat}
\bibliography{biblio}

\begin{thebibliography}{266}
\providecommand{\natexlab}[1]{#1}
\providecommand{\url}[1]{\texttt{#1}}
\expandafter\ifx\csname urlstyle\endcsname\relax
  \providecommand{\doi}[1]{doi: #1}\else
  \providecommand{\doi}{doi: \begingroup \urlstyle{rm}\Url}\fi

\bibitem[Baltrušaitis et~al.(2019)Baltrušaitis, Ahuja, and Morency]{8269806}
Tadas Baltrušaitis, Chaitanya Ahuja, and Louis-Philippe Morency.
\newblock Multimodal machine learning: A survey and taxonomy.
\newblock \emph{IEEE Transactions on Pattern Analysis and Machine
  Intelligence}, 41\penalty0 (2):\penalty0 423--443, 2019.
\newblock \doi{10.1109/TPAMI.2018.2798607}.

\bibitem[Hori et~al.(2017)Hori, Hori, Lee, Zhang, Harsham, Hershey, Marks, and
  Sumi]{8237712}
Chiori Hori, Takaaki Hori, Teng-Yok Lee, Ziming Zhang, Bret Harsham, John~R.
  Hershey, Tim~K. Marks, and Kazuhiko Sumi.
\newblock Attention-based multimodal fusion for video description.
\newblock In \emph{2017 IEEE International Conference on Computer Vision
  (ICCV)}, pages 4203--4212, 2017.
\newblock \doi{10.1109/ICCV.2017.450}.

\bibitem[Bagher~Zadeh et~al.(2018)Bagher~Zadeh, Liang, Poria, Cambria, and
  Morency]{bagher-zadeh-etal-2018-multimodal}
AmirAli Bagher~Zadeh, Paul~Pu Liang, Soujanya Poria, Erik Cambria, and
  Louis-Philippe Morency.
\newblock Multimodal language analysis in the wild: {CMU}-{MOSEI} dataset and
  interpretable dynamic fusion graph.
\newblock In \emph{Proceedings of the 56th Annual Meeting of the Association
  for Computational Linguistics (Volume 1: Long Papers)}, pages 2236--2246,
  Melbourne, Australia, July 2018. Association for Computational Linguistics.
\newblock \doi{10.18653/v1/P18-1208}.
\newblock URL \url{https://aclanthology.org/P18-1208}.

\bibitem[Zellers et~al.(2022)Zellers, Lu, Lu, Yu, Zhao, Salehi, Kusupati,
  Hessel, Farhadi, and Choi]{Zellers_2022_CVPR}
Rowan Zellers, Jiasen Lu, Ximing Lu, Youngjae Yu, Yanpeng Zhao, Mohammadreza
  Salehi, Aditya Kusupati, Jack Hessel, Ali Farhadi, and Yejin Choi.
\newblock Merlot reserve: Neural script knowledge through vision and language
  and sound.
\newblock In \emph{Proceedings of the IEEE/CVF Conference on Computer Vision
  and Pattern Recognition (CVPR)}, pages 16375--16387, June 2022.

\bibitem[Ektefaie et~al.(2023)Ektefaie, Dasoulas, Noori, Farhat, and
  Zitnik]{ektefaie2023multimodal}
Yasha Ektefaie, George Dasoulas, Ayush Noori, Maha Farhat, and Marinka Zitnik.
\newblock Multimodal learning with graphs.
\newblock \emph{Nature Machine Intelligence}, pages 1--11, 2023.

\bibitem[Wang et~al.(2021)Wang, Yu, Yu, Dai, Tsvetkov, and Cao]{wang2021simvlm}
Zirui Wang, Jiahui Yu, Adams~Wei Yu, Zihang Dai, Yulia Tsvetkov, and Yuan Cao.
\newblock Simvlm: Simple visual language model pretraining with weak
  supervision.
\newblock \emph{arXiv preprint arXiv:2108.10904}, 2021.

\bibitem[Alayrac et~al.(2022)Alayrac, Donahue, Luc, Miech, Barr, Hasson, Lenc,
  Mensch, Millican, Reynolds, et~al.]{alayrac2022flamingo}
Jean-Baptiste Alayrac, Jeff Donahue, Pauline Luc, Antoine Miech, Iain Barr,
  Yana Hasson, Karel Lenc, Arthur Mensch, Katie Millican, Malcolm Reynolds,
  et~al.
\newblock Flamingo: a visual language model for few-shot learning.
\newblock \emph{arXiv preprint arXiv:2204.14198}, 2022.

\bibitem[Park and Kim(2023)]{PARK2023100548}
Sang-Min Park and Young-Gab Kim.
\newblock Visual language integration: A survey and open challenges.
\newblock \emph{Computer Science Review}, 48:\penalty0 100548, 2023.
\newblock ISSN 1574-0137.
\newblock \doi{https://doi.org/10.1016/j.cosrev.2023.100548}.
\newblock URL
  \url{https://www.sciencedirect.com/science/article/pii/S1574013723000151}.

\bibitem[Antol et~al.(2015{\natexlab{a}})Antol, Agrawal, Lu, Mitchell, Batra,
  Zitnick, and Parikh]{Antol_2015_ICCV}
Stanislaw Antol, Aishwarya Agrawal, Jiasen Lu, Margaret Mitchell, Dhruv Batra,
  C.~Lawrence Zitnick, and Devi Parikh.
\newblock Vqa: Visual question answering.
\newblock In \emph{Proceedings of the IEEE International Conference on Computer
  Vision (ICCV)}, December 2015{\natexlab{a}}.

\bibitem[Ren et~al.(2015{\natexlab{a}})Ren, Kiros, and Zemel]{ren2015exploring}
Mengye Ren, Ryan Kiros, and Richard Zemel.
\newblock Exploring models and data for image question answering.
\newblock \emph{Advances in neural information processing systems}, 28,
  2015{\natexlab{a}}.

\bibitem[Zhang et~al.(2016)Zhang, Goyal, Summers-Stay, Batra, and
  Parikh]{Zhang_2016_CVPR}
Peng Zhang, Yash Goyal, Douglas Summers-Stay, Dhruv Batra, and Devi Parikh.
\newblock Yin and yang: Balancing and answering binary visual questions.
\newblock In \emph{Proceedings of the IEEE Conference on Computer Vision and
  Pattern Recognition (CVPR)}, June 2016.

\bibitem[Ma et~al.(2016)Ma, Lu, and Li]{ma2016learning}
Lin Ma, Zhengdong Lu, and Hang Li.
\newblock Learning to answer questions from image using convolutional neural
  network.
\newblock In \emph{Thirtieth AAAI Conference on Artificial Intelligence}, 2016.

\bibitem[Goyal et~al.(2017{\natexlab{a}})Goyal, Khot, Summers-Stay, Batra, and
  Parikh]{Goyal_2017_CVPR}
Yash Goyal, Tejas Khot, Douglas Summers-Stay, Dhruv Batra, and Devi Parikh.
\newblock Making the v in vqa matter: Elevating the role of image understanding
  in visual question answering.
\newblock In \emph{Proceedings of the IEEE Conference on Computer Vision and
  Pattern Recognition (CVPR)}, July 2017{\natexlab{a}}.

\bibitem[Antol et~al.(2015{\natexlab{b}})Antol, Agrawal, Lu, Mitchell, Batra,
  Zitnick, and Parikh]{VQA:2014}
Stanislaw Antol, Aishwarya Agrawal, Jiasen Lu, Margaret Mitchell, Dhruv Batra,
  C.~Lawrence Zitnick, and Devi Parikh.
\newblock {VQA}: {V}isual {Q}uestion {A}nswering.
\newblock In \emph{International Conference on Computer Vision (ICCV)},
  2015{\natexlab{b}}.

\bibitem[Krishna et~al.(2016)Krishna, Zhu, Groth, Johnson, Hata, Kravitz, Chen,
  Kalantidis, Li, Shamma, Bernstein, and Fei{-}Fei]{gva}
Ranjay Krishna, Yuke Zhu, Oliver Groth, Justin Johnson, Kenji Hata, Joshua
  Kravitz, Stephanie Chen, Yannis Kalantidis, Li{-}Jia Li, David~A. Shamma,
  Michael~S. Bernstein, and Li~Fei{-}Fei.
\newblock Visual genome: Connecting language and vision using crowdsourced
  dense image annotations.
\newblock \emph{CoRR}, abs/1602.07332, 2016.
\newblock URL \url{http://arxiv.org/abs/1602.07332}.

\bibitem[Young et~al.(2014)Young, Lai, Hodosh, and
  Hockenmaier]{young-etal-2014-image}
Peter Young, Alice Lai, Micah Hodosh, and Julia Hockenmaier.
\newblock From image descriptions to visual denotations: New similarity metrics
  for semantic inference over event descriptions.
\newblock \emph{Transactions of the Association for Computational Linguistics},
  2:\penalty0 67--78, 2014.
\newblock \doi{10.1162/tacl_a_00166}.
\newblock URL \url{https://aclanthology.org/Q14-1006}.

\bibitem[Kembhavi et~al.(2017{\natexlab{a}})Kembhavi, Seo, Schwenk, Choi,
  Farhadi, and Hajishirzi]{8100054}
Aniruddha Kembhavi, Minjoon Seo, Dustin Schwenk, Jonghyun Choi, Ali Farhadi,
  and Hannaneh Hajishirzi.
\newblock Are you smarter than a sixth grader? textbook question answering for
  multimodal machine comprehension.
\newblock In \emph{2017 IEEE Conference on Computer Vision and Pattern
  Recognition (CVPR)}, pages 5376--5384, 2017{\natexlab{a}}.
\newblock \doi{10.1109/CVPR.2017.571}.

\bibitem[Lin et~al.(2021)Lin, Zhang, Tac, Shi, Haffari, Wu, He, and
  Ge]{lin2021medical}
Zhihong Lin, Donghao Zhang, Qingyi Tac, Danli Shi, Gholamreza Haffari, Qi~Wu,
  Mingguang He, and Zongyuan Ge.
\newblock Medical visual question answering: A survey.
\newblock \emph{arXiv preprint arXiv:2111.10056}, 2021.

\bibitem[Gong et~al.(2021{\natexlab{a}})Gong, Chen, Liu, Yu, and
  Li]{10.1145/3460426.3463584}
Haifan Gong, Guanqi Chen, Sishuo Liu, Yizhou Yu, and Guanbin Li.
\newblock Cross-modal self-attention with multi-task pre-training for medical
  visual question answering.
\newblock In \emph{Proceedings of the 2021 International Conference on
  Multimedia Retrieval}, ICMR '21, page 456–460, New York, NY, USA,
  2021{\natexlab{a}}. Association for Computing Machinery.
\newblock ISBN 9781450384636.
\newblock \doi{10.1145/3460426.3463584}.
\newblock URL \url{https://doi.org/10.1145/3460426.3463584}.

\bibitem[Ambati and Reddy~Dudyala(2018)]{8987108}
Rahul Ambati and Chakravardhan Reddy~Dudyala.
\newblock A sequence-to-sequence model approach for imageclef 2018 medical
  domain visual question answering.
\newblock In \emph{2018 15th IEEE India Council International Conference
  (INDICON)}, pages 1--6, 2018.
\newblock \doi{10.1109/INDICON45594.2018.8987108}.

\bibitem[Allaouzi et~al.(2019)Allaouzi, Ahmed, and Benamrou]{Allaouzi2019AnEM}
Imane Allaouzi, Mohamed~Ben Ahmed, and Badr Benamrou.
\newblock An encoder-decoder model for visual question answering in the medical
  domain.
\newblock In \emph{Conference and Labs of the Evaluation Forum}, 2019.

\bibitem[Gong et~al.(2021{\natexlab{b}})Gong, Huang, Chen, and
  Li]{Gong2021SYSUHCPAV}
Haifan Gong, Ricong Huang, Guanqi Chen, and Guanbin Li.
\newblock Sysu-hcp at vqa-med 2021: A data-centric model with efficient
  training methodology for medical visual question answering.
\newblock In \emph{CLEF}, 2021{\natexlab{b}}.

\bibitem[Hasan et~al.(2018{\natexlab{a}})Hasan, Ling, Farri, Liu, Lungren, and
  M\"uller]{ImageCLEFVQA-Med2018}
Sadid~A. Hasan, Yuan Ling, Oladimeji Farri, Joey Liu, Matthew Lungren, and
  Henning M\"uller.
\newblock Overview of the {ImageCLEF} 2018 medical domain visual question
  answering task.
\newblock In \emph{CLEF2018 Working Notes}, {CEUR} Workshop Proceedings,
  Avignon, France, September 10-14 2018{\natexlab{a}}. CEUR-WS.org
  $<$http://ceur-ws.org$>$.

\bibitem[Lau et~al.(2018{\natexlab{a}})Lau, Gayen, Abacha, and
  Demner-Fushman]{Lau2018ADO}
Jason~J. Lau, Soumya Gayen, Asma~Ben Abacha, and Dina Demner-Fushman.
\newblock A dataset of clinically generated visual questions and answers about
  radiology images.
\newblock \emph{Scientific Data}, 5, 2018{\natexlab{a}}.

\bibitem[{Ben Abacha} et~al.(2021){Ben Abacha}, Sarrouti, Demner-Fushman,
  Hasan, and M\"uller]{ImageCLEF-VQA-Med2021}
Asma {Ben Abacha}, Mourad Sarrouti, Dina Demner-Fushman, Sadid~A. Hasan, and
  Henning M\"uller.
\newblock Overview of the vqa-med task at imageclef 2021: Visual question
  answering and generation in the medical domain.
\newblock In \emph{CLEF 2021 Working Notes}, {CEUR} Workshop Proceedings,
  Bucharest, Romania, September 21-24 2021. CEUR-WS.org.

\bibitem[Gordon et~al.(2018)Gordon, Kembhavi, Rastegari, Redmon, Fox, and
  Farhadi]{Gordon_2018_CVPR}
Daniel Gordon, Aniruddha Kembhavi, Mohammad Rastegari, Joseph Redmon, Dieter
  Fox, and Ali Farhadi.
\newblock Iqa: Visual question answering in interactive environments.
\newblock In \emph{Proceedings of the IEEE Conference on Computer Vision and
  Pattern Recognition (CVPR)}, June 2018.

\bibitem[Vinyals et~al.(2015)Vinyals, Toshev, Bengio, and
  Erhan]{Vinyals_2015_CVPR}
Oriol Vinyals, Alexander Toshev, Samy Bengio, and Dumitru Erhan.
\newblock Show and tell: A neural image caption generator.
\newblock In \emph{Proceedings of the IEEE Conference on Computer Vision and
  Pattern Recognition (CVPR)}, June 2015.

\bibitem[Mao et~al.(2016)Mao, Huang, Toshev, Camburu, Yuille, and
  Murphy]{Mao_2016_CVPR}
Junhua Mao, Jonathan Huang, Alexander Toshev, Oana Camburu, Alan~L. Yuille, and
  Kevin Murphy.
\newblock Generation and comprehension of unambiguous object descriptions.
\newblock In \emph{Proceedings of the IEEE Conference on Computer Vision and
  Pattern Recognition (CVPR)}, June 2016.

\bibitem[Deng et~al.(2021)Deng, Yang, Chen, Zhou, and Li]{Deng_2021_ICCV}
Jiajun Deng, Zhengyuan Yang, Tianlang Chen, Wengang Zhou, and Houqiang Li.
\newblock Transvg: End-to-end visual grounding with transformers.
\newblock In \emph{Proceedings of the IEEE/CVF International Conference on
  Computer Vision (ICCV)}, pages 1769--1779, October 2021.

\bibitem[Gao et~al.(2020)Gao, Li, Song, and Shen]{8620348}
Lianli Gao, Xiangpeng Li, Jingkuan Song, and Heng~Tao Shen.
\newblock Hierarchical lstms with adaptive attention for visual captioning.
\newblock \emph{IEEE Transactions on Pattern Analysis and Machine
  Intelligence}, 42\penalty0 (5):\penalty0 1112--1131, 2020.
\newblock \doi{10.1109/TPAMI.2019.2894139}.

\bibitem[Gan et~al.(2017)Gan, Gan, He, Pu, Tran, Gao, Carin, and
  Deng]{Gan_2017_CVPR}
Zhe Gan, Chuang Gan, Xiaodong He, Yunchen Pu, Kenneth Tran, Jianfeng Gao,
  Lawrence Carin, and Li~Deng.
\newblock Semantic compositional networks for visual captioning.
\newblock In \emph{Proceedings of the IEEE Conference on Computer Vision and
  Pattern Recognition (CVPR)}, July 2017.

\bibitem[Zhang et~al.(2019)Zhang, Cao, and Wu]{zhang2019information}
Dongxiang Zhang, Rui Cao, and Sai Wu.
\newblock Information fusion in visual question answering: A survey.
\newblock \emph{Information Fusion}, 52:\penalty0 268--280, 2019.

\bibitem[Kallooriyakath et~al.(2020)Kallooriyakath, Jithin, Bindu, and
  Adith]{kallooriyakath2020visual}
Liyana~Sahir Kallooriyakath, MV~Jithin, PV~Bindu, and PP~Adith.
\newblock Visual question answering: methodologies and challenges.
\newblock In \emph{2020 International Conference on Smart Technologies in
  Computing, Electrical and Electronics (ICSTCEE)}, pages 402--407. IEEE, 2020.

\bibitem[Manmadhan and Kovoor(2020)]{manmadhan2020visual}
Sruthy Manmadhan and Binsu~C Kovoor.
\newblock Visual question answering: a state-of-the-art review.
\newblock \emph{Artificial Intelligence Review}, 53:\penalty0 5705--5745, 2020.

\bibitem[Zou and Xie(2020)]{zou2020survey}
Yeyun Zou and Qiyu Xie.
\newblock A survey on vqa: Datasets and approaches.
\newblock In \emph{2020 2nd International Conference on Information Technology
  and Computer Application (ITCA)}, pages 289--297. IEEE, 2020.

\bibitem[Sharma and Jalal(2021)]{sharma2021survey}
Himanshu Sharma and Anand~Singh Jalal.
\newblock A survey of methods, datasets and evaluation metrics for visual
  question answering.
\newblock \emph{Image and Vision Computing}, 116:\penalty0 104327, 2021.

\bibitem[Lu et~al.(2019)Lu, Batra, Parikh, and Lee]{lu2019vilbert}
Jiasen Lu, Dhruv Batra, Devi Parikh, and Stefan Lee.
\newblock Vilbert: Pretraining task-agnostic visiolinguistic representations
  for vision-and-language tasks.
\newblock \emph{Advances in neural information processing systems}, 32, 2019.

\bibitem[Li et~al.(2019{\natexlab{a}})Li, Yatskar, Yin, Hsieh, and
  Chang]{li2019visualbert}
Liunian~Harold Li, Mark Yatskar, Da~Yin, Cho-Jui Hsieh, and Kai-Wei Chang.
\newblock Visualbert: A simple and performant baseline for vision and language.
\newblock \emph{arXiv preprint arXiv:1908.03557}, 2019{\natexlab{a}}.

\bibitem[Su et~al.(2019)Su, Zhu, Cao, Li, Lu, Wei, and Dai]{su2019vl}
Weijie Su, Xizhou Zhu, Yue Cao, Bin Li, Lewei Lu, Furu Wei, and Jifeng Dai.
\newblock Vl-bert: Pre-training of generic visual-linguistic representations.
\newblock \emph{arXiv preprint arXiv:1908.08530}, 2019.

\bibitem[Singh et~al.(2019)Singh, Natarajan, Shah, Jiang, Chen, Batra, Parikh,
  and Rohrbach]{singh2019towards}
Amanpreet Singh, Vivek Natarajan, Meet Shah, Yu~Jiang, Xinlei Chen, Dhruv
  Batra, Devi Parikh, and Marcus Rohrbach.
\newblock Towards vqa models that can read.
\newblock In \emph{Proceedings of the IEEE/CVF conference on computer vision
  and pattern recognition}, pages 8317--8326, 2019.

\bibitem[Wang et~al.(2022{\natexlab{a}})Wang, Gao, Hu, Selvaraju, Ramaiah, Xu,
  JaJa, and Davis]{wang2022tag}
Jun Wang, Mingfei Gao, Yuqian Hu, Ramprasaath~R Selvaraju, Chetan Ramaiah, Ran
  Xu, Joseph~F JaJa, and Larry~S Davis.
\newblock Tag: Boosting text-vqa via text-aware visual question-answer
  generation.
\newblock \emph{arXiv preprint arXiv:2208.01813}, 2022{\natexlab{a}}.

\bibitem[Mishra et~al.(2019)Mishra, Shekhar, Singh, and Chakraborty]{8978122}
Anand Mishra, Shashank Shekhar, Ajeet~Kumar Singh, and Anirban Chakraborty.
\newblock Ocr-vqa: Visual question answering by reading text in images.
\newblock In \emph{2019 International Conference on Document Analysis and
  Recognition (ICDAR)}, pages 947--952, 2019.
\newblock \doi{10.1109/ICDAR.2019.00156}.

\bibitem[Tanaka et~al.(2021)Tanaka, Nishida, and Yoshida]{tanaka2021visualmrc}
Ryota Tanaka, Kyosuke Nishida, and Sen Yoshida.
\newblock Visualmrc: Machine reading comprehension on document images.
\newblock In \emph{Proceedings of the AAAI Conference on Artificial
  Intelligence}, pages 13878--13888, 2021.

\bibitem[Biten et~al.(2019)Biten, Tito, Mafla, Gomez, Rusinol, Valveny,
  Jawahar, and Karatzas]{biten2019scene}
Ali~Furkan Biten, Ruben Tito, Andres Mafla, Lluis Gomez, Mar{\c{c}}al Rusinol,
  Ernest Valveny, CV~Jawahar, and Dimosthenis Karatzas.
\newblock Scene text visual question answering.
\newblock In \emph{Proceedings of the IEEE/CVF international conference on
  computer vision}, pages 4291--4301, 2019.

\bibitem[Mathew et~al.(2021)Mathew, Karatzas, and Jawahar]{mathew2021docvqa}
Minesh Mathew, Dimosthenis Karatzas, and CV~Jawahar.
\newblock Docvqa: A dataset for vqa on document images.
\newblock In \emph{Proceedings of the IEEE/CVF winter conference on
  applications of computer vision}, pages 2200--2209, 2021.

\bibitem[Wang et~al.(2015)Wang, Wu, Shen, Hengel, and Dick]{wang2015explicit}
Peng Wang, Qi~Wu, Chunhua Shen, Anton van~den Hengel, and Anthony Dick.
\newblock Explicit knowledge-based reasoning for visual question answering.
\newblock \emph{arXiv preprint arXiv:1511.02570}, 2015.

\bibitem[Wu et~al.(2016)Wu, Wang, Shen, Dick, and Van Den~Hengel]{wu2016ask}
Qi~Wu, Peng Wang, Chunhua Shen, Anthony Dick, and Anton Van Den~Hengel.
\newblock Ask me anything: Free-form visual question answering based on
  knowledge from external sources.
\newblock In \emph{Proceedings of the IEEE conference on computer vision and
  pattern recognition}, pages 4622--4630, 2016.

\bibitem[Marino et~al.(2019)Marino, Rastegari, Farhadi, and
  Mottaghi]{marino2019ok}
Kenneth Marino, Mohammad Rastegari, Ali Farhadi, and Roozbeh Mottaghi.
\newblock Ok-vqa: A visual question answering benchmark requiring external
  knowledge.
\newblock In \emph{Proceedings of the IEEE/cvf conference on computer vision
  and pattern recognition}, pages 3195--3204, 2019.

\bibitem[Schwenk et~al.(2022)Schwenk, Khandelwal, Clark, Marino, and
  Mottaghi]{schwenk2022okvqa}
Dustin Schwenk, Apoorv Khandelwal, Christopher Clark, Kenneth Marino, and
  Roozbeh Mottaghi.
\newblock A-okvqa: A benchmark for visual question answering using world
  knowledge.
\newblock In \emph{Computer Vision--ECCV 2022: 17th European Conference, Tel
  Aviv, Israel, October 23--27, 2022, Proceedings, Part VIII}, pages 146--162.
  Springer, 2022.

\bibitem[Chen et~al.(2022{\natexlab{a}})Chen, Anjum, and
  Gurari]{chen2022grounding}
Chongyan Chen, Samreen Anjum, and Danna Gurari.
\newblock Grounding answers for visual questions asked by visually impaired
  people.
\newblock In \emph{Proceedings of the IEEE/CVF Conference on Computer Vision
  and Pattern Recognition}, pages 19098--19107, 2022{\natexlab{a}}.

\bibitem[Ustalov et~al.(2023)Ustalov, Pavlichenko, Likhobaba, and
  Smirnova]{TolokaWSDMCup2023}
Dmitry Ustalov, Nikita Pavlichenko, Daniil Likhobaba, and Alisa Smirnova.
\newblock {WSDM~Cup 2023 Challenge on Visual Question Answering}.
\newblock In \emph{Proceedings of the 4th Crowd Science Workshop on
  Collaboration of Humans and Learning Algorithms for Data Labeling}, pages
  1--7, Singapore, 2023.
\newblock URL \url{http://ceur-ws.org/Vol-3357/invited1.pdf}.

\bibitem[Singh et~al.(2022)Singh, Hu, Goswami, Couairon, Galuba, Rohrbach, and
  Kiela]{singh2022flava}
Amanpreet Singh, Ronghang Hu, Vedanuj Goswami, Guillaume Couairon, Wojciech
  Galuba, Marcus Rohrbach, and Douwe Kiela.
\newblock Flava: A foundational language and vision alignment model.
\newblock In \emph{Proceedings of the IEEE/CVF Conference on Computer Vision
  and Pattern Recognition}, pages 15638--15650, 2022.

\bibitem[Kwon et~al.(2022)Kwon, Cai, Ravichandran, Bas, Bhotika, and
  Soatto]{kwon2022masked}
Gukyeong Kwon, Zhaowei Cai, Avinash Ravichandran, Erhan Bas, Rahul Bhotika, and
  Stefano Soatto.
\newblock Masked vision and language modeling for multi-modal representation
  learning.
\newblock \emph{arXiv preprint arXiv:2208.02131}, 2022.

\bibitem[Assran et~al.(2023)Assran, Duval, Misra, Bojanowski, Vincent, Rabbat,
  LeCun, and Ballas]{assran2023self}
Mahmoud Assran, Quentin Duval, Ishan Misra, Piotr Bojanowski, Pascal Vincent,
  Michael Rabbat, Yann LeCun, and Nicolas Ballas.
\newblock Self-supervised learning from images with a joint-embedding
  predictive architecture.
\newblock In \emph{Proceedings of the IEEE/CVF Conference on Computer Vision
  and Pattern Recognition}, pages 15619--15629, 2023.

\bibitem[He et~al.(2022)He, Chen, Xie, Li, Doll{\'a}r, and
  Girshick]{he2022masked}
Kaiming He, Xinlei Chen, Saining Xie, Yanghao Li, Piotr Doll{\'a}r, and Ross
  Girshick.
\newblock Masked autoencoders are scalable vision learners.
\newblock In \emph{Proceedings of the IEEE/CVF conference on computer vision
  and pattern recognition}, pages 16000--16009, 2022.

\bibitem[Radford et~al.(2021)Radford, Kim, Hallacy, Ramesh, Goh, Agarwal,
  Sastry, Askell, Mishkin, Clark, et~al.]{radford2021learning}
Alec Radford, Jong~Wook Kim, Chris Hallacy, Aditya Ramesh, Gabriel Goh,
  Sandhini Agarwal, Girish Sastry, Amanda Askell, Pamela Mishkin, Jack Clark,
  et~al.
\newblock Learning transferable visual models from natural language
  supervision.
\newblock In \emph{International conference on machine learning}, pages
  8748--8763. PMLR, 2021.

\bibitem[Jia et~al.(2021)Jia, Yang, Xia, Chen, Parekh, Pham, Le, Sung, Li, and
  Duerig]{jia2021scaling}
Chao Jia, Yinfei Yang, Ye~Xia, Yi-Ting Chen, Zarana Parekh, Hieu Pham, Quoc Le,
  Yun-Hsuan Sung, Zhen Li, and Tom Duerig.
\newblock Scaling up visual and vision-language representation learning with
  noisy text supervision.
\newblock In \emph{International Conference on Machine Learning}, pages
  4904--4916. PMLR, 2021.

\bibitem[Zhai et~al.(2023)Zhai, Mustafa, Kolesnikov, and
  Beyer]{zhai2023sigmoid}
Xiaohua Zhai, Basil Mustafa, Alexander Kolesnikov, and Lucas Beyer.
\newblock Sigmoid loss for language image pre-training, 2023.

\bibitem[Team(2024)]{chameleonteam2024chameleonmixedmodalearlyfusionfoundation}
Chameleon Team.
\newblock Chameleon: Mixed-modal early-fusion foundation models, 2024.
\newblock URL \url{https://arxiv.org/abs/2405.09818}.

\bibitem[OpenAI(2023)]{openai2023gpt4}
OpenAI.
\newblock Gpt-4 technical report, 2023.

\bibitem[Wang et~al.(2022{\natexlab{b}})Wang, Yang, Men, Lin, Bai, Li, Ma,
  Zhou, Zhou, and Yang]{wang2022ofa}
Peng Wang, An~Yang, Rui Men, Junyang Lin, Shuai Bai, Zhikang Li, Jianxin Ma,
  Chang Zhou, Jingren Zhou, and Hongxia Yang.
\newblock Ofa: Unifying architectures, tasks, and modalities through a simple
  sequence-to-sequence learning framework.
\newblock In \emph{International Conference on Machine Learning}, pages
  23318--23340. PMLR, 2022{\natexlab{b}}.

\bibitem[Liu et~al.(2023{\natexlab{a}})Liu, Li, Wu, and Lee]{liu2023llava}
Haotian Liu, Chunyuan Li, Qingyang Wu, and Yong~Jae Lee.
\newblock Visual instruction tuning, 2023{\natexlab{a}}.

\bibitem[Tsimpoukelli et~al.(2021)Tsimpoukelli, Menick, Cabi, Eslami, Vinyals,
  and Hill]{tsimpoukelli2021multimodalfewshotlearningfrozen}
Maria Tsimpoukelli, Jacob Menick, Serkan Cabi, S.~M.~Ali Eslami, Oriol Vinyals,
  and Felix Hill.
\newblock Multimodal few-shot learning with frozen language models, 2021.
\newblock URL \url{https://arxiv.org/abs/2106.13884}.

\bibitem[Li et~al.(2023{\natexlab{a}})Li, Li, Savarese, and
  Hoi]{li2023blip2bootstrappinglanguageimagepretraining}
Junnan Li, Dongxu Li, Silvio Savarese, and Steven Hoi.
\newblock Blip-2: Bootstrapping language-image pre-training with frozen image
  encoders and large language models, 2023{\natexlab{a}}.
\newblock URL \url{https://arxiv.org/abs/2301.12597}.

\bibitem[Bai et~al.(2023)Bai, Bai, Chu, Cui, Dang, Deng, Fan, Ge, Han, Huang,
  Hui, Ji, Li, Lin, Lin, Liu, Liu, Lu, Lu, Ma, Men, Ren, Ren, Tan, Tan, Tu,
  Wang, Wang, Wang, Wu, Xu, Xu, Yang, Yang, Yang, Yang, Yao, Yu, Yuan, Yuan,
  Zhang, Zhang, Zhang, Zhang, Zhou, Zhou, Zhou, and
  Zhu]{bai2023qwentechnicalreport}
Jinze Bai, Shuai Bai, Yunfei Chu, Zeyu Cui, Kai Dang, Xiaodong Deng, Yang Fan,
  Wenbin Ge, Yu~Han, Fei Huang, Binyuan Hui, Luo Ji, Mei Li, Junyang Lin, Runji
  Lin, Dayiheng Liu, Gao Liu, Chengqiang Lu, Keming Lu, Jianxin Ma, Rui Men,
  Xingzhang Ren, Xuancheng Ren, Chuanqi Tan, Sinan Tan, Jianhong Tu, Peng Wang,
  Shijie Wang, Wei Wang, Shengguang Wu, Benfeng Xu, Jin Xu, An~Yang, Hao Yang,
  Jian Yang, Shusheng Yang, Yang Yao, Bowen Yu, Hongyi Yuan, Zheng Yuan,
  Jianwei Zhang, Xingxuan Zhang, Yichang Zhang, Zhenru Zhang, Chang Zhou,
  Jingren Zhou, Xiaohuan Zhou, and Tianhang Zhu.
\newblock Qwen technical report, 2023.
\newblock URL \url{https://arxiv.org/abs/2309.16609}.

\bibitem[Santoro et~al.(2017)Santoro, Raposo, Barrett, Malinowski, Pascanu,
  Battaglia, and Lillicrap]{santoro2017simple}
Adam Santoro, David Raposo, David~G Barrett, Mateusz Malinowski, Razvan
  Pascanu, Peter Battaglia, and Timothy Lillicrap.
\newblock A simple neural network module for relational reasoning.
\newblock \emph{Advances in neural information processing systems}, 30, 2017.

\bibitem[Teney et~al.(2017)Teney, Liu, and van Den~Hengel]{teney2017graph}
Damien Teney, Lingqiao Liu, and Anton van Den~Hengel.
\newblock Graph-structured representations for visual question answering.
\newblock In \emph{Proceedings of the IEEE conference on computer vision and
  pattern recognition}, pages 1--9, 2017.

\bibitem[Fukui et~al.(2016)Fukui, Park, Yang, Rohrbach, Darrell, and
  Rohrbach]{fukui2016multimodal}
Akira Fukui, Dong~Huk Park, Daylen Yang, Anna Rohrbach, Trevor Darrell, and
  Marcus Rohrbach.
\newblock Multimodal compact bilinear pooling for visual question answering and
  visual grounding.
\newblock \emph{arXiv preprint arXiv:1606.01847}, 2016.

\bibitem[Ben-Younes et~al.(2017)Ben-Younes, Cadene, Cord, and
  Thome]{ben2017mutan}
Hedi Ben-Younes, R{\'e}mi Cadene, Matthieu Cord, and Nicolas Thome.
\newblock Mutan: Multimodal tucker fusion for visual question answering.
\newblock In \emph{Proceedings of the IEEE international conference on computer
  vision}, pages 2612--2620, 2017.

\bibitem[Andreas et~al.(2016)Andreas, Rohrbach, Darrell, and
  Klein]{andreas2016neural}
Jacob Andreas, Marcus Rohrbach, Trevor Darrell, and Dan Klein.
\newblock Neural module networks.
\newblock In \emph{Proceedings of the IEEE conference on computer vision and
  pattern recognition}, pages 39--48, 2016.

\bibitem[Hu et~al.(2017)Hu, Andreas, Rohrbach, Darrell, and
  Saenko]{hu2017learning}
Ronghang Hu, Jacob Andreas, Marcus Rohrbach, Trevor Darrell, and Kate Saenko.
\newblock Learning to reason: End-to-end module networks for visual question
  answering.
\newblock In \emph{Proceedings of the IEEE international conference on computer
  vision}, pages 804--813, 2017.

\bibitem[Yu et~al.(2019)Yu, Yu, Cui, Tao, and Tian]{yu2019deep}
Zhou Yu, Jun Yu, Yuhao Cui, Dacheng Tao, and Qi~Tian.
\newblock Deep modular co-attention networks for visual question answering.
\newblock In \emph{Proceedings of the IEEE/CVF conference on computer vision
  and pattern recognition}, pages 6281--6290, 2019.

\bibitem[Rahman et~al.(2021)Rahman, Chou, Sigal, and
  Carenini]{rahman2021improved}
Tanzila Rahman, Shih-Han Chou, Leonid Sigal, and Giuseppe Carenini.
\newblock An improved attention for visual question answering.
\newblock In \emph{Proceedings of the IEEE/CVF Conference on Computer Vision
  and Pattern Recognition}, pages 1653--1662, 2021.

\bibitem[Lu et~al.(2016)Lu, Yang, Batra, and Parikh]{lu2016hierarchical}
Jiasen Lu, Jianwei Yang, Dhruv Batra, and Devi Parikh.
\newblock Hierarchical question-image co-attention for visual question
  answering.
\newblock \emph{Advances in neural information processing systems}, 29, 2016.

\bibitem[Nam et~al.(2017)Nam, Ha, and Kim]{nam2017dual}
Hyeonseob Nam, Jung-Woo Ha, and Jeonghee Kim.
\newblock Dual attention networks for multimodal reasoning and matching.
\newblock In \emph{Proceedings of the IEEE conference on computer vision and
  pattern recognition}, pages 299--307, 2017.

\bibitem[Yang et~al.(2016)Yang, He, Gao, Deng, and Smola]{Yang_2016_CVPR}
Zichao Yang, Xiaodong He, Jianfeng Gao, Li~Deng, and Alex Smola.
\newblock Stacked attention networks for image question answering.
\newblock In \emph{Proceedings of the IEEE Conference on Computer Vision and
  Pattern Recognition (CVPR)}, June 2016.

\bibitem[Anderson et~al.(2018)Anderson, He, Buehler, Teney, Johnson, Gould, and
  Zhang]{anderson2018bottom}
Peter Anderson, Xiaodong He, Chris Buehler, Damien Teney, Mark Johnson, Stephen
  Gould, and Lei Zhang.
\newblock Bottom-up and top-down attention for image captioning and visual
  question answering.
\newblock In \emph{Proceedings of the IEEE conference on computer vision and
  pattern recognition}, pages 6077--6086, 2018.

\bibitem[Malinowski et~al.(2015)Malinowski, Rohrbach, and
  Fritz]{malinowski2015ask}
Mateusz Malinowski, Marcus Rohrbach, and Mario Fritz.
\newblock Ask your neurons: A neural-based approach to answering questions
  about images.
\newblock In \emph{Proceedings of the IEEE international conference on computer
  vision}, pages 1--9, 2015.

\bibitem[Devlin et~al.(2019)Devlin, Chang, Lee, and
  Toutanova]{devlin-etal-2019-bert}
Jacob Devlin, Ming-Wei Chang, Kenton Lee, and Kristina Toutanova.
\newblock {BERT}: Pre-training of deep bidirectional transformers for language
  understanding.
\newblock In \emph{Proceedings of the 2019 Conference of the North {A}merican
  Chapter of the Association for Computational Linguistics: Human Language
  Technologies, Volume 1 (Long and Short Papers)}, pages 4171--4186,
  Minneapolis, Minnesota, June 2019. Association for Computational Linguistics.
\newblock \doi{10.18653/v1/N19-1423}.
\newblock URL \url{https://aclanthology.org/N19-1423}.

\bibitem[Liu et~al.(2019{\natexlab{a}})Liu, Ott, Goyal, Du, Joshi, Chen, Levy,
  Lewis, Zettlemoyer, and Stoyanov]{DBLP:journals/corr/abs-1907-11692}
Yinhan Liu, Myle Ott, Naman Goyal, Jingfei Du, Mandar Joshi, Danqi Chen, Omer
  Levy, Mike Lewis, Luke Zettlemoyer, and Veselin Stoyanov.
\newblock Roberta: {A} robustly optimized {BERT} pretraining approach.
\newblock \emph{CoRR}, abs/1907.11692, 2019{\natexlab{a}}.
\newblock URL \url{http://arxiv.org/abs/1907.11692}.

\bibitem[Touvron et~al.(2023{\natexlab{a}})Touvron, Lavril, Izacard, Martinet,
  Lachaux, Lacroix, Rozi{\`e}re, Goyal, Hambro, Azhar,
  et~al.]{touvron2023llama}
Hugo Touvron, Thibaut Lavril, Gautier Izacard, Xavier Martinet, Marie-Anne
  Lachaux, Timoth{\'e}e Lacroix, Baptiste Rozi{\`e}re, Naman Goyal, Eric
  Hambro, Faisal Azhar, et~al.
\newblock Llama: Open and efficient foundation language models.
\newblock \emph{arXiv preprint arXiv:2302.13971}, 2023{\natexlab{a}}.

\bibitem[Chiang et~al.(2023)Chiang, Li, Lin, Sheng, Wu, Zhang, Zheng, Zhuang,
  Zhuang, Gonzalez, Stoica, and Xing]{vicuna2023}
Wei-Lin Chiang, Zhuohan Li, Zi~Lin, Ying Sheng, Zhanghao Wu, Hao Zhang, Lianmin
  Zheng, Siyuan Zhuang, Yonghao Zhuang, Joseph~E. Gonzalez, Ion Stoica, and
  Eric~P. Xing.
\newblock Vicuna: An open-source chatbot impressing gpt-4 with 90\%* chatgpt
  quality, March 2023.
\newblock URL \url{https://lmsys.org/blog/2023-03-30-vicuna/}.

\bibitem[Schmidhuber et~al.(1997)Schmidhuber, Hochreiter,
  et~al.]{schmidhuber1997long}
J{\"u}rgen Schmidhuber, Sepp Hochreiter, et~al.
\newblock Long short-term memory.
\newblock \emph{Neural Comput}, 9\penalty0 (8):\penalty0 1735--1780, 1997.

\bibitem[Schuster and Paliwal(1997)]{650093}
M.~Schuster and K.K. Paliwal.
\newblock Bidirectional recurrent neural networks.
\newblock \emph{IEEE Transactions on Signal Processing}, 45\penalty0
  (11):\penalty0 2673--2681, 1997.
\newblock \doi{10.1109/78.650093}.

\bibitem[Cho et~al.(2014)Cho, van Merrienboer, G{\"{u}}l{\c{c}}ehre, Bougares,
  Schwenk, and Bengio]{DBLP:journals/corr/ChoMGBSB14}
Kyunghyun Cho, Bart van Merrienboer, {\c{C}}aglar G{\"{u}}l{\c{c}}ehre, Fethi
  Bougares, Holger Schwenk, and Yoshua Bengio.
\newblock Learning phrase representations using {RNN} encoder-decoder for
  statistical machine translation.
\newblock \emph{CoRR}, abs/1406.1078, 2014.
\newblock URL \url{http://arxiv.org/abs/1406.1078}.

\bibitem[Mikolov et~al.(2013)Mikolov, Chen, Corrado, and
  Dean]{mikolov2013efficient}
Tomas Mikolov, Kai Chen, Greg Corrado, and Jeffrey Dean.
\newblock Efficient estimation of word representations in vector space.
\newblock \emph{arXiv preprint arXiv:1301.3781}, 2013.

\bibitem[Pennington et~al.(2014)Pennington, Socher, and
  Manning]{pennington-etal-2014-glove}
Jeffrey Pennington, Richard Socher, and Christopher Manning.
\newblock {G}lo{V}e: Global vectors for word representation.
\newblock In \emph{Proceedings of the 2014 Conference on Empirical Methods in
  Natural Language Processing ({EMNLP})}, pages 1532--1543, Doha, Qatar,
  October 2014. Association for Computational Linguistics.
\newblock \doi{10.3115/v1/D14-1162}.
\newblock URL \url{https://aclanthology.org/D14-1162}.

\bibitem[Dosovitskiy et~al.(2020)Dosovitskiy, Beyer, Kolesnikov, Weissenborn,
  Zhai, Unterthiner, Dehghani, Minderer, Heigold, Gelly,
  et~al.]{dosovitskiy2020image}
Alexey Dosovitskiy, Lucas Beyer, Alexander Kolesnikov, Dirk Weissenborn,
  Xiaohua Zhai, Thomas Unterthiner, Mostafa Dehghani, Matthias Minderer, Georg
  Heigold, Sylvain Gelly, et~al.
\newblock An image is worth 16x16 words: Transformers for image recognition at
  scale.
\newblock \emph{arXiv preprint arXiv:2010.11929}, 2020.

\bibitem[Redmon et~al.(2016)Redmon, Divvala, Girshick, and
  Farhadi]{Redmon_2016_CVPR}
Joseph Redmon, Santosh Divvala, Ross Girshick, and Ali Farhadi.
\newblock You only look once: Unified, real-time object detection.
\newblock In \emph{Proceedings of the IEEE Conference on Computer Vision and
  Pattern Recognition (CVPR)}, June 2016.

\bibitem[Ren et~al.(2015{\natexlab{b}})Ren, He, Girshick, and
  Sun]{NIPS2015_14bfa6bb}
Shaoqing Ren, Kaiming He, Ross Girshick, and Jian Sun.
\newblock Faster r-cnn: Towards real-time object detection with region proposal
  networks.
\newblock In C.~Cortes, N.~Lawrence, D.~Lee, M.~Sugiyama, and R.~Garnett,
  editors, \emph{Advances in Neural Information Processing Systems}, volume~28.
  Curran Associates, Inc., 2015{\natexlab{b}}.
\newblock URL
  \url{https://proceedings.neurips.cc/paper/2015/file/14bfa6bb14875e45bba028a21ed38046-Paper.pdf}.

\bibitem[He et~al.(2015)He, Zhang, Ren, and Sun]{7005506}
Kaiming He, Xiangyu Zhang, Shaoqing Ren, and Jian Sun.
\newblock Spatial pyramid pooling in deep convolutional networks for visual
  recognition.
\newblock \emph{IEEE Transactions on Pattern Analysis and Machine
  Intelligence}, 37\penalty0 (9):\penalty0 1904--1916, 2015.
\newblock \doi{10.1109/TPAMI.2015.2389824}.

\bibitem[LeCun et~al.(1989)LeCun, Boser, Denker, Henderson, Howard, Hubbard,
  and Jackel]{6795724}
Y.~LeCun, B.~Boser, J.~S. Denker, D.~Henderson, R.~E. Howard, W.~Hubbard, and
  L.~D. Jackel.
\newblock Backpropagation applied to handwritten zip code recognition.
\newblock \emph{Neural Computation}, 1\penalty0 (4):\penalty0 541--551, 1989.
\newblock \doi{10.1162/neco.1989.1.4.541}.

\bibitem[Deng et~al.(2009)Deng, Dong, Socher, Li, Li, and Fei-Fei]{5206848}
Jia Deng, Wei Dong, Richard Socher, Li-Jia Li, Kai Li, and Li~Fei-Fei.
\newblock Imagenet: A large-scale hierarchical image database.
\newblock In \emph{2009 IEEE Conference on Computer Vision and Pattern
  Recognition}, pages 248--255, 2009.
\newblock \doi{10.1109/CVPR.2009.5206848}.

\bibitem[Simonyan and Zisserman(2015)]{DBLP:journals/corr/SimonyanZ14a}
Karen Simonyan and Andrew Zisserman.
\newblock Very deep convolutional networks for large-scale image recognition.
\newblock In Yoshua Bengio and Yann LeCun, editors, \emph{3rd International
  Conference on Learning Representations, {ICLR} 2015, San Diego, CA, USA, May
  7-9, 2015, Conference Track Proceedings}, 2015.
\newblock URL \url{http://arxiv.org/abs/1409.1556}.

\bibitem[He et~al.(2016)He, Zhang, Ren, and Sun]{7780459}
Kaiming He, Xiangyu Zhang, Shaoqing Ren, and Jian Sun.
\newblock Deep residual learning for image recognition.
\newblock In \emph{2016 IEEE Conference on Computer Vision and Pattern
  Recognition (CVPR)}, pages 770--778, 2016.
\newblock \doi{10.1109/CVPR.2016.90}.

\bibitem[Szegedy et~al.(2015)Szegedy, Liu, Jia, Sermanet, Reed, Anguelov,
  Erhan, Vanhoucke, and Rabinovich]{7298594}
Christian Szegedy, Wei Liu, Yangqing Jia, Pierre Sermanet, Scott Reed, Dragomir
  Anguelov, Dumitru Erhan, Vincent Vanhoucke, and Andrew Rabinovich.
\newblock Going deeper with convolutions.
\newblock In \emph{2015 IEEE Conference on Computer Vision and Pattern
  Recognition (CVPR)}, pages 1--9, 2015.
\newblock \doi{10.1109/CVPR.2015.7298594}.

\bibitem[Jiang et~al.(2020)Jiang, Misra, Rohrbach, Learned-Miller, and
  Chen]{Jiang_2020_CVPR}
Huaizu Jiang, Ishan Misra, Marcus Rohrbach, Erik Learned-Miller, and Xinlei
  Chen.
\newblock In defense of grid features for visual question answering.
\newblock In \emph{Proceedings of the IEEE/CVF Conference on Computer Vision
  and Pattern Recognition (CVPR)}, June 2020.

\bibitem[Banerjee et~al.(2021)Banerjee, Gokhale, Yang, and
  Baral]{banerjee-etal-2021-weaqa}
Pratyay Banerjee, Tejas Gokhale, Yezhou Yang, and Chitta Baral.
\newblock {W}ea{QA}: Weak supervision via captions for visual question
  answering.
\newblock In \emph{Findings of the Association for Computational Linguistics:
  ACL-IJCNLP 2021}, pages 3420--3435, Online, August 2021. Association for
  Computational Linguistics.
\newblock \doi{10.18653/v1/2021.findings-acl.302}.
\newblock URL \url{https://aclanthology.org/2021.findings-acl.302}.

\bibitem[Yu et~al.(2023{\natexlab{a}})Yu, Li, Shi, Li, and Xiao]{YU2023119148}
Yonglin Yu, Haifeng Li, Hanrong Shi, Lin Li, and Jun Xiao.
\newblock Question-guided feature pyramid network for medical visual question
  answering.
\newblock \emph{Expert Systems with Applications}, 214:\penalty0 119148,
  2023{\natexlab{a}}.
\newblock ISSN 0957-4174.
\newblock \doi{https://doi.org/10.1016/j.eswa.2022.119148}.
\newblock URL
  \url{https://www.sciencedirect.com/science/article/pii/S0957417422021662}.

\bibitem[Jaderberg et~al.(2015)Jaderberg, Simonyan, Zisserman, and
  kavukcuoglu]{NIPS2015_33ceb07b}
Max Jaderberg, Karen Simonyan, Andrew Zisserman, and koray kavukcuoglu.
\newblock Spatial transformer networks.
\newblock In C.~Cortes, N.~Lawrence, D.~Lee, M.~Sugiyama, and R.~Garnett,
  editors, \emph{Advances in Neural Information Processing Systems}, volume~28.
  Curran Associates, Inc., 2015.
\newblock URL
  \url{https://proceedings.neurips.cc/paper_files/paper/2015/file/33ceb07bf4eeb3da587e268d663aba1a-Paper.pdf}.

\bibitem[Burt et~al.(2017)Burt, Cudic, and Principe]{7965954}
Ryan Burt, Mihael Cudic, and Jose~C. Principe.
\newblock Fusing attention with visual question answering.
\newblock In \emph{2017 International Joint Conference on Neural Networks
  (IJCNN)}, pages 949--953, 2017.
\newblock \doi{10.1109/IJCNN.2017.7965954}.

\bibitem[Guo et~al.(2023)Guo, Li, Li, Tiong, Li, Tao, and HOI]{guo2023from}
Jiaxian Guo, Junnan Li, Dongxu Li, Anthony Tiong, Boyang Li, Dacheng Tao, and
  Steven HOI.
\newblock From images to textual prompts: Zero-shot {VQA} with frozen large
  language models, 2023.
\newblock URL \url{https://openreview.net/forum?id=Ck1UtnVukP8}.

\bibitem[Lin et~al.(2017)Lin, Goyal, Girshick, He, and Dollar]{Lin_2017_ICCV}
Tsung-Yi Lin, Priya Goyal, Ross Girshick, Kaiming He, and Piotr Dollar.
\newblock Focal loss for dense object detection.
\newblock In \emph{Proceedings of the IEEE International Conference on Computer
  Vision (ICCV)}, Oct 2017.

\bibitem[Kim et~al.(2021)Kim, Son, and Kim]{kim2021vilt}
Wonjae Kim, Bokyung Son, and Ildoo Kim.
\newblock Vilt: Vision-and-language transformer without convolution or region
  supervision.
\newblock In \emph{International Conference on Machine Learning}, pages
  5583--5594. PMLR, 2021.

\bibitem[Bao et~al.(2021)Bao, Wang, Dong, Liu, Mohammed, Aggarwal, Som, and
  Wei]{bao2021vlmo}
Hangbo Bao, Wenhui Wang, Li~Dong, Qiang Liu, Owais~Khan Mohammed, Kriti
  Aggarwal, Subhojit Som, and Furu Wei.
\newblock Vlmo: Unified vision-language pre-training with
  mixture-of-modality-experts.
\newblock \emph{arXiv preprint arXiv:2111.02358}, 2021.

\bibitem[Zeng et~al.(2022{\natexlab{a}})Zeng, Zhang, and Li]{zeng2022multi}
Yan Zeng, Xinsong Zhang, and Hang Li.
\newblock Multi-grained vision language pre-training: Aligning texts with
  visual concepts.
\newblock \emph{Proceedings of The 33rd International Conference on Machine
  Learning}, 2022{\natexlab{a}}.

\bibitem[Zeng et~al.(2022{\natexlab{b}})Zeng, Zhang, Li, Wang, Zhang, and
  Zhou]{zeng2022x}
Yan Zeng, Xinsong Zhang, Hang Li, Jiawei Wang, Jipeng Zhang, and Wangchunshu
  Zhou.
\newblock X2-vlm: All-in-one pre-trained model for vision-language tasks.
\newblock \emph{arXiv preprint arXiv:2211.12402}, 2022{\natexlab{b}}.

\bibitem[Li et~al.(2021)Li, Selvaraju, Gotmare, Joty, Xiong, and
  Hoi]{li2021align}
Junnan Li, Ramprasaath Selvaraju, Akhilesh Gotmare, Shafiq Joty, Caiming Xiong,
  and Steven Chu~Hong Hoi.
\newblock Align before fuse: Vision and language representation learning with
  momentum distillation.
\newblock \emph{Advances in neural information processing systems},
  34:\penalty0 9694--9705, 2021.

\bibitem[Kirillov et~al.(2023)Kirillov, Mintun, Ravi, Mao, Rolland, Gustafson,
  Xiao, Whitehead, Berg, Lo, Dollar, and Girshick]{Kirillov_2023_ICCV}
Alexander Kirillov, Eric Mintun, Nikhila Ravi, Hanzi Mao, Chloe Rolland, Laura
  Gustafson, Tete Xiao, Spencer Whitehead, Alexander~C. Berg, Wan-Yen Lo, Piotr
  Dollar, and Ross Girshick.
\newblock Segment anything.
\newblock In \emph{Proceedings of the IEEE/CVF International Conference on
  Computer Vision (ICCV)}, pages 4015--4026, October 2023.

\bibitem[Birkl et~al.(2023)Birkl, Wofk, and M{\"u}ller]{birkl2023midas}
Reiner Birkl, D~Wofk, and M~M{\"u}ller.
\newblock Midas v3. 1--a model zoo for robust monocular relative depth
  estimation. arxiv.
\newblock \emph{arXiv preprint arXiv:2307.14460}, 2023.

\bibitem[Lee et~al.(2024)Lee, Park, Kim, and Ro]{lee2024moai}
Byung-Kwan Lee, Beomchan Park, Chae~Won Kim, and Yong~Man Ro.
\newblock Moai: Mixture of all intelligence for large language and vision
  models.
\newblock \emph{arXiv preprint arXiv:2403.07508}, 2024.

\bibitem[Tong et~al.(2024)Tong, Brown, Wu, Woo, Middepogu, Akula, Yang, Yang,
  Iyer, Pan, Wang, Fergus, LeCun, and Xie]{tong2024cambrian1}
Shengbang Tong, Ellis Brown, Penghao Wu, Sanghyun Woo, Manoj Middepogu,
  Sai~Charitha Akula, Jihan Yang, Shusheng Yang, Adithya Iyer, Xichen Pan,
  Austin Wang, Rob Fergus, Yann LeCun, and Saining Xie.
\newblock Cambrian-1: A fully open, vision-centric exploration of multimodal
  llms, 2024.

\bibitem[Kiros et~al.(2015)Kiros, Zhu, Salakhutdinov, Zemel, Urtasun, Torralba,
  and Fidler]{kiros2015skip}
Ryan Kiros, Yukun Zhu, Russ~R Salakhutdinov, Richard Zemel, Raquel Urtasun,
  Antonio Torralba, and Sanja Fidler.
\newblock Skip-thought vectors.
\newblock \emph{Advances in neural information processing systems}, 28, 2015.

\bibitem[Vaswani et~al.(2017)Vaswani, Shazeer, Parmar, Uszkoreit, Jones, Gomez,
  Kaiser, and Polosukhin]{vaswani2017attention}
Ashish Vaswani, Noam Shazeer, Niki Parmar, Jakob Uszkoreit, Llion Jones,
  Aidan~N Gomez, {\L}ukasz Kaiser, and Illia Polosukhin.
\newblock Attention is all you need.
\newblock \emph{Advances in neural information processing systems}, 30, 2017.

\bibitem[Devlin et~al.(2018)Devlin, Chang, Lee, and Toutanova]{devlin2018bert}
Jacob Devlin, Ming-Wei Chang, Kenton Lee, and Kristina Toutanova.
\newblock Bert: Pre-training of deep bidirectional transformers for language
  understanding.
\newblock \emph{arXiv preprint arXiv:1810.04805}, 2018.

\bibitem[Liu et~al.(2019{\natexlab{b}})Liu, Ott, Goyal, Du, Joshi, Chen, Levy,
  Lewis, Zettlemoyer, and Stoyanov]{liu2019roberta}
Yinhan Liu, Myle Ott, Naman Goyal, Jingfei Du, Mandar Joshi, Danqi Chen, Omer
  Levy, Mike Lewis, Luke Zettlemoyer, and Veselin Stoyanov.
\newblock Roberta: A robustly optimized bert pretraining approach.
\newblock \emph{arXiv preprint arXiv:1907.11692}, 2019{\natexlab{b}}.

\bibitem[Xue et~al.(2020)Xue, Constant, Roberts, Kale, Al-Rfou, Siddhant,
  Barua, and Raffel]{xue2020mt5}
Linting Xue, Noah Constant, Adam Roberts, Mihir Kale, Rami Al-Rfou, Aditya
  Siddhant, Aditya Barua, and Colin Raffel.
\newblock mt5: A massively multilingual pre-trained text-to-text transformer.
\newblock \emph{arXiv preprint arXiv:2010.11934}, 2020.

\bibitem[Touvron et~al.(2023{\natexlab{b}})Touvron, Martin, Stone, Albert,
  Almahairi, Babaei, Bashlykov, Batra, Bhargava, Bhosale,
  et~al.]{touvron2023llama2}
Hugo Touvron, Louis Martin, Kevin Stone, Peter Albert, Amjad Almahairi, Yasmine
  Babaei, Nikolay Bashlykov, Soumya Batra, Prajjwal Bhargava, Shruti Bhosale,
  et~al.
\newblock Llama 2: Open foundation and fine-tuned chat models.
\newblock \emph{arXiv preprint arXiv:2307.09288}, 2023{\natexlab{b}}.

\bibitem[Jabri et~al.(2016)Jabri, Joulin, and Maaten]{jabri2016revisiting}
Allan Jabri, Armand Joulin, and Laurens van~der Maaten.
\newblock Revisiting visual question answering baselines.
\newblock In \emph{European conference on computer vision}, pages 727--739.
  Springer, 2016.

\bibitem[Saito et~al.(2017)Saito, Shin, Ushiku, and Harada]{saito2017dualnet}
Kuniaki Saito, Andrew Shin, Yoshitaka Ushiku, and Tatsuya Harada.
\newblock Dualnet: Domain-invariant network for visual question answering.
\newblock In \emph{2017 IEEE International Conference on Multimedia and Expo
  (ICME)}, pages 829--834. IEEE, 2017.

\bibitem[Gao et~al.(2015)Gao, Mao, Zhou, Huang, Wang, and Xu]{gao2015you}
Haoyuan Gao, Junhua Mao, Jie Zhou, Zhiheng Huang, Lei Wang, and Wei Xu.
\newblock Are you talking to a machine? dataset and methods for multilingual
  image question.
\newblock \emph{Advances in neural information processing systems}, 28, 2015.

\bibitem[Kafle and Kanan(2016)]{kafle2016answer}
Kushal Kafle and Christopher Kanan.
\newblock Answer-type prediction for visual question answering.
\newblock In \emph{Proceedings of the IEEE conference on computer vision and
  pattern recognition}, pages 4976--4984, 2016.

\bibitem[Bahdanau et~al.(2014)Bahdanau, Cho, and Bengio]{bahdanau2014neural}
Dzmitry Bahdanau, Kyunghyun Cho, and Yoshua Bengio.
\newblock Neural machine translation by jointly learning to align and
  translate.
\newblock \emph{arXiv preprint arXiv:1409.0473}, 2014.

\bibitem[Kim et~al.(2016{\natexlab{a}})Kim, Lee, Kwak, Heo, Kim, Ha, and
  Zhang]{kim2016multimodal}
Jin-Hwa Kim, Sang-Woo Lee, Donghyun Kwak, Min-Oh Heo, Jeonghee Kim, Jung-Woo
  Ha, and Byoung-Tak Zhang.
\newblock Multimodal residual learning for visual qa.
\newblock \emph{Advances in neural information processing systems}, 29,
  2016{\natexlab{a}}.

\bibitem[Sukhbaatar et~al.(2015)Sukhbaatar, Weston, Fergus,
  et~al.]{sukhbaatar2015end}
Sainbayar Sukhbaatar, Jason Weston, Rob Fergus, et~al.
\newblock End-to-end memory networks.
\newblock \emph{Advances in neural information processing systems}, 28, 2015.

\bibitem[Xu and Saenko(2016)]{xu2016ask}
Huijuan Xu and Kate Saenko.
\newblock Ask, attend and answer: Exploring question-guided spatial attention
  for visual question answering.
\newblock In \emph{European conference on computer vision}, pages 451--466.
  Springer, 2016.

\bibitem[Xiong et~al.(2016)Xiong, Merity, and Socher]{pmlr-v48-xiong16}
Caiming Xiong, Stephen Merity, and Richard Socher.
\newblock Dynamic memory networks for visual and textual question answering.
\newblock In Maria~Florina Balcan and Kilian~Q. Weinberger, editors,
  \emph{Proceedings of The 33rd International Conference on Machine Learning},
  volume~48 of \emph{Proceedings of Machine Learning Research}, pages
  2397--2406, New York, New York, USA, 20--22 Jun 2016. PMLR.
\newblock URL \url{https://proceedings.mlr.press/v48/xiong16.html}.

\bibitem[Kumar et~al.(2016)Kumar, Irsoy, Ondruska, Iyyer, Bradbury, Gulrajani,
  Zhong, Paulus, and Socher]{kumar2016ask}
Ankit Kumar, Ozan Irsoy, Peter Ondruska, Mohit Iyyer, James Bradbury, Ishaan
  Gulrajani, Victor Zhong, Romain Paulus, and Richard Socher.
\newblock Ask me anything: Dynamic memory networks for natural language
  processing.
\newblock In \emph{International conference on machine learning}, pages
  1378--1387. PMLR, 2016.

\bibitem[Shih et~al.(2016)Shih, Singh, and Hoiem]{shih2016look}
Kevin~J Shih, Saurabh Singh, and Derek Hoiem.
\newblock Where to look: Focus regions for visual question answering.
\newblock In \emph{Proceedings of the IEEE conference on computer vision and
  pattern recognition}, pages 4613--4621, 2016.

\bibitem[Ilievski et~al.(2016)Ilievski, Yan, and Feng]{ilievski2016focused}
Ilija Ilievski, Shuicheng Yan, and Jiashi Feng.
\newblock A focused dynamic attention model for visual question answering.
\newblock \emph{arXiv preprint arXiv:1604.01485}, 2016.

\bibitem[Teney et~al.(2018)Teney, Anderson, He, and Van
  Den~Hengel]{teney2018tips}
Damien Teney, Peter Anderson, Xiaodong He, and Anton Van Den~Hengel.
\newblock Tips and tricks for visual question answering: Learnings from the
  2017 challenge.
\newblock In \emph{Proceedings of the IEEE conference on computer vision and
  pattern recognition}, pages 4223--4232, 2018.

\bibitem[Song et~al.(2018)Song, Zeng, Gao, and Shen]{ijcai2018p126}
Jingkuan Song, Pengpeng Zeng, Lianli Gao, and Heng~Tao Shen.
\newblock From pixels to objects: Cubic visual attention for visual question
  answering.
\newblock In \emph{Proceedings of the Twenty-Seventh International Joint
  Conference on Artificial Intelligence, {IJCAI-18}}, pages 906--912.
  International Joint Conferences on Artificial Intelligence Organization, 7
  2018.
\newblock \doi{10.24963/ijcai.2018/126}.
\newblock URL \url{https://doi.org/10.24963/ijcai.2018/126}.

\bibitem[Kim et~al.(2018)Kim, Jun, and Zhang]{kim2018bilinear}
Jin-Hwa Kim, Jaehyun Jun, and Byoung-Tak Zhang.
\newblock Bilinear attention networks.
\newblock \emph{Advances in neural information processing systems}, 31, 2018.

\bibitem[Nguyen and Okatani(2018)]{nguyen2018improved}
Duy-Kien Nguyen and Takayuki Okatani.
\newblock Improved fusion of visual and language representations by dense
  symmetric co-attention for visual question answering.
\newblock In \emph{Proceedings of the IEEE conference on computer vision and
  pattern recognition}, pages 6087--6096, 2018.

\bibitem[Huang et~al.(2019)Huang, Wang, Chen, and Wei]{huang2019attention}
Lun Huang, Wenmin Wang, Jie Chen, and Xiao-Yong Wei.
\newblock Attention on attention for image captioning.
\newblock In \emph{Proceedings of the IEEE/CVF international conference on
  computer vision}, pages 4634--4643, 2019.

\bibitem[Zhou et~al.(2021)Zhou, Ren, Zhu, Sun, Liu, Ding, Xu, and
  Ji]{zhou2021trar}
Yiyi Zhou, Tianhe Ren, Chaoyang Zhu, Xiaoshuai Sun, Jianzhuang Liu, Xinghao
  Ding, Mingliang Xu, and Rongrong Ji.
\newblock Trar: Routing the attention spans in transformer for visual question
  answering.
\newblock In \emph{Proceedings of the IEEE/CVF International Conference on
  Computer Vision}, pages 2074--2084, 2021.

\bibitem[Steitz et~al.(2022)Steitz, Pfeiffer, Gurevych, and
  Roth]{steitz2022txt}
Jan-Martin~O Steitz, Jonas Pfeiffer, Iryna Gurevych, and Stefan Roth.
\newblock Txt: Crossmodal end-to-end learning with transformers.
\newblock In \emph{Pattern Recognition: 43rd DAGM German Conference, DAGM GCPR
  2021, Bonn, Germany, September 28--October 1, 2021, Proceedings}, pages
  405--420. Springer, 2022.

\bibitem[Johnson et~al.(2017{\natexlab{a}})Johnson, Hariharan, Van Der~Maaten,
  Hoffman, Fei-Fei, Lawrence~Zitnick, and Girshick]{johnson2017inferring}
Justin Johnson, Bharath Hariharan, Laurens Van Der~Maaten, Judy Hoffman,
  Li~Fei-Fei, C~Lawrence~Zitnick, and Ross Girshick.
\newblock Inferring and executing programs for visual reasoning.
\newblock In \emph{Proceedings of the IEEE international conference on computer
  vision}, pages 2989--2998, 2017{\natexlab{a}}.

\bibitem[Hu et~al.(2018)Hu, Andreas, Darrell, and Saenko]{hu2018explainable}
Ronghang Hu, Jacob Andreas, Trevor Darrell, and Kate Saenko.
\newblock Explainable neural computation via stack neural module networks.
\newblock In \emph{Proceedings of the European conference on computer vision
  (ECCV)}, pages 53--69, 2018.

\bibitem[Yi et~al.(2018)Yi, Wu, Gan, Torralba, Kohli, and
  Tenenbaum]{yi2018neural}
Kexin Yi, Jiajun Wu, Chuang Gan, Antonio Torralba, Pushmeet Kohli, and Josh
  Tenenbaum.
\newblock Neural-symbolic vqa: Disentangling reasoning from vision and language
  understanding.
\newblock \emph{Advances in neural information processing systems}, 31, 2018.

\bibitem[Vedantam et~al.(2019)Vedantam, Desai, Lee, Rohrbach, Batra, and
  Parikh]{vedantam2019probabilistic}
Ramakrishna Vedantam, Karan Desai, Stefan Lee, Marcus Rohrbach, Dhruv Batra,
  and Devi Parikh.
\newblock Probabilistic neural symbolic models for interpretable visual
  question answering.
\newblock In \emph{International Conference on Machine Learning}, pages
  6428--6437. PMLR, 2019.

\bibitem[Charikar et~al.(2002)Charikar, Chen, and
  Farach-Colton]{charikar2002finding}
Moses Charikar, Kevin Chen, and Martin Farach-Colton.
\newblock Finding frequent items in data streams.
\newblock In \emph{International Colloquium on Automata, Languages, and
  Programming}, pages 693--703. Springer, 2002.

\bibitem[Kim et~al.(2016{\natexlab{b}})Kim, On, Lim, Kim, Ha, and
  Zhang]{kim2016hadamard}
Jin-Hwa Kim, Kyoung-Woon On, Woosang Lim, Jeonghee Kim, Jung-Woo Ha, and
  Byoung-Tak Zhang.
\newblock Hadamard product for low-rank bilinear pooling.
\newblock \emph{arXiv preprint arXiv:1610.04325}, 2016{\natexlab{b}}.

\bibitem[Yu et~al.(2017)Yu, Yu, Fan, and Tao]{yu2017multi}
Zhou Yu, Jun Yu, Jianping Fan, and Dacheng Tao.
\newblock Multi-modal factorized bilinear pooling with co-attention learning
  for visual question answering.
\newblock In \emph{Proceedings of the IEEE international conference on computer
  vision}, pages 1821--1830, 2017.

\bibitem[Perez et~al.(2018)Perez, Strub, De~Vries, Dumoulin, and
  Courville]{perez2018film}
Ethan Perez, Florian Strub, Harm De~Vries, Vincent Dumoulin, and Aaron
  Courville.
\newblock Film: Visual reasoning with a general conditioning layer.
\newblock In \emph{Proceedings of the AAAI Conference on Artificial
  Intelligence}, volume~32, 2018.

\bibitem[Ben-Younes et~al.(2019)Ben-Younes, Cadene, Thome, and
  Cord]{ben2019block}
Hedi Ben-Younes, Remi Cadene, Nicolas Thome, and Matthieu Cord.
\newblock Block: Bilinear superdiagonal fusion for visual question answering
  and visual relationship detection.
\newblock In \emph{Proceedings of the AAAI conference on artificial
  intelligence}, volume~33, pages 8102--8109, 2019.

\bibitem[LeCun et~al.(2006)LeCun, Chopra, Hadsell, Ranzato, and
  Huang]{lecun2006tutorial}
Yann LeCun, Sumit Chopra, Raia Hadsell, M~Ranzato, and Fujie Huang.
\newblock A tutorial on energy-based learning.
\newblock \emph{Predicting structured data}, 1\penalty0 (0), 2006.

\bibitem[Mu et~al.(2022)Mu, Kirillov, Wagner, and Xie]{mu2022slip}
Norman Mu, Alexander Kirillov, David Wagner, and Saining Xie.
\newblock Slip: Self-supervision meets language-image pre-training.
\newblock In \emph{European conference on computer vision}, pages 529--544.
  Springer, 2022.

\bibitem[Lavoie et~al.(2024)Lavoie, Kirichenko, Ibrahim, Assran, Wilson,
  Courville, and Ballas]{lavoie2024modeling}
Samuel Lavoie, Polina Kirichenko, Mark Ibrahim, Mahmoud Assran, Andrew~Gordon
  Wilson, Aaron Courville, and Nicolas Ballas.
\newblock Modeling caption diversity in contrastive vision-language
  pretraining, 2024.

\bibitem[Tan and Bansal(2019)]{tan2019lxmert}
Hao Tan and Mohit Bansal.
\newblock Lxmert: Learning cross-modality encoder representations from
  transformers.
\newblock \emph{arXiv preprint arXiv:1908.07490}, 2019.

\bibitem[Yu et~al.(2022)Yu, Wang, Vasudevan, Yeung, Seyedhosseini, and
  Wu]{yu2022coca}
Jiahui Yu, Zirui Wang, Vijay Vasudevan, Legg Yeung, Mojtaba Seyedhosseini, and
  Yonghui Wu.
\newblock Coca: Contrastive captioners are image-text foundation models.
\newblock \emph{arXiv preprint arXiv:2205.01917}, 2022.

\bibitem[Sauer et~al.(2024)Sauer, Boesel, Dockhorn, Blattmann, Esser, and
  Rombach]{sauer2024fast}
Axel Sauer, Frederic Boesel, Tim Dockhorn, Andreas Blattmann, Patrick Esser,
  and Robin Rombach.
\newblock Fast high-resolution image synthesis with latent adversarial
  diffusion distillation.
\newblock \emph{arXiv preprint arXiv:2403.12015}, 2024.

\bibitem[Stability.ai(2024)]{Stable_Diffusion_3}
Stability.ai.
\newblock Stable diffusion 3: Research paper.
\newblock \emph{https://stability.ai/news/stable-diffusion-3-research-paper},
  2024.

\bibitem[Meta(2023)]{llama3}
Meta.
\newblock meta-llama-3, 2023.
\newblock URL \url{https://ai.meta.com/blog/meta-llama-3}.

\bibitem[TII(2024)]{falcon2}
TII.
\newblock Falcon 2.
\newblock In \emph{European conference on computer vision}, 2024.
\newblock URL \url{https://falconllm.tii.ae/}.

\bibitem[Liu et~al.(2024{\natexlab{a}})Liu, Li, Li, Li, Zhang, Shen, and
  Lee]{liu2024llavanext}
Haotian Liu, Chunyuan Li, Yuheng Li, Bo~Li, Yuanhan Zhang, Sheng Shen, and
  Yong~Jae Lee.
\newblock Llava-next: Improved reasoning, ocr, and world knowledge, January
  2024{\natexlab{a}}.
\newblock URL \url{https://llava-vl.github.io/blog/2024-01-30-llava-next/}.

\bibitem[Zhu et~al.(2023)Zhu, Chen, Shen, Li, and Elhoseiny]{zhu2023minigpt}
Deyao Zhu, Jun Chen, Xiaoqian Shen, Xiang Li, and Mohamed Elhoseiny.
\newblock Minigpt-4: Enhancing vision-language understanding with advanced
  large language models.
\newblock \emph{arXiv preprint arXiv:2304.10592}, 2023.

\bibitem[Krishna et~al.(2017)Krishna, Zhu, Groth, Johnson, Hata, Kravitz, Chen,
  Kalantidis, Li, Shamma, et~al.]{krishna2017visual}
Ranjay Krishna, Yuke Zhu, Oliver Groth, Justin Johnson, Kenji Hata, Joshua
  Kravitz, Stephanie Chen, Yannis Kalantidis, Li-Jia Li, David~A Shamma, et~al.
\newblock Visual genome: Connecting language and vision using crowdsourced
  dense image annotations.
\newblock \emph{International journal of computer vision}, 123\penalty0
  (1):\penalty0 32--73, 2017.

\bibitem[Sheng et~al.(2016)Sheng, Van~Gool, and Moens]{sheng2016dataset}
Shurong Sheng, Luc Van~Gool, and Marie-Francine Moens.
\newblock A dataset for multimodal question answering in the cultural heritage
  domain.
\newblock In \emph{Proceedings of the COLING 2016 Workshop on Language
  Technology Resources and Tools for Digital Humanities (LT4DH)}, pages 10--17.
  ACL, 2016.

\bibitem[Wang et~al.(2017)Wang, Wu, Shen, Dick, and Van
  Den~Hengel]{wang2017fvqa}
Peng Wang, Qi~Wu, Chunhua Shen, Anthony Dick, and Anton Van Den~Hengel.
\newblock Fvqa: Fact-based visual question answering.
\newblock \emph{IEEE transactions on pattern analysis and machine
  intelligence}, 40\penalty0 (10):\penalty0 2413--2427, 2017.

\bibitem[Malinowski and Fritz(2014)]{malinowski2014multi}
Mateusz Malinowski and Mario Fritz.
\newblock A multi-world approach to question answering about real-world scenes
  based on uncertain input.
\newblock \emph{Advances in neural information processing systems}, 27, 2014.

\bibitem[Zhu et~al.(2016)Zhu, Groth, Bernstein, and Fei-Fei]{zhu2016visual7w}
Yuke Zhu, Oliver Groth, Michael Bernstein, and Li~Fei-Fei.
\newblock Visual7w: Grounded question answering in images.
\newblock In \emph{Proceedings of the IEEE conference on computer vision and
  pattern recognition}, pages 4995--5004, 2016.

\bibitem[Goyal et~al.(2017{\natexlab{b}})Goyal, Khot, Summers-Stay, Batra, and
  Parikh]{goyal2017making}
Yash Goyal, Tejas Khot, Douglas Summers-Stay, Dhruv Batra, and Devi Parikh.
\newblock Making the v in vqa matter: Elevating the role of image understanding
  in visual question answering.
\newblock In \emph{Proceedings of the IEEE conference on computer vision and
  pattern recognition}, pages 6904--6913, 2017{\natexlab{b}}.

\bibitem[Johnson et~al.(2017{\natexlab{b}})Johnson, Hariharan, Van Der~Maaten,
  Fei-Fei, Lawrence~Zitnick, and Girshick]{johnson2017clevr}
Justin Johnson, Bharath Hariharan, Laurens Van Der~Maaten, Li~Fei-Fei,
  C~Lawrence~Zitnick, and Ross Girshick.
\newblock Clevr: A diagnostic dataset for compositional language and elementary
  visual reasoning.
\newblock In \emph{Proceedings of the IEEE conference on computer vision and
  pattern recognition}, pages 2901--2910, 2017{\natexlab{b}}.

\bibitem[Agrawal et~al.(2018)Agrawal, Batra, Parikh, and
  Kembhavi]{agrawal2018don}
Aishwarya Agrawal, Dhruv Batra, Devi Parikh, and Aniruddha Kembhavi.
\newblock Don't just assume; look and answer: Overcoming priors for visual
  question answering.
\newblock In \emph{Proceedings of the IEEE conference on computer vision and
  pattern recognition}, pages 4971--4980, 2018.

\bibitem[Hussain et~al.(2017)Hussain, Zhang, Zhang, Ye, Thomas, Agha, Ong, and
  Kovashka]{hussain2017automatic}
Zaeem Hussain, Mingda Zhang, Xiaozhong Zhang, Keren Ye, Christopher Thomas,
  Zuha Agha, Nathan Ong, and Adriana Kovashka.
\newblock Automatic understanding of image and video advertisements.
\newblock In \emph{Proceedings of the IEEE conference on computer vision and
  pattern recognition}, pages 1705--1715, 2017.

\bibitem[Kembhavi et~al.(2017{\natexlab{b}})Kembhavi, Seo, Schwenk, Choi,
  Farhadi, and Hajishirzi]{kembhavi2017you}
Aniruddha Kembhavi, Minjoon Seo, Dustin Schwenk, Jonghyun Choi, Ali Farhadi,
  and Hannaneh Hajishirzi.
\newblock Are you smarter than a sixth grader? textbook question answering for
  multimodal machine comprehension.
\newblock In \emph{Proceedings of the IEEE Conference on Computer Vision and
  Pattern recognition}, pages 4999--5007, 2017{\natexlab{b}}.

\bibitem[Kahou et~al.(2017)Kahou, Michalski, Atkinson, K{\'a}d{\'a}r,
  Trischler, and Bengio]{kahou2017figureqa}
Samira~Ebrahimi Kahou, Vincent Michalski, Adam Atkinson, {\'A}kos
  K{\'a}d{\'a}r, Adam Trischler, and Yoshua Bengio.
\newblock Figureqa: An annotated figure dataset for visual reasoning.
\newblock \emph{arXiv preprint arXiv:1710.07300}, 2017.

\bibitem[Hasan et~al.(2018{\natexlab{b}})Hasan, Ling, Farri, Liu, M{\"u}ller,
  and Lungren]{hasan2018overview}
Sadid~A Hasan, Yuan Ling, Oladimeji Farri, Joey Liu, Henning M{\"u}ller, and
  Matthew~P Lungren.
\newblock Overview of imageclef 2018 medical domain visual question answering
  task.
\newblock In \emph{CLEF (Working Notes)}, 2018{\natexlab{b}}.

\bibitem[Gurari et~al.(2018{\natexlab{a}})Gurari, Li, Stangl, Guo, Lin,
  Grauman, Luo, and Bigham]{8578478}
Danna Gurari, Qing Li, Abigale~J. Stangl, Anhong Guo, Chi Lin, Kristen Grauman,
  Jiebo Luo, and Jeffrey~P. Bigham.
\newblock Vizwiz grand challenge: Answering visual questions from blind people.
\newblock In \emph{2018 IEEE/CVF Conference on Computer Vision and Pattern
  Recognition}, pages 3608--3617, 2018{\natexlab{a}}.
\newblock \doi{10.1109/CVPR.2018.00380}.

\bibitem[Acharya et~al.(2019)Acharya, Kafle, and Kanan]{acharya2019tallyqa}
Manoj Acharya, Kushal Kafle, and Christopher Kanan.
\newblock Tallyqa: Answering complex counting questions.
\newblock In \emph{Proceedings of the AAAI conference on artificial
  intelligence}, volume~33, pages 8076--8084, 2019.

\bibitem[Kafle et~al.(2018)Kafle, Price, Cohen, and Kanan]{kafle2018dvqa}
Kushal Kafle, Brian Price, Scott Cohen, and Christopher Kanan.
\newblock Dvqa: Understanding data visualizations via question answering.
\newblock In \emph{Proceedings of the IEEE conference on computer vision and
  pattern recognition}, pages 5648--5656, 2018.

\bibitem[Abacha et~al.(2019)Abacha, Hasan, Datla, Liu, Demner-Fushman, and
  M{\"u}ller]{abacha2019vqa}
Asma~Ben Abacha, Sadid~A Hasan, Vivek~V Datla, Joey Liu, Dina Demner-Fushman,
  and Henning M{\"u}ller.
\newblock Vqa-med: Overview of the medical visual question answering task at
  imageclef 2019.
\newblock \emph{CLEF (working notes)}, 2\penalty0 (6), 2019.

\bibitem[Wang et~al.(2020)Wang, Liu, Shen, Ng, Luo, Jin, Chan, Hengel, and
  Wang]{wang2020general}
Xinyu Wang, Yuliang Liu, Chunhua Shen, Chun~Chet Ng, Canjie Luo, Lianwen Jin,
  Chee~Seng Chan, Anton van~den Hengel, and Liangwei Wang.
\newblock On the general value of evidence, and bilingual scene-text visual
  question answering.
\newblock In \emph{Proceedings of the IEEE/CVF Conference on Computer Vision
  and Pattern Recognition}, pages 10126--10135, 2020.

\bibitem[Hudson and Manning(2019{\natexlab{a}})]{hudson2019gqa}
Drew~A Hudson and Christopher~D Manning.
\newblock Gqa: A new dataset for real-world visual reasoning and compositional
  question answering.
\newblock In \emph{Proceedings of the IEEE/CVF conference on computer vision
  and pattern recognition}, pages 6700--6709, 2019{\natexlab{a}}.

\bibitem[Chaudhry et~al.(2020)Chaudhry, Shekhar, Gupta, Maneriker, Bansal, and
  Joshi]{chaudhry2020leaf}
Ritwick Chaudhry, Sumit Shekhar, Utkarsh Gupta, Pranav Maneriker, Prann Bansal,
  and Ajay Joshi.
\newblock Leaf-qa: Locate, encode \& attend for figure question answering.
\newblock In \emph{Proceedings of the IEEE/CVF Winter Conference on
  Applications of Computer Vision}, pages 3512--3521, 2020.

\bibitem[Zellers et~al.(2019)Zellers, Bisk, Farhadi, and
  Choi]{Zellers_2019_CVPR}
Rowan Zellers, Yonatan Bisk, Ali Farhadi, and Yejin Choi.
\newblock From recognition to cognition: Visual commonsense reasoning.
\newblock In \emph{Proceedings of the IEEE/CVF Conference on Computer Vision
  and Pattern Recognition (CVPR)}, June 2019.

\bibitem[Garcia et~al.(2020)Garcia, Ye, Liu, Hu, Otani, Chu, Nakashima, and
  Mitamura]{garcia2020dataset}
Noa Garcia, Chentao Ye, Zihua Liu, Qingtao Hu, Mayu Otani, Chenhui Chu, Yuta
  Nakashima, and Teruko Mitamura.
\newblock A dataset and baselines for visual question answering on art.
\newblock In \emph{Computer Vision--ECCV 2020 Workshops: Glasgow, UK, August
  23--28, 2020, Proceedings, Part II 16}, pages 92--108. Springer, 2020.

\bibitem[Abacha et~al.(2020)Abacha, Datla, Hasan, Demner-Fushman, and
  M{\"u}ller]{abacha2020overview}
Asma~Ben Abacha, Vivek~V Datla, Sadid~A Hasan, Dina Demner-Fushman, and Henning
  M{\"u}ller.
\newblock Overview of the vqa-med task at imageclef 2020: Visual question
  answering and generation in the medical domain.
\newblock In \emph{CLEF (Working Notes)}, 2020.

\bibitem[Kovaleva et~al.(2020)Kovaleva, Shivade, Kashyap, Kanjaria, Wu, Ballah,
  Coy, Karargyris, Guo, Beymer, et~al.]{kovaleva2020towards}
Olga Kovaleva, Chaitanya Shivade, Satyananda Kashyap, Karina Kanjaria, Joy Wu,
  Deddeh Ballah, Adam Coy, Alexandros Karargyris, Yufan Guo, David~Beymer
  Beymer, et~al.
\newblock Towards visual dialog for radiology.
\newblock In \emph{Proceedings of the 19th SIGBioMed Workshop on Biomedical
  Language Processing}, pages 60--69, 2020.

\bibitem[He et~al.(2020{\natexlab{a}})He, Zhang, Mou, Xing, and
  Xie]{he2020pathvqa}
Xuehai He, Yichen Zhang, Luntian Mou, Eric Xing, and Pengtao Xie.
\newblock Pathvqa: 30000+ questions for medical visual question answering.
\newblock \emph{arXiv preprint arXiv:2003.10286}, 2020{\natexlab{a}}.

\bibitem[Methani et~al.(2020)Methani, Ganguly, Khapra, and
  Kumar]{Methani_2020_WACV}
Nitesh Methani, Pritha Ganguly, Mitesh~M. Khapra, and Pratyush Kumar.
\newblock Plotqa: Reasoning over scientific plots.
\newblock In \emph{Proceedings of the IEEE/CVF Winter Conference on
  Applications of Computer Vision (WACV)}, March 2020.

\bibitem[Kiela et~al.(2020)Kiela, Firooz, Mohan, Goswami, Singh, Ringshia, and
  Testuggine]{kiela2020hateful}
Douwe Kiela, Hamed Firooz, Aravind Mohan, Vedanuj Goswami, Amanpreet Singh,
  Pratik Ringshia, and Davide Testuggine.
\newblock The hateful memes challenge: Detecting hate speech in multimodal
  memes.
\newblock \emph{Advances in neural information processing systems},
  33:\penalty0 2611--2624, 2020.

\bibitem[Liu et~al.(2021{\natexlab{a}})Liu, Zhan, Xu, Ma, Yang, and
  Wu]{liu2021slake}
Bo~Liu, Li-Ming Zhan, Li~Xu, Lin Ma, Yan Yang, and Xiao-Ming Wu.
\newblock Slake: a semantically-labeled knowledge-enhanced dataset for medical
  visual question answering.
\newblock In \emph{2021 IEEE 18th International Symposium on Biomedical Imaging
  (ISBI)}, pages 1650--1654. IEEE, 2021{\natexlab{a}}.

\bibitem[Chen et~al.(2021)Chen, Tang, Qin, Liang, Liu, Xing, and
  Lin]{chen2021geoqa}
Jiaqi Chen, Jianheng Tang, Jinghui Qin, Xiaodan Liang, Lingbo Liu, Eric~P Xing,
  and Liang Lin.
\newblock Geoqa: A geometric question answering benchmark towards multimodal
  numerical reasoning.
\newblock \emph{arXiv preprint arXiv:2105.14517}, 2021.

\bibitem[Lu et~al.(2021)Lu, Qiu, Chen, Xia, Zhao, Zhang, Yu, Liang, and
  Zhu]{lu2021iconqa}
Pan Lu, Liang Qiu, Jiaqi Chen, Tony Xia, Yizhou Zhao, Wei Zhang, Zhou Yu,
  Xiaodan Liang, and Song-Chun Zhu.
\newblock Iconqa: A new benchmark for abstract diagram understanding and visual
  language reasoning.
\newblock In \emph{The 35th Conference on Neural Information Processing Systems
  (NeurIPS) Track on Datasets and Benchmarks}, 2021.

\bibitem[Lindstr{\"o}m and Abraham(2022)]{lindstrom2022clevr}
Adam~Dahlgren Lindstr{\"o}m and Savitha~Sam Abraham.
\newblock Clevr-math: A dataset for compositional language, visual and
  mathematical reasoning.
\newblock \emph{arXiv preprint arXiv:2208.05358}, 2022.

\bibitem[Lu et~al.(2022)Lu, Mishra, Xia, Qiu, Chang, Zhu, Tafjord, Clark, and
  Kalyan]{lu2022learn}
Pan Lu, Swaroop Mishra, Tanglin Xia, Liang Qiu, Kai-Wei Chang, Song-Chun Zhu,
  Oyvind Tafjord, Peter Clark, and Ashwin Kalyan.
\newblock Learn to explain: Multimodal reasoning via thought chains for science
  question answering.
\newblock \emph{Advances in Neural Information Processing Systems},
  35:\penalty0 2507--2521, 2022.

\bibitem[Masry et~al.(2022{\natexlab{a}})Masry, Do, Tan, Joty, and
  Hoque]{masry-etal-2022-chartqa}
Ahmed Masry, Xuan~Long Do, Jia~Qing Tan, Shafiq Joty, and Enamul Hoque.
\newblock {C}hart{QA}: A benchmark for question answering about charts with
  visual and logical reasoning.
\newblock In Smaranda Muresan, Preslav Nakov, and Aline Villavicencio, editors,
  \emph{Findings of the Association for Computational Linguistics: ACL 2022},
  pages 2263--2279, Dublin, Ireland, May 2022{\natexlab{a}}. Association for
  Computational Linguistics.
\newblock \doi{10.18653/v1/2022.findings-acl.177}.
\newblock URL \url{https://aclanthology.org/2022.findings-acl.177}.

\bibitem[Liu et~al.(2023{\natexlab{b}})Liu, Duan, Zhang, Li, Zhang, Zhao, Yuan,
  Wang, He, Liu, et~al.]{liu2023mmbench}
Yuan Liu, Haodong Duan, Yuanhan Zhang, Bo~Li, Songyang Zhang, Wangbo Zhao, Yike
  Yuan, Jiaqi Wang, Conghui He, Ziwei Liu, et~al.
\newblock Mmbench: Is your multi-modal model an all-around player?
\newblock \emph{arXiv preprint arXiv:2307.06281}, 2023{\natexlab{b}}.

\bibitem[Li et~al.(2023{\natexlab{b}})Li, Wang, Wang, Ge, Ge, and
  Shan]{li2023seed}
Bohao Li, Rui Wang, Guangzhi Wang, Yuying Ge, Yixiao Ge, and Ying Shan.
\newblock Seed-bench: Benchmarking multimodal llms with generative
  comprehension.
\newblock \emph{arXiv preprint arXiv:2307.16125}, 2023{\natexlab{b}}.

\bibitem[Yu et~al.(2023{\natexlab{b}})Yu, Yang, Li, Wang, Lin, Liu, Wang, and
  Wang]{yu2023mm}
Weihao Yu, Zhengyuan Yang, Linjie Li, Jianfeng Wang, Kevin Lin, Zicheng Liu,
  Xinchao Wang, and Lijuan Wang.
\newblock Mm-vet: Evaluating large multimodal models for integrated
  capabilities.
\newblock \emph{arXiv preprint arXiv:2308.02490}, 2023{\natexlab{b}}.

\bibitem[Yue et~al.(2024)Yue, Ni, Zhang, Zheng, Liu, Zhang, Stevens, Jiang,
  Ren, Sun, Wei, Yu, Yuan, Sun, Yin, Zheng, Yang, Liu, Huang, Sun, Su, and
  Chen]{Yue_2024_CVPR}
Xiang Yue, Yuansheng Ni, Kai Zhang, Tianyu Zheng, Ruoqi Liu, Ge~Zhang, Samuel
  Stevens, Dongfu Jiang, Weiming Ren, Yuxuan Sun, Cong Wei, Botao Yu, Ruibin
  Yuan, Renliang Sun, Ming Yin, Boyuan Zheng, Zhenzhu Yang, Yibo Liu, Wenhao
  Huang, Huan Sun, Yu~Su, and Wenhu Chen.
\newblock Mmmu: A massive multi-discipline multimodal understanding and
  reasoning benchmark for expert agi.
\newblock In \emph{Proceedings of the IEEE/CVF Conference on Computer Vision
  and Pattern Recognition (CVPR)}, pages 9556--9567, June 2024.

\bibitem[Yu et~al.(2024)Yu, Yang, Ren, Li, Wang, Lin, Lin, Liu, Wang, and
  Wang]{yu2024mm}
Weihao Yu, Zhengyuan Yang, Linfeng Ren, Linjie Li, Jianfeng Wang, Kevin Lin,
  Chung-Ching Lin, Zicheng Liu, Lijuan Wang, and Xinchao Wang.
\newblock Mm-vet v2: A challenging benchmark to evaluate large multimodal
  models for integrated capabilities.
\newblock \emph{arXiv preprint arXiv:2408.00765}, 2024.

\bibitem[Lu et~al.(2023)Lu, Bansal, Xia, Liu, Li, Hajishirzi, Cheng, Chang,
  Galley, and Gao]{lu2023mathvista}
Pan Lu, Hritik Bansal, Tony Xia, Jiacheng Liu, Chunyuan Li, Hannaneh
  Hajishirzi, Hao Cheng, Kai-Wei Chang, Michel Galley, and Jianfeng Gao.
\newblock Mathvista: Evaluating mathematical reasoning of foundation models in
  visual contexts.
\newblock \emph{arXiv preprint arXiv:2310.02255}, 2023.

\bibitem[Lin et~al.(2014)Lin, Maire, Belongie, Hays, Perona, Ramanan,
  Doll{\'a}r, and Zitnick]{lin2014microsoft}
Tsung-Yi Lin, Michael Maire, Serge Belongie, James Hays, Pietro Perona, Deva
  Ramanan, Piotr Doll{\'a}r, and C~Lawrence Zitnick.
\newblock Microsoft coco: Common objects in context.
\newblock In \emph{European conference on computer vision}, pages 740--755.
  Springer, 2014.

\bibitem[Davis(2020)]{davis2020unanswerable}
Ernest Davis.
\newblock Unanswerable questions about images and texts.
\newblock \emph{Frontiers in Artificial Intelligence}, 3:\penalty0 51, 2020.

\bibitem[Ray et~al.(2016)Ray, Christie, Bansal, Batra, and
  Parikh]{ray2016question}
Arijit Ray, Gordon Christie, Mohit Bansal, Dhruv Batra, and Devi Parikh.
\newblock Question relevance in vqa: identifying non-visual and false-premise
  questions.
\newblock \emph{arXiv preprint arXiv:1606.06622}, 2016.

\bibitem[Stengel-Eskin et~al.(2022)Stengel-Eskin, Guallar-Blasco, Zhou, and
  Van~Durme]{stengel2022did}
Elias Stengel-Eskin, Jimena Guallar-Blasco, Yi~Zhou, and Benjamin Van~Durme.
\newblock Why did the chicken cross the road? rephrasing and analyzing
  ambiguous questions in vqa.
\newblock \emph{arXiv preprint arXiv:2211.07516}, 2022.

\bibitem[Bhattacharya et~al.(2019)Bhattacharya, Li, and
  Gurari]{bhattacharya2019does}
Nilavra Bhattacharya, Qing Li, and Danna Gurari.
\newblock Why does a visual question have different answers?
\newblock In \emph{Proceedings of the IEEE/CVF International Conference on
  Computer Vision}, pages 4271--4280, 2019.

\bibitem[Rajpurkar et~al.(2018)Rajpurkar, Jia, and Liang]{rajpurkar2018know}
Pranav Rajpurkar, Robin Jia, and Percy Liang.
\newblock Know what you don't know: Unanswerable questions for squad.
\newblock \emph{arXiv preprint arXiv:1806.03822}, 2018.

\bibitem[Van~Landeghem et~al.(2023)Van~Landeghem, Tito, Borchmann, Pietruszka,
  Joziak, Powalski, Jurkiewicz, Coustaty, Anckaert, Valveny,
  et~al.]{van2023document}
Jordy Van~Landeghem, Rub{\`e}n Tito, {\L}ukasz Borchmann, Micha{\l} Pietruszka,
  Pawel Joziak, Rafal Powalski, Dawid Jurkiewicz, Micka{\"e}l Coustaty,
  Bertrand Anckaert, Ernest Valveny, et~al.
\newblock Document understanding dataset and evaluation (dude).
\newblock In \emph{Proceedings of the IEEE/CVF International Conference on
  Computer Vision}, pages 19528--19540, 2023.

\bibitem[Toor et~al.(2017)Toor, Wechsler, and Nappi]{toor2017question}
Andeep~S Toor, Harry Wechsler, and Michele Nappi.
\newblock Question part relevance and editing for cooperative and context-aware
  vqa (c2vqa).
\newblock In \emph{Proceedings of the 15th International Workshop on
  Content-Based Multimedia Indexing}, pages 1--6, 2017.

\bibitem[Chandrasekaran et~al.(2018)Chandrasekaran, Prabhu, Yadav,
  Chattopadhyay, and Parikh]{chandrasekaran2018explanations}
Arjun Chandrasekaran, Viraj Prabhu, Deshraj Yadav, Prithvijit Chattopadhyay,
  and Devi Parikh.
\newblock Do explanations make vqa models more predictable to a human?
\newblock \emph{arXiv preprint arXiv:1810.12366}, 2018.

\bibitem[Mashrur et~al.(2023)Mashrur, Luo, Zaidi, and
  Robles-Kelly]{mashrur2023robust}
Akib Mashrur, Wei Luo, Nayyar~A Zaidi, and Antonio Robles-Kelly.
\newblock Robust visual question answering via semantic cross modal
  augmentation.
\newblock \emph{Computer Vision and Image Understanding}, page 103862, 2023.

\bibitem[Mahendru et~al.(2017)Mahendru, Prabhu, Mohapatra, Batra, and
  Lee]{mahendru2017promise}
Aroma Mahendru, Viraj Prabhu, Akrit Mohapatra, Dhruv Batra, and Stefan Lee.
\newblock The promise of premise: Harnessing question premises in visual
  question answering.
\newblock \emph{arXiv preprint arXiv:1705.00601}, 2017.

\bibitem[Miller(1995)]{miller1995wordnet}
George~A Miller.
\newblock Wordnet: a lexical database for english.
\newblock \emph{Communications of the ACM}, 38\penalty0 (11):\penalty0 39--41,
  1995.

\bibitem[Lau et~al.(2018{\natexlab{b}})Lau, Gayen, Ben~Abacha, and
  Demner-Fushman]{lau2018dataset}
Jason~J Lau, Soumya Gayen, Asma Ben~Abacha, and Dina Demner-Fushman.
\newblock A dataset of clinically generated visual questions and answers about
  radiology images.
\newblock \emph{Scientific data}, 5\penalty0 (1):\penalty0 1--10,
  2018{\natexlab{b}}.

\bibitem[Zhan et~al.(2020)Zhan, Liu, Fan, Chen, and Wu]{zhan2020medical}
Li-Ming Zhan, Bo~Liu, Lu~Fan, Jiaxin Chen, and Xiao-Ming Wu.
\newblock Medical visual question answering via conditional reasoning.
\newblock In \emph{Proceedings of the 28th ACM International Conference on
  Multimedia}, pages 2345--2354, 2020.

\bibitem[Do et~al.(2021)Do, Nguyen, Tjiputra, Tran, Tran, and
  Nguyen]{do2021multiple}
Tuong Do, Binh~X Nguyen, Erman Tjiputra, Minh Tran, Quang~D Tran, and Anh
  Nguyen.
\newblock Multiple meta-model quantifying for medical visual question
  answering.
\newblock In \emph{Medical Image Computing and Computer Assisted
  Intervention--MICCAI 2021: 24th International Conference, Strasbourg, France,
  September 27--October 1, 2021, Proceedings, Part V 24}, pages 64--74.
  Springer, 2021.

\bibitem[Nguyen et~al.(2019)Nguyen, Do, Nguyen, Do, Tjiputra, and
  Tran]{nguyen2019overcoming}
Binh~D Nguyen, Thanh-Toan Do, Binh~X Nguyen, Tuong Do, Erman Tjiputra, and
  Quang~D Tran.
\newblock Overcoming data limitation in medical visual question answering.
\newblock In \emph{Medical Image Computing and Computer Assisted
  Intervention--MICCAI 2019: 22nd International Conference, Shenzhen, China,
  October 13--17, 2019, Proceedings, Part IV 22}, pages 522--530. Springer,
  2019.

\bibitem[Liu et~al.(2021{\natexlab{b}})Liu, Zhan, and Wu]{liu2021contrastive}
Bo~Liu, Li-Ming Zhan, and Xiao-Ming Wu.
\newblock Contrastive pre-training and representation distillation for medical
  visual question answering based on radiology images.
\newblock In \emph{Medical Image Computing and Computer Assisted
  Intervention--MICCAI 2021: 24th International Conference, Strasbourg, France,
  September 27--October 1, 2021, Proceedings, Part II 24}, pages 210--220.
  Springer, 2021{\natexlab{b}}.

\bibitem[Peng et~al.(2018)Peng, Liu, and Rosen]{peng2018umass}
Yalei Peng, Feifan Liu, and Max~P Rosen.
\newblock Umass at imageclef medical visual question answering (med-vqa) 2018
  task.
\newblock In \emph{CLEF (Working Notes)}, 2018.

\bibitem[Gurari et~al.(2018{\natexlab{b}})Gurari, Li, Stangl, Guo, Lin,
  Grauman, Luo, and Bigham]{gurari2018vizwiz}
Danna Gurari, Qing Li, Abigale~J Stangl, Anhong Guo, Chi Lin, Kristen Grauman,
  Jiebo Luo, and Jeffrey~P Bigham.
\newblock Vizwiz grand challenge: Answering visual questions from blind people.
\newblock In \emph{Proceedings of the IEEE conference on computer vision and
  pattern recognition}, pages 3608--3617, 2018{\natexlab{b}}.

\bibitem[Li et~al.(2020{\natexlab{a}})Li, Yin, Li, Zhang, Hu, Zhang, Wang, Hu,
  Dong, Wei, et~al.]{li2020oscar}
Xiujun Li, Xi~Yin, Chunyuan Li, Pengchuan Zhang, Xiaowei Hu, Lei Zhang, Lijuan
  Wang, Houdong Hu, Li~Dong, Furu Wei, et~al.
\newblock Oscar: Object-semantics aligned pre-training for vision-language
  tasks.
\newblock In \emph{European Conference on Computer Vision}, pages 121--137.
  Springer, 2020{\natexlab{a}}.

\bibitem[Lobry et~al.(2019)Lobry, Murray, Marcos, and Tuia]{lobry2019visual}
Sylvain Lobry, Jesse Murray, Diego Marcos, and Devis Tuia.
\newblock Visual question answering from remote sensing images.
\newblock In \emph{IGARSS 2019-2019 IEEE International Geoscience and Remote
  Sensing Symposium}, pages 4951--4954. IEEE, 2019.

\bibitem[Lobry et~al.(2020)Lobry, Marcos, Murray, and Tuia]{lobry2020rsvqa}
Sylvain Lobry, Diego Marcos, Jesse Murray, and Devis Tuia.
\newblock Rsvqa: Visual question answering for remote sensing data.
\newblock \emph{IEEE Transactions on Geoscience and Remote Sensing},
  58\penalty0 (12):\penalty0 8555--8566, 2020.

\bibitem[Chappuis et~al.(2021)Chappuis, Lobry, Kellenberger, Saux, and
  Tuia]{chappuis2021find}
Christel Chappuis, Sylvain Lobry, Benjamin Kellenberger, Bertrand~Le Saux, and
  Devis Tuia.
\newblock How to find a good image-text embedding for remote sensing visual
  question answering?
\newblock \emph{arXiv preprint arXiv:2109.11848}, 2021.

\bibitem[Felix et~al.(2021)Felix, Repasky, Hodge, Zolfaghari, Abbasnejad, and
  Sherrah]{felix2021cross}
Rafael Felix, Boris Repasky, Samuel Hodge, Reza Zolfaghari, Ehsan Abbasnejad,
  and Jamie Sherrah.
\newblock Cross-modal visual question answering for remote sensing data: The
  international conference on digital image computing: Techniques and
  applications (dicta 2021).
\newblock In \emph{2021 Digital Image Computing: Techniques and Applications
  (DICTA)}, pages 1--9. IEEE, 2021.

\bibitem[Bazi et~al.(2022)Bazi, Al~Rahhal, Mekhalfi, Al~Zuair, and
  Melgani]{bazi2022bi}
Yakoub Bazi, Mohamad~Mahmoud Al~Rahhal, Mohamed~Lamine Mekhalfi,
  Mansour~Abdulaziz Al~Zuair, and Farid Melgani.
\newblock Bi-modal transformer-based approach for visual question answering in
  remote sensing imagery.
\newblock \emph{IEEE Transactions on Geoscience and Remote Sensing},
  60:\penalty0 1--11, 2022.

\bibitem[Zheng et~al.(2021)Zheng, Wang, Du, and Lu]{zheng2021mutual}
Xiangtao Zheng, Binqiang Wang, Xingqian Du, and Xiaoqiang Lu.
\newblock Mutual attention inception network for remote sensing visual question
  answering.
\newblock \emph{IEEE Transactions on Geoscience and Remote Sensing},
  60:\penalty0 1--14, 2021.

\bibitem[Zhang et~al.(2023{\natexlab{a}})Zhang, Jiao, Li, Liu, Chen, Liu, Li,
  and Guo]{zhang2023spatial}
Zixiao Zhang, Licheng Jiao, Lingling Li, Xu~Liu, Puhua Chen, Fang Liu, Yuxuan
  Li, and Zhicheng Guo.
\newblock A spatial hierarchical reasoning network for remote sensing visual
  question answering.
\newblock \emph{IEEE Transactions on Geoscience and Remote Sensing},
  2023{\natexlab{a}}.

\bibitem[Brown et~al.(2020)Brown, Mann, Ryder, Subbiah, Kaplan, Dhariwal,
  Neelakantan, Shyam, Sastry, Askell, et~al.]{brown2020language}
Tom Brown, Benjamin Mann, Nick Ryder, Melanie Subbiah, Jared~D Kaplan, Prafulla
  Dhariwal, Arvind Neelakantan, Pranav Shyam, Girish Sastry, Amanda Askell,
  et~al.
\newblock Language models are few-shot learners.
\newblock \emph{Advances in neural information processing systems},
  33:\penalty0 1877--1901, 2020.

\bibitem[Zhou et~al.(2020)Zhou, Mishra, Verma, Bhamidipati, and
  Wang]{zhou2020recommending}
Yichao Zhou, Shaunak Mishra, Manisha Verma, Narayan Bhamidipati, and Wei Wang.
\newblock Recommending themes for ad creative design via visual-linguistic
  representations.
\newblock In \emph{Proceedings of The Web Conference 2020}, pages 2521--2527,
  2020.

\bibitem[Kembhavi et~al.(2016)Kembhavi, Salvato, Kolve, Seo, Hajishirzi, and
  Farhadi]{kembhavi2016diagram}
Aniruddha Kembhavi, Mike Salvato, Eric Kolve, Minjoon Seo, Hannaneh Hajishirzi,
  and Ali Farhadi.
\newblock A diagram is worth a dozen images.
\newblock In \emph{Computer Vision--ECCV 2016: 14th European Conference,
  Amsterdam, The Netherlands, October 11--14, 2016, Proceedings, Part IV 14},
  pages 235--251. Springer, 2016.

\bibitem[Gomez-Perez and Ortega(2020)]{gomez2020isaaq}
Jose~Manuel Gomez-Perez and Raul Ortega.
\newblock Isaaq--mastering textbook questions with pre-trained transformers and
  bottom-up and top-down attention.
\newblock \emph{arXiv preprint arXiv:2010.00562}, 2020.

\bibitem[Haurilet et~al.(2018)Haurilet, Al-Halah, and
  Stiefelhagen]{haurilet2018moqa}
Monica Haurilet, Ziad Al-Halah, and Rainer Stiefelhagen.
\newblock Moqa-a multi-modal question answering architecture.
\newblock In \emph{Proceedings of the European Conference on Computer Vision
  (ECCV) Workshops}, pages 0--0, 2018.

\bibitem[Li et~al.(2018{\natexlab{a}})Li, Su, Zhu, Wang, and
  Zhang]{li2018textbook}
Juzheng Li, Hang Su, Jun Zhu, Siyu Wang, and Bo~Zhang.
\newblock Textbook question answering under instructor guidance with memory
  networks.
\newblock In \emph{Proceedings of the IEEE Conference on Computer Vision and
  Pattern Recognition}, pages 3655--3663, 2018{\natexlab{a}}.

\bibitem[Wang et~al.(2023)Wang, Wei, Liu, Lin, Zhang, and Wu]{9996417}
Yaxian Wang, Bifan Wei, Jun Liu, Qika Lin, Lingling Zhang, and Yaqiang Wu.
\newblock Spatial-semantic collaborative graph network for textbook question
  answering.
\newblock \emph{IEEE Transactions on Circuits and Systems for Video
  Technology}, 33\penalty0 (7):\penalty0 3214--3228, 2023.
\newblock \doi{10.1109/TCSVT.2022.3231463}.

\bibitem[Ma et~al.(2022)Ma, Chai, Huang, Liu, You, and Zheng]{ma2022weakly}
Jie Ma, Qi~Chai, Jingyue Huang, Jun Liu, Yang You, and Qinghua Zheng.
\newblock Weakly supervised learning for textbook question answering.
\newblock \emph{IEEE Transactions on Image Processing}, 31:\penalty0
  7378--7388, 2022.

\bibitem[Li et~al.(2018{\natexlab{b}})Li, Su, Zhu, and Zhang]{li2018essay}
Juzheng Li, Hang Su, Jun Zhu, and Bo~Zhang.
\newblock Essay-anchor attentive multi-modal bilinear pooling for textbook
  question answering.
\newblock In \emph{2018 IEEE International Conference on Multimedia and Expo
  (ICME)}, pages 1--6. IEEE, 2018{\natexlab{b}}.

\bibitem[Chen et~al.(2022{\natexlab{b}})Chen, Li, Qin, Lu, Lin, Chen, and
  Liang]{chen2022unigeo}
Jiaqi Chen, Tong Li, Jinghui Qin, Pan Lu, Liang Lin, Chongyu Chen, and Xiaodan
  Liang.
\newblock Unigeo: Unifying geometry logical reasoning via reformulating
  mathematical expression.
\newblock \emph{arXiv preprint arXiv:2212.02746}, 2022{\natexlab{b}}.

\bibitem[Zhang et~al.(2023{\natexlab{b}})Zhang, Yin, and Liu]{zhang2023multi}
Ming-Liang Zhang, Fei Yin, and Cheng-Lin Liu.
\newblock A multi-modal neural geometric solver with textual clauses parsed
  from diagram.
\newblock \emph{arXiv preprint arXiv:2302.11097}, 2023{\natexlab{b}}.

\bibitem[Cao and Xiao(2022)]{cao-xiao-2022-augmented}
Jie Cao and Jing Xiao.
\newblock An augmented benchmark dataset for geometric question answering
  through dual parallel text encoding.
\newblock In \emph{Proceedings of the 29th International Conference on
  Computational Linguistics}, pages 1511--1520, Gyeongju, Republic of Korea,
  October 2022. International Committee on Computational Linguistics.
\newblock URL \url{https://aclanthology.org/2022.coling-1.130}.

\bibitem[He et~al.(2017)He, Xia, Yu, Jian, Meng, and Chen]{he2017educational}
Bin He, Meng Xia, Xinguo Yu, Pengpeng Jian, Hao Meng, and Zhanwen Chen.
\newblock An educational robot system of visual question answering for
  preschoolers.
\newblock In \emph{2017 2nd international conference on robotics and automation
  engineering (ICRAE)}, pages 441--445. IEEE, 2017.

\bibitem[Dou et~al.(2022)Dou, Xu, Gan, Wang, Wang, Wang, Zhu, Zhang, Yuan,
  Peng, et~al.]{dou2022empirical}
Zi-Yi Dou, Yichong Xu, Zhe Gan, Jianfeng Wang, Shuohang Wang, Lijuan Wang,
  Chenguang Zhu, Pengchuan Zhang, Lu~Yuan, Nanyun Peng, et~al.
\newblock An empirical study of training end-to-end vision-and-language
  transformers.
\newblock In \emph{Proceedings of the IEEE/CVF Conference on Computer Vision
  and Pattern Recognition}, pages 18166--18176, 2022.

\bibitem[Chen et~al.(2015)Chen, Wang, Chen, Gao, Xu, and Nevatia]{chen2015abc}
Kan Chen, Jiang Wang, Liang-Chieh Chen, Haoyuan Gao, Wei Xu, and Ram Nevatia.
\newblock Abc-cnn: An attention based convolutional neural network for visual
  question answering.
\newblock \emph{arXiv preprint arXiv:1511.05960}, 2015.

\bibitem[Jiang et~al.(2018)Jiang, Natarajan, Chen, Rohrbach, Batra, and
  Parikh]{jiang2018pythia}
Yu~Jiang, Vivek Natarajan, Xinlei Chen, Marcus Rohrbach, Dhruv Batra, and Devi
  Parikh.
\newblock Pythia v0. 1: the winning entry to the vqa challenge 2018.
\newblock \emph{arXiv preprint arXiv:1807.09956}, 2018.

\bibitem[Zhao et~al.(2021)Zhao, Samel, Chen, et~al.]{zhao2021proto}
Zelin Zhao, Karan Samel, Binghong Chen, et~al.
\newblock Proto: Program-guided transformer for program-guided tasks.
\newblock \emph{Advances in Neural Information Processing Systems},
  34:\penalty0 17021--17036, 2021.

\bibitem[Norcliffe-Brown et~al.(2018)Norcliffe-Brown, Vafeias, and
  Parisot]{norcliffe2018learning}
Will Norcliffe-Brown, Stathis Vafeias, and Sarah Parisot.
\newblock Learning conditioned graph structures for interpretable visual
  question answering.
\newblock \emph{Advances in neural information processing systems}, 31, 2018.

\bibitem[Gao et~al.(2019)Gao, You, Zhang, Wang, and Li]{gao2019multi}
Peng Gao, Haoxuan You, Zhanpeng Zhang, Xiaogang Wang, and Hongsheng Li.
\newblock Multi-modality latent interaction network for visual question
  answering.
\newblock In \emph{Proceedings of the IEEE/CVF international conference on
  computer vision}, pages 5825--5835, 2019.

\bibitem[Wu et~al.(2018)Wu, Liu, Wang, and Dong]{wu2018chain}
Chenfei Wu, Jinlai Liu, Xiaojie Wang, and Xuan Dong.
\newblock Chain of reasoning for visual question answering.
\newblock \emph{Advances in Neural Information Processing Systems}, 31, 2018.

\bibitem[Li et~al.(2019{\natexlab{b}})Li, Gan, Cheng, and Liu]{li2019relation}
Linjie Li, Zhe Gan, Yu~Cheng, and Jingjing Liu.
\newblock Relation-aware graph attention network for visual question answering.
\newblock In \emph{Proceedings of the IEEE/CVF international conference on
  computer vision}, pages 10313--10322, 2019{\natexlab{b}}.

\bibitem[Hudson and Manning(2019{\natexlab{b}})]{hudson2019learning}
Drew Hudson and Christopher~D Manning.
\newblock Learning by abstraction: The neural state machine.
\newblock \emph{Advances in Neural Information Processing Systems}, 32,
  2019{\natexlab{b}}.

\bibitem[Huang et~al.(2020)Huang, Zeng, Liu, Fu, and Fu]{huang2020pixel}
Zhicheng Huang, Zhaoyang Zeng, Bei Liu, Dongmei Fu, and Jianlong Fu.
\newblock Pixel-bert: Aligning image pixels with text by deep multi-modal
  transformers.
\newblock \emph{arXiv preprint arXiv:2004.00849}, 2020.

\bibitem[Chen et~al.(2020)Chen, Li, Yu, El~Kholy, Ahmed, Gan, Cheng, and
  Liu]{chen2020uniter}
Yen-Chun Chen, Linjie Li, Licheng Yu, Ahmed El~Kholy, Faisal Ahmed, Zhe Gan,
  Yu~Cheng, and Jingjing Liu.
\newblock Uniter: Universal image-text representation learning.
\newblock In \emph{European conference on computer vision}, pages 104--120.
  Springer, 2020.

\bibitem[Shen et~al.(2021)Shen, Li, Tan, Bansal, Rohrbach, Chang, Yao, and
  Keutzer]{shen2021much}
Sheng Shen, Liunian~Harold Li, Hao Tan, Mohit Bansal, Anna Rohrbach, Kai-Wei
  Chang, Zhewei Yao, and Kurt Keutzer.
\newblock How much can clip benefit vision-and-language tasks?
\newblock \emph{arXiv preprint arXiv:2107.06383}, 2021.

\bibitem[Li et~al.(2020{\natexlab{b}})Li, Gao, Niu, Xiao, Liu, Liu, Wu, and
  Wang]{li2020unimo}
Wei Li, Can Gao, Guocheng Niu, Xinyan Xiao, Hao Liu, Jiachen Liu, Hua Wu, and
  Haifeng Wang.
\newblock Unimo: Towards unified-modal understanding and generation via
  cross-modal contrastive learning.
\newblock \emph{arXiv preprint arXiv:2012.15409}, 2020{\natexlab{b}}.

\bibitem[Zhang et~al.(2021)Zhang, Li, Hu, Yang, Zhang, Wang, Choi, and
  Gao]{zhang2021vinvl}
Pengchuan Zhang, Xiujun Li, Xiaowei Hu, Jianwei Yang, Lei Zhang, Lijuan Wang,
  Yejin Choi, and Jianfeng Gao.
\newblock Vinvl: Revisiting visual representations in vision-language models.
\newblock In \emph{Proceedings of the IEEE/CVF Conference on Computer Vision
  and Pattern Recognition}, pages 5579--5588, 2021.

\bibitem[Cho et~al.(2021)Cho, Lei, Tan, and Bansal]{cho2021unifying}
Jaemin Cho, Jie Lei, Hao Tan, and Mohit Bansal.
\newblock Unifying vision-and-language tasks via text generation.
\newblock In \emph{International Conference on Machine Learning}, pages
  1931--1942. PMLR, 2021.

\bibitem[Kamath et~al.(2021)Kamath, Singh, LeCun, Synnaeve, Misra, and
  Carion]{kamath2021mdetr}
Aishwarya Kamath, Mannat Singh, Yann LeCun, Gabriel Synnaeve, Ishan Misra, and
  Nicolas Carion.
\newblock Mdetr-modulated detection for end-to-end multi-modal understanding.
\newblock In \emph{Proceedings of the IEEE/CVF International Conference on
  Computer Vision}, pages 1780--1790, 2021.

\bibitem[Yu et~al.(2021)Yu, Tang, Yin, Sun, Tian, Wu, and Wang]{yu2021ernie}
Fei Yu, Jiji Tang, Weichong Yin, Yu~Sun, Hao Tian, Hua Wu, and Haifeng Wang.
\newblock Ernie-vil: Knowledge enhanced vision-language representations through
  scene graphs.
\newblock In \emph{Proceedings of the AAAI Conference on Artificial
  Intelligence}, volume~35, pages 3208--3216, 2021.

\bibitem[Yuan et~al.(2021)Yuan, Chen, Chen, Codella, Dai, Gao, Hu, Huang, Li,
  Li, et~al.]{yuan2021florence}
Lu~Yuan, Dongdong Chen, Yi-Ling Chen, Noel Codella, Xiyang Dai, Jianfeng Gao,
  Houdong Hu, Xuedong Huang, Boxin Li, Chunyuan Li, et~al.
\newblock Florence: A new foundation model for computer vision.
\newblock \emph{arXiv preprint arXiv:2111.11432}, 2021.

\bibitem[Li et~al.(2022{\natexlab{a}})Li, Gao, Niu, Xiao, Liu, Liu, Wu, and
  Wang]{li2022unimo}
Wei Li, Can Gao, Guocheng Niu, Xinyan Xiao, Hao Liu, Jiachen Liu, Hua Wu, and
  Haifeng Wang.
\newblock Unimo-2: End-to-end unified vision-language grounded learning.
\newblock \emph{arXiv preprint arXiv:2203.09067}, 2022{\natexlab{a}}.

\bibitem[Wang et~al.(2022{\natexlab{c}})Wang, Yang, Hu, Li, Lin, Gan, Liu, Liu,
  and Wang]{wang2022git}
Jianfeng Wang, Zhengyuan Yang, Xiaowei Hu, Linjie Li, Kevin Lin, Zhe Gan,
  Zicheng Liu, Ce~Liu, and Lijuan Wang.
\newblock Git: A generative image-to-text transformer for vision and language.
\newblock \emph{arXiv preprint arXiv:2205.14100}, 2022{\natexlab{c}}.

\bibitem[Wang et~al.(2022{\natexlab{d}})Wang, Bao, Dong, Bjorck, Peng, Liu,
  Aggarwal, Mohammed, Singhal, Som, et~al.]{wang2022image}
Wenhui Wang, Hangbo Bao, Li~Dong, Johan Bjorck, Zhiliang Peng, Qiang Liu, Kriti
  Aggarwal, Owais~Khan Mohammed, Saksham Singhal, Subhojit Som, et~al.
\newblock Image as a foreign language: Beit pretraining for all vision and
  vision-language tasks.
\newblock \emph{arXiv preprint arXiv:2208.10442}, 2022{\natexlab{d}}.

\bibitem[Brock et~al.(2021)Brock, De, Smith, and Simonyan]{brock2021high}
Andy Brock, Soham De, Samuel~L Smith, and Karen Simonyan.
\newblock High-performance large-scale image recognition without normalization.
\newblock In \emph{International Conference on Machine Learning}, pages
  1059--1071. PMLR, 2021.

\bibitem[Chen et~al.(2022{\natexlab{c}})Chen, Wang, Changpinyo, Piergiovanni,
  Padlewski, Salz, Goodman, Grycner, Mustafa, Beyer, et~al.]{chen2022pali}
Xi~Chen, Xiao Wang, Soravit Changpinyo, AJ~Piergiovanni, Piotr Padlewski,
  Daniel Salz, Sebastian Goodman, Adam Grycner, Basil Mustafa, Lucas Beyer,
  et~al.
\newblock Pali: A jointly-scaled multilingual language-image model.
\newblock \emph{arXiv preprint arXiv:2209.06794}, 2022{\natexlab{c}}.

\bibitem[Li et~al.(2022{\natexlab{b}})Li, Xu, Tian, Wang, Yan, Bi, Ye, Chen,
  Xu, Cao, et~al.]{li2022mplug}
Chenliang Li, Haiyang Xu, Junfeng Tian, Wei Wang, Ming Yan, Bin Bi, Jiabo Ye,
  Hehong Chen, Guohai Xu, Zheng Cao, et~al.
\newblock mplug: Effective and efficient vision-language learning by
  cross-modal skip-connections.
\newblock \emph{arXiv preprint arXiv:2205.12005}, 2022{\natexlab{b}}.

\bibitem[Masry et~al.(2022{\natexlab{b}})Masry, Do, Tan, Joty, and
  Hoque]{bai2023qwen}
Ahmed Masry, Xuan~Long Do, Jia~Qing Tan, Shafiq Joty, and Enamul Hoque.
\newblock {C}hart{QA}: A benchmark for question answering about charts with
  visual and logical reasoning.
\newblock In Smaranda Muresan and Aline Nakov, Preslav
  andbai2023qwen~Villavicencio, editors, \emph{Findings of the Association for
  Computational Linguistics: ACL 2022}, pages 2263--2279, Dublin, Ireland, May
  2022{\natexlab{b}}. Association for Computational Linguistics.
\newblock \doi{10.18653/v1/2022.findings-acl.177}.
\newblock URL \url{https://aclanthology.org/2022.findings-acl.177}.

\bibitem[Liu et~al.(2024{\natexlab{b}})Liu, Li, Li, and Lee]{liu2024improved}
Haotian Liu, Chunyuan Li, Yuheng Li, and Yong~Jae Lee.
\newblock Improved baselines with visual instruction tuning.
\newblock In \emph{Proceedings of the IEEE/CVF Conference on Computer Vision
  and Pattern Recognition}, pages 26296--26306, 2024{\natexlab{b}}.

\bibitem[Zheng et~al.(2023)Zheng, Chiang, Sheng, Zhuang, Wu, Zhuang, Lin, Li,
  Li, Xing, et~al.]{zheng2023judging}
Lianmin Zheng, Wei-Lin Chiang, Ying Sheng, Siyuan Zhuang, Zhanghao Wu, Yonghao
  Zhuang, Zi~Lin, Zhuohan Li, Dacheng Li, Eric Xing, et~al.
\newblock Judging llm-as-a-judge with mt-bench and chatbot arena.
\newblock \emph{Advances in Neural Information Processing Systems},
  36:\penalty0 46595--46623, 2023.

\bibitem[Clark et~al.(2020)Clark, Luong, Le, and Manning]{clark2020electra}
Kevin Clark, Minh-Thang Luong, Quoc~V Le, and Christopher~D Manning.
\newblock Electra: Pre-training text encoders as discriminators rather than
  generators.
\newblock \emph{arXiv preprint arXiv:2003.10555}, 2020.

\bibitem[Lan et~al.(2019)Lan, Chen, Goodman, Gimpel, Sharma, and
  Soricut]{lan2019albert}
Zhenzhong Lan, Mingda Chen, Sebastian Goodman, Kevin Gimpel, Piyush Sharma, and
  Radu Soricut.
\newblock Albert: A lite bert for self-supervised learning of language
  representations.
\newblock \emph{arXiv preprint arXiv:1909.11942}, 2019.

\bibitem[He et~al.(2020{\natexlab{b}})He, Liu, Gao, and Chen]{he2020deberta}
Pengcheng He, Xiaodong Liu, Jianfeng Gao, and Weizhu Chen.
\newblock Deberta: Decoding-enhanced bert with disentangled attention.
\newblock \emph{arXiv preprint arXiv:2006.03654}, 2020{\natexlab{b}}.

\bibitem[Yu et~al.(2018)Yu, Yu, Xiang, Fan, and Tao]{yu2018beyond}
Zhou Yu, Jun Yu, Chenchao Xiang, Jianping Fan, and Dacheng Tao.
\newblock Beyond bilinear: Generalized multimodal factorized high-order pooling
  for visual question answering.
\newblock \emph{IEEE transactions on neural networks and learning systems},
  29\penalty0 (12):\penalty0 5947--5959, 2018.

\end{thebibliography}

\end{document}